\documentclass{article} 
\usepackage[preprint]{colm2025_conference}
\usepackage{microtype}
\usepackage{graphicx}
\usepackage{booktabs} 
\usepackage{natbib}
\usepackage{multirow}
\usepackage{minitoc}

\usepackage{algorithm}
\usepackage{algorithmic}

\usepackage{amsmath}
\usepackage{amssymb}
\usepackage{mathtools}
\usepackage{amsthm}
\usepackage{bbm}
\usepackage{wrapfig} 

\usepackage{graphicx}
\usepackage[skip=8pt]{caption}
\usepackage{subcaption}
\usepackage{ifthen}

\usepackage{varioref}
\usepackage{hyperref}
\usepackage[dvipsnames]{xcolor}
\hypersetup{
    colorlinks,
    linkcolor={red!50!black},
    citecolor={blue!50!black},
    urlcolor={blue!80!black}
}
\usepackage[capitalize,noabbrev]{cleveref}

\definecolor{antiquefuchsia}{rgb}{0.75, 0.55, 0.65}

\theoremstyle{plain}
\newtheorem{theorem}{Theorem}[section]

\newtheorem{lemma}{Lemma}

\newtheorem{corollary}[theorem]{Corollary}
\theoremstyle{definition}

\theoremstyle{remark}

\usepackage{enumitem}
\usepackage{tcolorbox}
\tcbuselibrary{skins, breakable}

\usepackage{amsmath,amsfonts,bm}

\def\eqref#1{equation~\ref{#1}}

\def\1{\bm{1}}

\def\vg{{\bm{g}}}
\def\vh{{\bm{h}}}

\def\vu{{\bm{u}}}
\def\vv{{\bm{v}}}

\def\mG{{\bm{G}}}

\def\mI{{\bm{I}}}

\def\mM{{\bm{M}}}

\DeclareMathAlphabet{\mathsfit}{\encodingdefault}{\sfdefault}{m}{sl}
\SetMathAlphabet{\mathsfit}{bold}{\encodingdefault}{\sfdefault}{bx}{n}

\def\gA{{\mathcal{A}}}

\def\gN{{\mathcal{N}}}

\def\gP{{\mathcal{P}}}

\newcommand{\E}{\mathbb{E}}

\newcommand{\R}{\mathbb{R}}

\DeclareMathOperator{\Tr}{Tr}

\newtheorem*{proposition*}{Proposition}

\newcounter{resultcounter}

\crefname{result}{R}{R}
\Crefname{result}{R}{R}

\crefalias{resultcounter}{result}

\newcommand{\resultformat}[1]{\textbf{R#1}}

\crefformat{result}{(\resultformat{#1})}
\Crefformat{result}{(\resultformat{#1})}

\newcommand{\norm}[1]{\left\lVert #1 \right\rVert_2}

\def\teacher{\pi_{\cal T}}
\def\student{\pi_{\cal S}}

\def\prompt{\mathbf{x}}
\def\gen{\mathbf{y}}

\def\promptset{\mathcal{X}}

\def\response{y}

\def\loss{\mathcal{L}}

\def\numprompts{n}
\def\numgenerations{m}
\def\Numprompts{N}
\def\Numgenerations{M}
\def\gradientdim{d}
\def \modelparams {P}

\def\gsm{GSM8K}
\def\math{MATH}
\def\llamaS{LLaMA-1B}
\def\llamaM{LLaMA-3B}
\def\olmoS{OLMo-1B}

\def\gemmaS{Gemma-2B}

\def \gnorm {\text{G-Norm}}
\def \gradCov {\boldsymbol\Sigma}
\def \gvendi {\text{G-Vendi}}
\def \gradcv {\text{GRACE}}

\def \empD {\mathcal{D}}
\def \numCV {C}
\def \eval {\lambda} 
\def \evec {\vu} 

\def \precond {\mathbf{M}}

\def \idx {u}

\def \mm {\mathbf{m}}
\def \mv {\mathbf{v}}
\def \mg {\mathbf{g}}

\author{Abhishek Panigrahi$^{\dagger}$\text{ } Bingbin Liu $^{\alpha}$ \text{ }  Sadhika Malladi $^{\theta}$ \text{ }  Sham Kakade $^{\alpha}$ \text{ }  Surbhi Goel $^{\zeta}$  \\
$^{\dagger}$Princeton Language and Intelligence \quad $^{\alpha}$Kempner Institute, Harvard \\
 $^{\theta}$Microsoft Research, New York \quad 
$^{\zeta}$University of Pennsylvania \\
}

\title{In Good GRACEs: Principled Teacher Selection for Knowledge Distillation}

\begin{document}
\maketitle
\begingroup
\renewcommand\thefootnote{}\footnotetext{Correspondence to: \texttt{ap34@princeton.edu}, \texttt{bliu@g.harvard.edu}}
\addtocounter{footnote}{-1}
\endgroup

\begin{abstract}

\looseness-1Knowledge distillation is an efficient strategy to use data generated by large ``teacher'' language models to train smaller capable ``student'' models, but selecting the optimal teacher for a specific student-task combination requires expensive trial-and-error.
We propose a lightweight score called \gradcv{} to quantify how effective a teacher will be for post-training a student model.
\gradcv{} measures distributional properties of the student's gradients without access to a verifier, teacher logits, teacher internals, or test data. 
From an information-theoretic perspective, \gradcv{} connects to leave-one-out stability of gradient-based algorithms, which controls the generalization performance of the distilled students.
On \gsm{} and \math{}, \gradcv{} correlates strongly (up to $86\%$ Spearman correlation) with the performance of the distilled LLaMA and OLMo students.
In particular, training a student using the \gradcv{}-selected teacher can improve the performance by up to $7.4\%$ over naively using the best-performing teacher.
Further, \gradcv{} can provide guidance on crucial design choices in distillation, including (1) the best temperature to use when generating from the teacher, (2) the best teacher to use given a size constraint, and (3) the best teacher to use within a specific model family. 
Altogether, our findings demonstrate that \gradcv{} can efficiently and effectively identify a strongly compatible teacher for a given student and provide fine-grained guidance on how to perform distillation.
\end{abstract}

\section{Introduction}
\label{sec:intro}
\looseness-1Distillation is an efficient and effective method to produce capable small models from existing, powerful teacher models.  
In this work, we focus on the specific case of training autoregressive language models on text generated by a teacher model. 
It is difficult to select the right teacher for a given student and task:
a counterintuitive fact is that \textit{a stronger-performing model is not always a better teacher}, which has been observed in classic classification/regression settings~\citep{mirzadeh2019improved,harutyunyan23supervision,panigrahi2025progressive} and more recently in the context of language models~\citep{zhang2023lifting,zhang2023towards,peng2024pretraining,razin2025makes}.
Given the large number of available models as potential teachers,
the current approach of guess-and-check is costly, because it requires collecting generations from a capable teacher and subsequently training a student on those generations. 
Additionally, the specific hyperparameters used in both phases can dramatically affect the final performance of the student, underscoring the need for careful, repeated testing to select the right teacher.
As such, the current work seeks to address the following question:
\begin{center}
\emph{Given a pool of candidates, can we efficiently identify the best teacher for a given student and task?}
\end{center}

\begin{figure}
    \centering
    \includegraphics[width=0.48\linewidth]{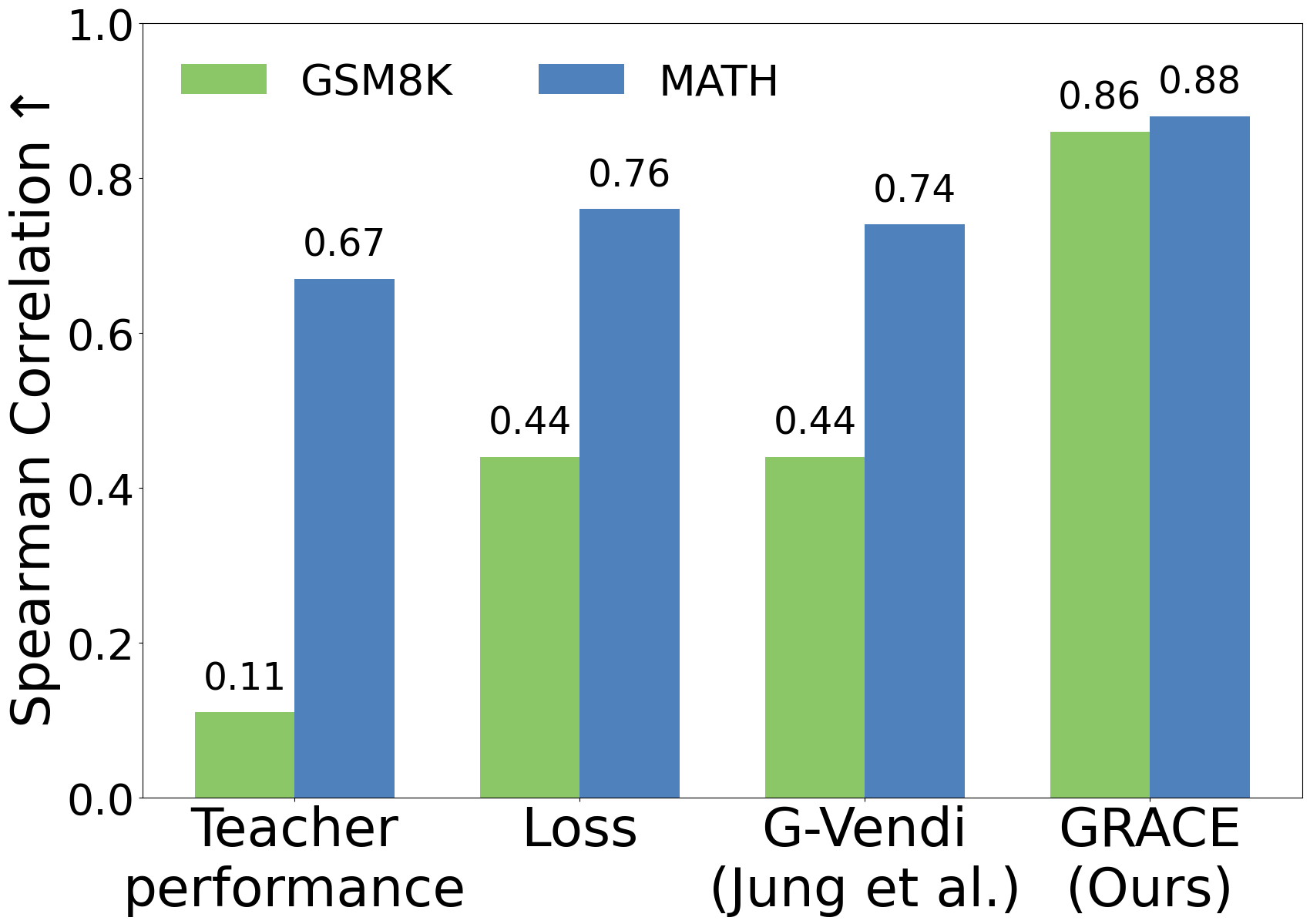}
    \includegraphics[width=0.5\linewidth]{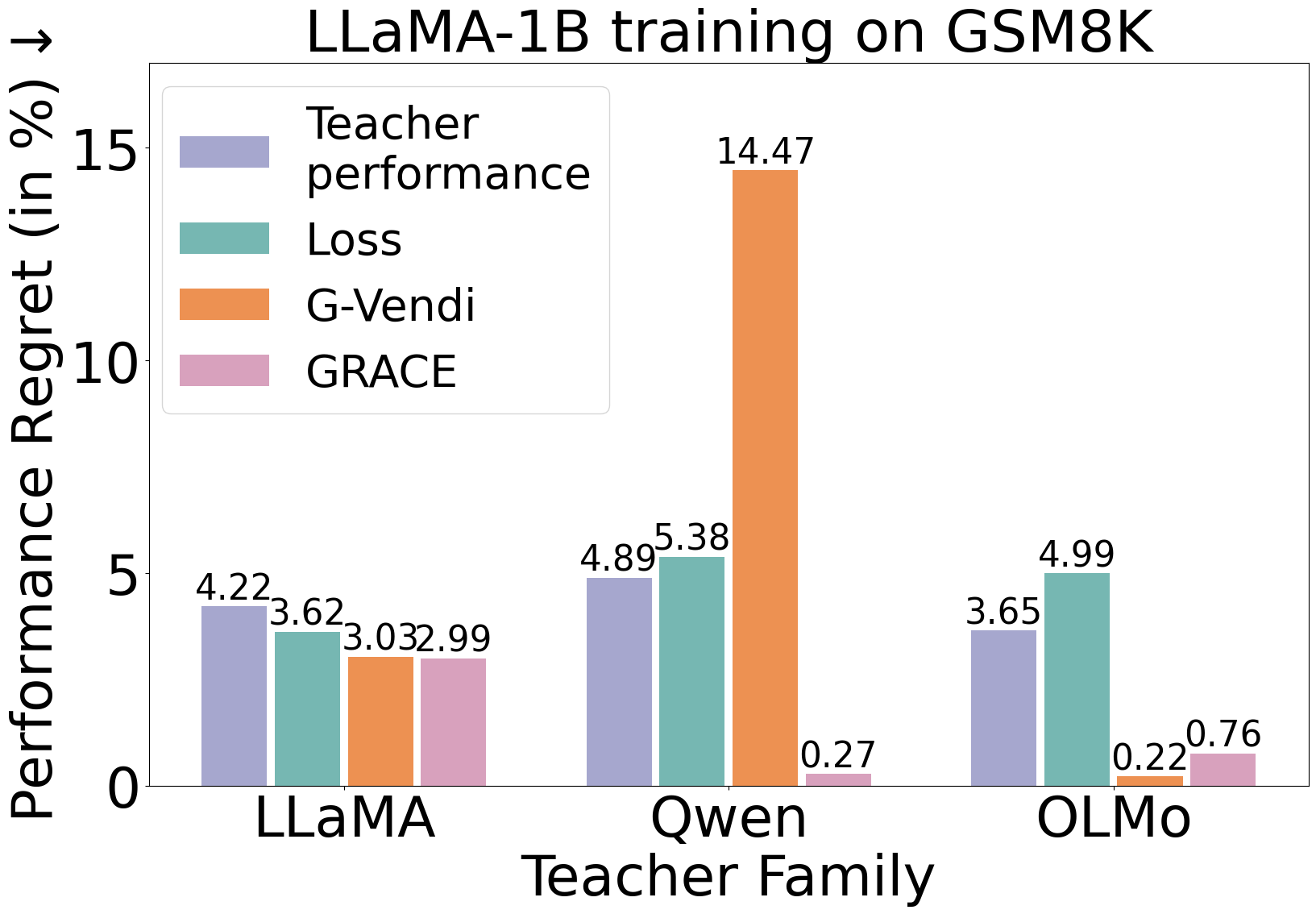}
    \caption{\textbf{
    \gradcv{} correlates with student performance after distillation on math-related reasoning tasks.} 
    We evaluate \llamaS{} and \llamaM{} students on GSM8K and MATH respectively with 15 teachers across LLaMA, Gemma, Qwen, OLMo, and Phi families.
    (Left) \textit{\gradcv{} shows the strongest correlation with student accuracy} among four scores---teacher performance, student's loss (before training) on the teacher generations, \gvendi{}, and \gradcv{}.
    (Right) \textit{\gradcv{} reliably selects near-optimal teachers within each teacher family}, measured by its small \textit{teacher-student regret}, which is the absolute gap in final performance between the best overall student and that obtained from the teacher chosen by each score. Performance is measured by average accuracy over 16 generations per prompt.
    }
    \label{fig:main_metrics}
\end{figure}

We argue that identifying a good teacher requires jointly considering the teacher and the student.
To this end, we study the teacher's influence on the student's optimization process by analyzing the student's gradients.
We propose a score, \textbf{\gradcv{}} (GRAdient Cross-validation Evaluation), which captures the distributional properties of the student's gradients on a small set of teacher-generated data, enabling efficient and effective identification of the most compatible teacher (\Cref{def:gradcv}).
Inspired by prior data selection and distillation works, \gradcv{} unifies data diversity and teacher-student alignment desiderata into a single score.
Specifically, \gradcv{} leverages a cross-validation structure,
where it takes gradients from one split of data and measures their magnitudes re-weighted under the spectrum of the other split.
The gradient spectrum reflects data diversity, while the gradient norm encodes teacher-student alignment.
This cross-validation structure also allows us to draw a natural connection to conditional mutual information-based generalization bounds~\citep{steinke2020reasoning,rammal2022leave1one1out}, providing insight into why \gradcv{} works (\Cref{lem:cmi_result}). 
Importantly, \gradcv{} is efficient to compute and does not rely on access to an external verifier, teacher logits, teacher representations, or test data.

We conduct thorough experiments to verify that 
the \gradcv{} scores 
of teachers accurately reflect the student's performance when trained on those teachers.
We focus on math-related datasets, namely GSM8K~\citep{cobbe2021gsm8k} and MATH~\citep{hendrycks2021MATH}, because the broad community interest in mathematical reasoning has led to the development of a large, diverse set of teachers that are readily available and suitable for distillation. 
We train LLaMA-1B-Base, OLMo-1B-Base and Gemma-2B-Base (for GSM8K) as well as LLaMA-3B-Base (for MATH) using generations sampled from $15$ candidate teachers drawn from the LLaMA~\citep{grattafiori2024llama}, OLMo~\citep{olmo20242}, Qwen~\citep{qwen25technicalreport}, Gemma~\citep{team2024gemma}, and Phi~\citep{abdin2024phi4} families. 
Our results show that:
\begin{itemize}[leftmargin=*]
    \item \gradcv{} correlates strongly with the student's distillation performance (\Cref{fig:main_metrics}, left), outperforming strong baselines such as \gvendi{}~\citep{jung2025prismatic}.
    
    \item \gradcv{} identifies teachers that are strongly compatible to a given student. Selecting teachers using \gradcv{} yields $7\%$  and $2\%$ improvement in student performance compared to using the best-performing teacher, on \gsm{} and \math{} respectively.

    \item \gradcv{} offers actionable insights to practitioners. It helps identify close to 1) optimal generation temperature for a given teacher model,
    2) best teacher within a size constraint,
    and 3) best teacher within a model family.
    
\end{itemize}
Given the limited research on teacher selection in LLM distillation, our findings indicate that \gradcv{} improves performance and offers reliable guidance for effective distillation.

\section{GRACE: Gradient Cross-Validation Evaluation}
\label{sec:grace}

We consider the case of using distillation to fine-tune a pre-trained student model to solve specific downstream tasks.
For each of the $\Numprompts$ prompts $\prompt \in \promptset$, we autoregressively generate $\Numgenerations$ responses $\response_1, ..., \response_M$ from a teacher distribution $\teacher$. This distribution encodes the temperature it may be sampled at from the teacher as well. We then fine-tune the pre-trained student with the standard autoregressive cross-entropy objective $\loss$ on a dataset $\empD^{\text{distill}}_{\cal T}$ containing $\Numprompts\times \Numgenerations$ teacher generations. 
In contrast to logit-based distillation, learning from generations permits distillation across architectures and in cases where the teacher's logits are not available. 
We measure the performance of students and teachers as the average accuracy of $k$ sampled responses for a given prompt (i.e., average-at-k). 
We will use $\student$ to denote the pre-trained student, and refer to its parameters as $\Theta_{\cal S} \in \R^D$ when necessary.

\subsection{Gradient-Based Scores}

The problem of selecting a teacher for distillation is closely connected to the well-studied field of data selection:
choosing the best teacher based on its generations can be viewed as selecting the best subset from the union of all teachers' generations, with the constraint that each subset must come from a single teacher.
For language models, many successful data selection methods \citep{pruthi2020estimating,xia24less,engstrom2024dsdm} rely on first- or second-order gradient information to identify useful data for a given task.

In contrast to selecting individual datapoints from a dataset,
we would like to select a data distribution (i.e., corresponding to a teacher). 
As such, instead of quantifying the value of individual datapoints, we turn our attention to gradient-based approaches to measure data quality in terms of its \textit{distributional features}.
For a teacher $\teacher$, we assume access to only a subsampled dataset $\empD^\text{eval}_{\cal T} \subset \empD^\text{distill}_{\cal T}$ 
containing $\numprompts \times \numgenerations$ prompt-generation pairs, where $\numprompts, \numgenerations$ may be much smaller than $\Numprompts,\Numgenerations$. In our experiments (\cref{sec:expr}), $\numprompts \times \numgenerations$ is $60 \times$ smaller compared to the $\Numprompts \times \Numgenerations$.

In the following, we will first describe two baselines (i.e., \gvendi{}~\citep{jung2025prismatic} and \gnorm{}), then introduce the proposed \gradcv{} score combining the advantages of both. 
The two baselines measure data diversity (via directional entropy) and teacher-student alignment (via gradient norms).
\gradcv{} constructs a single score that balances these two desiderata by penalizing large gradients along low-eigenvalue directions.

\looseness-1\paragraph{Gradients.} We establish some useful notation to work with gradients. 
Let $\vg(\prompt, \gen) := \frac{1}{|\gen|} \nabla \loss(\gen|\prompt; \Theta_{\cal S})$ be the student's gradient on the response $\gen$ conditioned on prompt $\prompt$, 
which is averaged over response tokens, following prior works on data selection.
Since all gradients are computed with respect to the student model's parameters, we omit the explicit dependency on $\Theta_{\cal{S}}$ for notational clarity.

We process the gradient with two steps.
First, for computational reasons, we work with a random low-dimensional projection of the gradient, denoted $\Pi \vg \in \R^\gradientdim$ with $\boldsymbol{\Pi} \in \{\pm 1/\sqrt{D}\}^{\gradientdim \times D}$~\citep{park2023trak}.
We also rescale the gradient by $\log |\gen|$ to account for the response length $|\gen|$.
This is motivated by the empirical observation that the gradient norm averaged over a length-$T$ sequence roughly decreases as $1/\log T$ (\Cref{fig:grad_norm_length}), which can cause gradient-based computations to unduly favor short sequences~\citep{xia24less}.

The processed gradient is denoted $\vh(\prompt, \gen) := \log(|\gen|) \cdot \Pi\vg(\prompt, \gen)$.
For a dataset $\empD$ of generations, we define the matrix consisting of processed gradients (i.e., $\vh$) as $\mG (\empD) \in \R^{\numprompts{\numgenerations} \times \gradientdim}$, and define the matrix consisting of processed and \textit{normalized} gradients (i.e., $\tilde\vh = \vh/\|\vh\|$) as $\tilde{\mG} (\empD) \in \R^{\numprompts{\numgenerations} \times \gradientdim}$.
Then, we define the mean and second moment of the unnormalized gradients as:
\begin{align}
\label{eq:mean_cov}
    \mu(\empD) := \frac{1}{{\numprompts} {\numgenerations}} \mG (\empD)^\top\mathbf{1},
    \quad
    \gradCov(\empD) :=& 
    \frac{1}{{\numprompts} {\numgenerations}} \mG(\empD)^\top \mG(\empD).
\end{align}
$\tilde\mu(\empD), \tilde\gradCov(\empD)$ are defined similarly. We additionally use $\Hat{{\boldsymbol\Sigma}}(\empD) = \tilde{\boldsymbol\Sigma}(\empD) + \frac{\nu}{\gradientdim} \mI $ with smoothing parameter $\nu >0$ for numerical stability.

\looseness-1Gradient information has been used for data selection, where samples with higher influence to a validation set is preferred~\citep{pruthi2020estimating,xia24less}.
One can adapt these methods for teacher selection. However, we find them to be not strongly predictive empirically; we provide details in \Cref{sec:gradient_influence}. Consequently, we evaluate training data quality by analyzing its distributional properties, without relying on a validation set.

\paragraph{G-Vendi~\citep{jung2025prismatic}.}
One natural distributional measure of data quality is diversity.
Along these lines, \citet{jung2025prismatic} propose the G-Vendi score, which measures the directional coverage of $\empD$ as the entropy of the eigenvalues of $\tilde\gradCov(\empD)$ (\cref{eq:mean_cov}):
\begin{align}
    \gvendi(\empD)  := \text{Entropy}(\eval(\tilde\gradCov(\empD)))
    = -\sum_{\lambda \in \eval(\tilde\gradCov(\empD))} \eval \log \eval, \label{eq:g-vendi}
\end{align}
where $\eval(\tilde\gradCov(\empD))$ denotes the set of eigenvalues of $\tilde\gradCov(\empD)$, with $|\eval(\tilde\gradCov(\empD))| = \min\{\numprompts\numgenerations, \gradientdim\}$.
A larger \gvendi{} score is preferred as it corresponds to better diversity.
\cite{jung2025prismatic} use \gvendi{} to select an optimal subset of training data $\empD$ from a full dataset generated by a single teacher.  
However, using \gvendi{} to select a teacher out of many candidates may yield suboptimal choices.
For example, when the student serves as its own teacher, we find that its \gvendi{} score (5.93) is higher than that of all other teachers (\cref{fig:across_all_temps}), even though the self-distilled student has a low accuracy of 4\%.
This occurs because an untrained model can produce random responses, which exhibit high gradient entropy and therefore a high \gvendi{} score, yet offer no meaningful learning signal for distillation.

\paragraph{\gnorm{}.} 
The limitation of \gvendi{} leads us to investigate another gradient-based distributional score, namely the gradient-norm score (\gnorm{}):
\begin{align}
    \gnorm(\empD)
     := \Tr\left(\gradCov(\empD)\right)
     = \frac{1}{\numprompts\numgenerations} \sum_{(\prompt, \gen) \in \empD}\|\vh(\prompt, \gen)\|^2.
    \label{eq:mean_grad_norm}
\end{align}
Because gradients become smaller near an optimum, a lower \gnorm{} indicates that the student requires less parameter updates to reduce loss on the given task.
\gnorm{} therefore serves as a proxy for teacher-student \textit{alignment}.
This explains why high-performing models may be poor teachers:
although Gemma-2 Instruct models achieve strong accuracy on \gsm{} and \math{}, the student model shows high \gnorm{} on their generated responses suggesting weak teacher-student alignment and shows weaker post-distillation performance (\cref{fig:across_all_temps}).
For students which have been pretrained, \gnorm{} can additionally reflect the teacher's generation \textit{quality},
as \gnorm{} is high when a teacher provides uninformative or random responses.
In particular, using the student itself as the teacher results in the highest \gnorm{}, showing that its outputs do not provide a meaningful distillation signal.

Nevertheless, \gnorm{} alone remains insufficient.
For instance, on \gsm{}, we find that \gnorm{}'s value largely differentiates different teacher families and does not correlate with the student's performance when generation temperature from the teacher is varied (\Cref{fig:across_all_temps,fig:gsm8k-comparative}).
This is because \gnorm{} captures only the magnitude of gradients, without considering the directional spread of the gradients across dimensions.


\looseness-1\gnorm{} and \gvendi{}  capture complementary distributional properties and can sometimes trend in different directions.
For instance, we find that increasing the teacher's generation temperature 
increases $\gnorm$, consistent with the observation that higher temperatures induce worse data, but at the same time increases $\gvendi$, indicating higher data diversity (\cref{fig:gsm8k-comparative}).
As such, we treat \gnorm{} and \gvendi{} as baselines and propose \gradcv{} as a single score that unifies both.

\subsection{The GRACE Score}\label{def:gradcv}
\looseness-1GRAdient Cross-validation Evaluation (\gradcv{}) computes the norm of the gradients weighted under the spectrum of the normalized gradient second-moment matrix,
thereby combining the advantages of \gvendi{} and \gnorm{}.
\gradcv{} is computed solely using the student's gradients on the teacher's generations, and does not rely on a verifier or access to test samples.
In the following, we will first define the \gradcv{} score, and then describe its connection to leave-one-out conditional mutual information.

\paragraph{GRACE.} For a dataset $\empD$ of teacher generations containing $\numprompts \times \numgenerations$ prompt-generation pairs and a choice of hyperparameter $\numCV$, construct $\numCV$ partitions of the prompts in the dataset $\empD$, denoted $\{\empD_{i}\}_{i=1}^{\numCV}$, each containing $\numprompts/\numCV$ prompts and their generations. Let $\empD_{-i}$ denote the concatenation of all partitions except the partition $\empD_{i}$. Then, \gradcv{} is defined as
    \begingroup
    \allowdisplaybreaks
    \begin{align}
        \gradcv{}(\empD) &= 
        \frac{1}{C} \sum_{i=1}^C\Tr\left(\Hat{{\boldsymbol\Sigma}}(\empD_{-i})^{-1} \gradCov(\empD_i)\right) \label{eq:gradcv}\\
        &= \frac{1}{\numprompts\numgenerations} \sum_{i=1}^C \sum_{(\prompt, \gen) \in \empD_{i}} \|\Hat{{\boldsymbol\Sigma}}(\empD_{-i} )^{-1/2}\vh(\prompt,\gen)\|^2,
    \end{align}
    \endgroup
    where $\Hat{{\boldsymbol\Sigma}}(\empD_{-i}) = \tilde{\boldsymbol\Sigma}(\empD_{-i}) + \frac{\nu}{\gradientdim} \mI $ with smoothing parameter $\nu >0$ for numerical stability.

\gradcv{} computes the expected squared norm of spectral-weighted gradients.
Given a random partition $(\empD_i, \empD_{-i})$, let $\{\eval_j, \evec_j\}_{j \in [\gradientdim]}$ denote the set of eigenvalues and eigenvectors for $\tilde{\boldsymbol\Sigma}(\empD_{-i})$.
For this partition, \gradcv{} computes
\begin{equation}
     \sum_{j \in [\gradientdim]} \frac{1}{\eval_j + \frac{\nu}{d}} \left(\frac{1}{|\empD_i|}\sum_{(\prompt, \gen) \in \empD_i}(\vh(\prompt, \gen)^{\top} \evec_j)^2\right).
 \end{equation}

\textbf{A bias-variance tradeoff.} \gradcv{} can be split into two terms:
\begin{align*}
    \gradcv{}(\empD) \approx \underbrace{
\frac{1}{\numprompts\numgenerations} \sum_{i=1}^C
\Tr\!\left(\Hat{{\boldsymbol\Sigma}}(\empD_{-i})^{-1}\mG_{\mu}(\empD_i)^\top\mG_{\mu}(\empD_i)\right)
}_{\text{\gradcv{}-Variance}(\empD)}
+
\underbrace{
\frac{1}{\numprompts\numgenerations} \sum_{i=1}^C
\mu(\empD)^{\top}\Hat{{\boldsymbol\Sigma}}(\empD_{-i})^{-1}\mu(\empD)
}_{\text{\gradcv{}-Bias}(\empD)},
\end{align*}
where $\mG_{\mu}(\empD) := \mG(\empD) - \mathbf{1}\mu(\empD)^{\top}$ denotes the centered gradient matrix.
The term $\text{\gradcv{}-Variance}(\empD)$ captures the variance in gradients under the spectrum of $\Hat{{\boldsymbol\Sigma}}$, while $\text{\gradcv{}-Bias}(\empD)$ measures the norm of the mean gradient under the spectrum of $\Hat{{\boldsymbol\Sigma}}$. 

Repeating our intuitions from \gnorm{}, $\text{\gradcv{}-Bias}$ is highly useful for identifying pathological teachers: for example, if a teacher provides random responses, $\text{\gradcv{}-Bias}$ becomes large, indicating that such data is not suitable for distillation.

\looseness-1When we do not encounter such teachers, most of the predictive power comes from $\text{\gradcv{}-Variance}$. In our experiments, $\text{\gradcv{}-Variance}$ dominates $\text{\gradcv{}-Bias}$, and our conclusions remain the same whether we consider $\text{\gradcv{}}$ or $\text{\gradcv{}-Variance}$.
A smaller $\text{\gradcv{}-Variance}$ indicates a better distillation teacher, as it corresponds to a smaller variance of gradients along all eigenvectors of $\tilde{\boldsymbol\Sigma}$. 
Gradient variance in directions with smaller eigenvalues are penalized more heavily,
which is desirable as high gradient variance along such directions can more easily induce instability or poor generalization. We formalize this view in \cref{sec:cmi}. The directional spectrum is taken from the \textit{normalized} gradients, since the gradient norm is less relevant than the gradient direction with the use of adaptive optimizers and normalization layers \citep{loshchilov2017decoupled,ba2016layer,li2022robust}.



\subsubsection{Connecting \gradcv{} to leave-one-out CMI}
\label{sec:cmi}
\gradcv{} connects naturally to leave-one-out conditional mutual information (CMI), a frequently used concept in studying generalization~\citep{xu2017information,steinke2020reasoning,rammal2022leave1one1out}.
CMI captures how sensitive the learning outcome is to the removal of a single sample.
A higher sensitivity suggests heavier memorization,
which can lead to low generalization to unseen test examples. Under this  framework, we show that \gradcv{} successfully unifies \gnorm{} and \gvendi{}.

Formally, we overload $\vg(\empD; \Theta) = \frac{1}{|\empD|} \sum_{(\prompt,\gen) \in \empD}\vg(\prompt, \gen; \Theta)$ to denote the average gradient on a dataset $\empD$. To keep our discussion general, we consider the following gradient update that uses gradients and a preconditioner matrix $\precond$:
\begin{align*}
     \Theta \leftarrow \Theta - \eta(\precond(\empD; \Theta) \vg(\empD;\Theta) + \epsilon),
\end{align*}
where $\eta$ denotes the learning rate and $\epsilon \sim \mathcal{N}(0, \sigma^2\mathbf{I})$ denotes the gradient noise.
Setting $\precond$ as identity recovers gradient descent, 
and setting $\precond$ as a function of gradient second moments recovers various adaptive algorithms in practice.

Let $\Theta'_{\empD}$ denote the resulting parameters after a gradient update with $\empD$, and $\Theta'_{\empD \setminus \{(\prompt, \cdot)\}}$ denote the parameters from a set where all training data connected to a uniformly sampled prompt $\prompt$ are dropped from the training set $\empD$. CMI measures the mutual information between the parameters $\Theta'_{\empD \setminus \{(\prompt, \cdot)\}}$ and the dropped prompt $\prompt$.
We show that CMI can be bounded as follows:

\begin{lemma}[Informal; cf \cref{cor:precond}]
    Define \gradcv{} with $\numCV = \numprompts$.
    For any $\empD'$,
    take $\precond( \empD', \Theta) := \Hat{{\boldsymbol\Sigma}}(\empD')^{-1/2}$, where 
    then CMI $\lesssim \frac{1}{\sigma^2 \numprompts^2}$ \gradcv{}-Variance $(\empD)$  $\lesssim \frac{1}{\sigma^2 \numprompts^2}$ \gradcv{} $(\empD)$.
    \label{lem:cmi_result}
\end{lemma}

\looseness-1Thus, \gradcv{} offers an upper bound on the student’s generalization performance after training with a single gradient update under the pre-conditioner matrix $ \Hat{{\boldsymbol\Sigma}}^{-1/2}$. Intuitively, \gradcv{} captures how uniformly spread the gradients are across examples, which mirrors the core idea behind CMI. Higher gradient uniformity leads to more stable optimization, which in turn can help stronger performance on held-out evaluation.


\textbf{Choice of $\precond$ for \gradcv{}.} We defined \gradcv{} based on a particular choice of the pre-conditioner matrix (i.e. $\precond(\empD', \Theta) = \Hat{{\boldsymbol\Sigma}}(\empD')^{-1/2}$) in the definition of CMI, which is motivated by adaptive optimizers used in practice \citep{kingma2014adam,loshchilov2017decoupled,duchi2011adaptive}. 
In principle, one could obtain even tighter upper bounds on CMI by choosing $\precond$ optimally. We leave a more thorough exploration of this direction to future work.

\section{Experiments} 
\label{sec:expr}

We compare the three scores mentioned in the previous section, \gnorm{}, \gvendi{}, and \gradcv{}, on two common math reasoning datasets, \gsm{}~\citep{cobbe2021gsm8k} and \math{}~\citep{hendrycks2021MATH}.
These datasets have a diverse set of strong teacher models readily available, due to the broad community interest in mathematical reasoning.
For each prompt-response pair, the model receives a binary correctness score, and we quantify its performance by the average accuracy (in percentage) achieved when sampling $k$ responses for each prompt, referred to as average-at-$k$.







\paragraph{Settings.}
The student model is taken to be \llamaS{}-base, \olmoS{}-base or \gemmaS{}-base on \gsm{} \citep{cobbe2021gsm8k}, and \llamaM{}-base on \math{} \citep{hendrycks2021MATH}.
We compare 15 text-only trained teachers:
LLaMA-(3.2/3.3) 3/8/70B Instruct models, Qwen-2.5 1.5/3/7/14B Instruct models, Qwen-2.5 Math 1.5/7B Instruct models, Gemma-2 2/9/27B Instruct models, OLMo 7/13B Instruct models, and Phi-4 on both \math{} and \gsm{} \citep{dubey2024llama,abdin2024phi4,yang2024qwen2,qwen25technicalreport,gemmateam2024gemma2improvingopen}.\footnote{We use short chain-of-thought reasoning models as teachers, given the extensive number of such available models, which also enables extensive ablation studies. Moreover, our experiments focus on small-scale students, for which prior work has shown that long chain-of-thought teachers do not necessarily benefit such models~\citep{li2025small}. 
}
The teacher's generation temperature is varied from $0.3$ to $1.0$ in increments of $0.1$. 
The computational costs of \gradcv{} are detailed in \Cref{app:compute}. We also compare against several common baselines, described in detail in \cref{app:other_metrics}.
All baselines considered do not require access to a verifier except for teacher performance.

\begin{figure}[tbp]
    \centering
    \includegraphics[width=\linewidth]{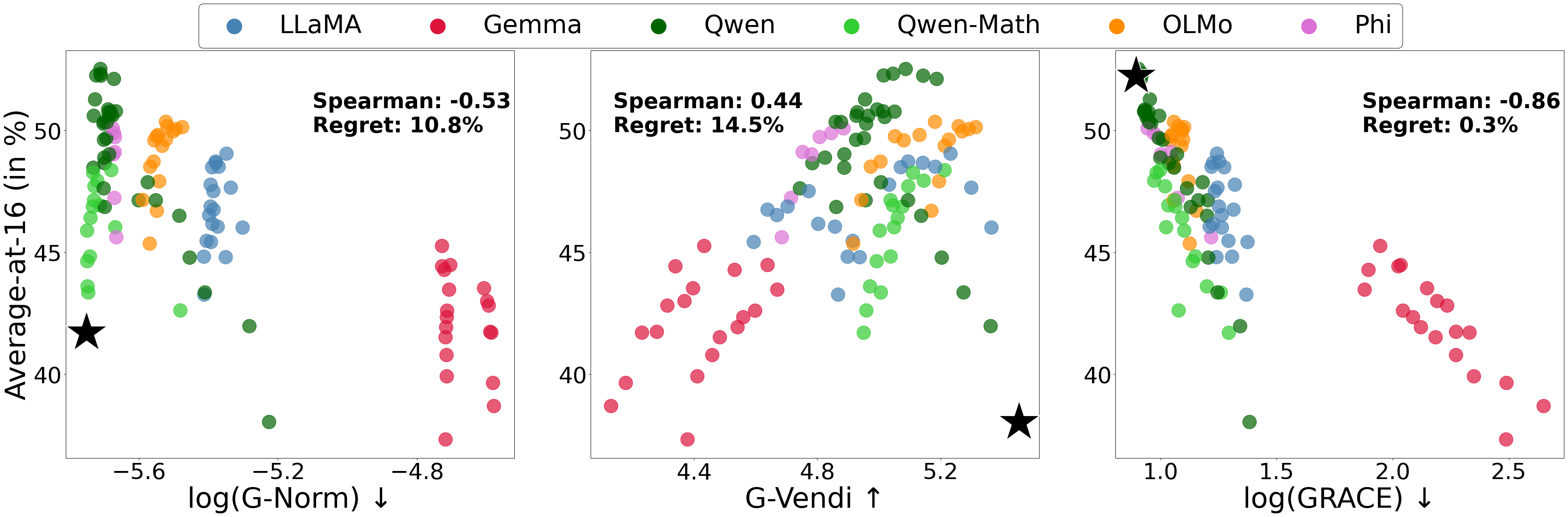}
    \caption{ \textbf{\gradcv{} achieves $86\%$ Spearman correlation to \llamaS{}'s post-distillation performance on \gsm{}}, much higher than \gnorm{} (53\%) and \gvendi{} (44\%).
    When evaluated by teacher-student regret, \gradcv{} selects a teacher with regret of $0.3\%$, outperforming \gnorm{} and \gvendi{}, which incur regrets of $10.8\%$ and $14.5\%$, respectively.
     Stars denote students trained from the teacher chosen by each score. Gemma teachers are outliers, because they give extremely concise responses to each prompt. More discussion is in \cref{app:response_analysis}
    }
    \label{fig:across_all_temps}
\end{figure}

To compute scores, we use a subset of $\numprompts=512$ prompts randomly selected from the training set, with $\numgenerations=4$ generations per prompt.
For $\gradcv{}$, we use $10$-way cross validation (i.e. $\numCV =10$).
The student gradients are randomly projected to dimension $\gradientdim = \numprompts = 512$; we provide ablation results on these hyperparameter choices in \Cref{sec:ablation}.

Each distillation training run uses learning rate\footnote{We search over learning rates $\{5 \times 10^{-5}, 10^{-5}, 5\times10^{-6}\}$ and find $10^{-5}$ to be consistently the best.
} $10^{-5}$ and 4 epochs over the training set.
We use the cosine learning rate schedule with $5\%$ warmup, 0 weight decay, and batch size $64$.
We generate $\Numgenerations=16$ responses per prompt from each teacher and fine-tune the student on all generations without filtering for correctness of the final answer.\footnote{Surprisingly, our ablations in \cref{sec:filter_v_nofilter} show that our results are not significantly affected if we filter by correctness.}

\textbf{Evaluation.} We evaluate the utility of each score with two tests. First, we compute the Spearman correlation of each score with the performance of the trained student models. Second, we measure the \emph{teacher-student regret}, defined as the absolute difference between the final student performance when trained by the best available teacher and the final student performance when trained under the best teacher identified by each score.

We measure performance of the student by average-at-16 performance based on $16$ generations at temperature $1.0$. Average-at-16 is a more stringent evaluation metric that tests consistency of generation from the model by requiring to solve a problem correctly across multiple attempts. It has recently been used for assessing robust performance on olympiad-level benchmarks and also serves as a key loss component when training models via reinforcement learning \citep{guo2025deepseek}.
We discuss later in \cref{sec:ablation} how the results change when we look at other performance metrics.

\begin{figure}[tbp]
    \centering
    \includegraphics[width=\linewidth]{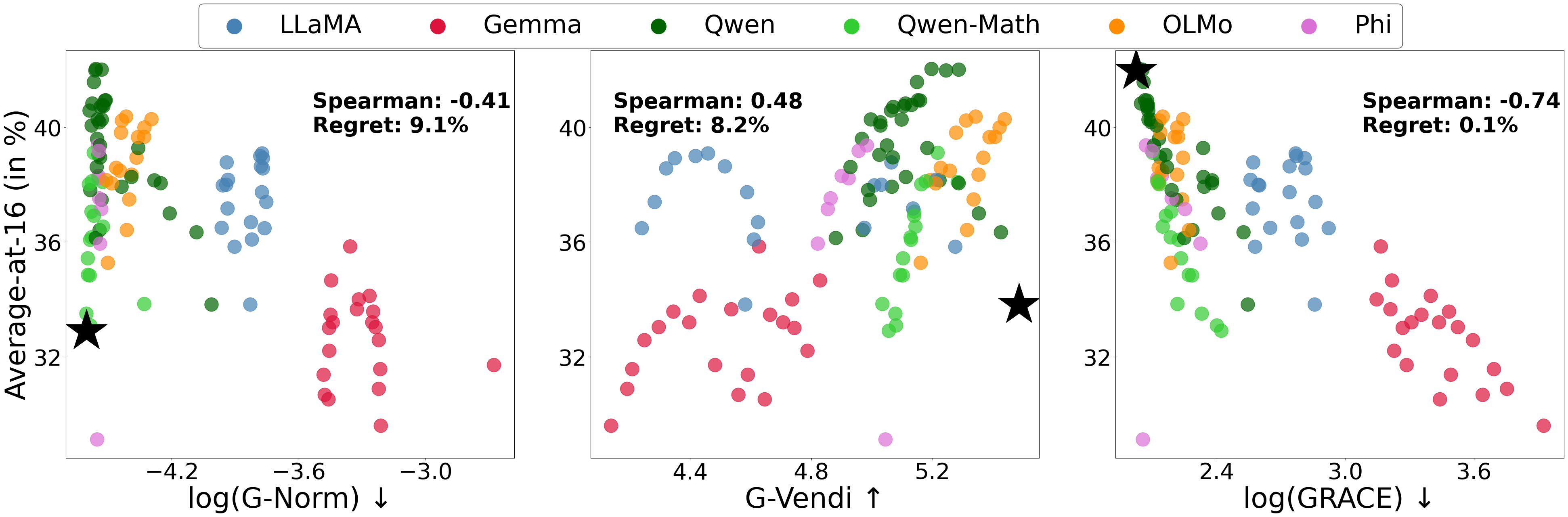}
    \caption{ 
    \textbf{\textbf{\gradcv{} achieves $74\%$ Spearman correlation to \olmoS{}'s post-distillation performance on \gsm{}}},
    significantly outperforming \gnorm{} (41\%) and \gvendi{} (48\%). 
    When evaluated by teacher-student regret, \gradcv{} selects a teacher with regret of $0.1\%$, outperforming \gnorm{} and \gvendi{}, which incur regrets of $9.1\%$ and $8.2\%$, respectively.
    Stars denote students trained from the teacher chosen by each score.
    Similar observations hold for a \gemmaS{} student (\cref{fig:across_all_temps_gsm_gemma}). 
    } 
    \label{fig:across_all_temps_olmo}
\end{figure}

\textbf{Scatter plots.} We present our findings in form of scatter plots (e.g. \cref{fig:across_all_temps,fig:across_all_temps_olmo}). In all scatter plots, each scatter point represents the student’s distillation performance from a specific teacher–temperature pair. Different teacher families are shown in distinct colors for clarity. We mark best teacher selected by each score by ``star'' in its corresponding subplot.


\subsection{\gradcv{} correlates well with student's performance and returns low regret}
\looseness-1\cref{fig:across_all_temps} shows that for a \llamaS{} model trained on \gsm{}, \gradcv{} achieves the best Spearman correlation with the student performance (0.86) when compared against \gnorm{} (0.55) and \gvendi{} (0.44). Furthermore, \gradcv{} returns a lower regret (0.3 $\%$) when selecting the best teacher, compared to \gnorm{} (4.9 $\%$) and \gvendi{} ($14.5\%$) respectively.

\begin{figure}[h]
    \centering
    \includegraphics[width=\linewidth]{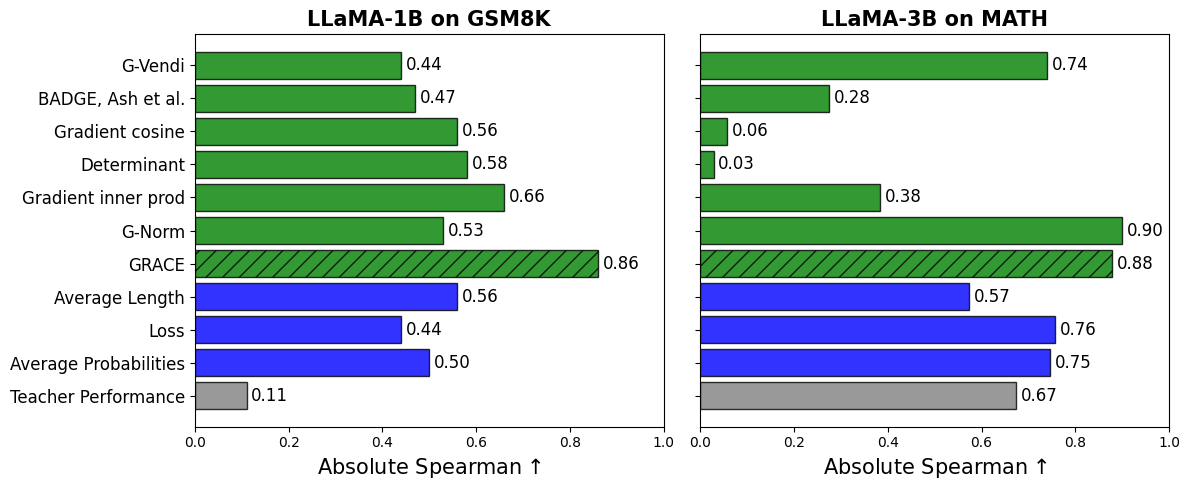}
    \caption{\looseness-1\textbf{\gradcv{} achieves the strongest correlation to student performance},
    among all scores for \llamaS{} on \gsm{} and \llamaM{} on \math{}.
    Blue bars represent gradient-based scores, green bars denote student logit-based scores on the training data, and the gray bar corresponds to teacher performance. 
    Teacher performance and the student's loss on teacher generations (Loss) before training show only weak correlations. 
    While \gnorm{} correlates well with student performance on \math{}, it is significantly worse on \gsm{}.}
    \label{fig:comp_to_baselines}
\end{figure}

Additional experiments with \olmoS{} and \gemmaS{} models trained on \gsm{} (\cref{fig:across_all_temps_olmo,fig:across_all_temps_gsm_gemma}) as well as with a \llamaM{} model trained on \math{} (\cref{fig:across_all_temps_math}) verify the utility of \gradcv{}. 
We also compare against other data selection baselines in \cref{fig:comp_to_baselines}. 
Across both \gsm{} and \math{}, \gradcv{} stands out as the only score maintaining a strong correlation ($>85\%$) with student performance, and simultaneously yields the lowest regret, $0.3\%$ on \gsm{} and $3.9\%$ on \math{}.


\textbf{Comparisons with teacher performance and student loss.}
Two intuitive baselines fail to reflect the student's distillation performance. The first is the teacher's own performance, which only shows a weak correlation of $11\%$ for \llamaS{} on \gsm{}. This agrees with findings in prior works~\citep{mirzadeh2019improved,harutyunyan23supervision,panigrahi2025progressive,zhang2023lifting,zhang2023towards,peng2024pretraining,razin2025makes}.
As an example, LLaMA-70B Instruct has the best performance among all teachers, but a student trained with LLaMA-70B Instruct reaches only $44.5\%$ average-at-16 performance, which returns a regret of $7.7\%$ when compared against the best performing student.
Similarly, the student's loss on teacher's generations, measured before training, is also poorly correlated with the student's  performance ($44\%$ with \llamaS{} training on \gsm{}) and returns a regret of $5.4\%$.


\textbf{Qualitative study: Teacher selection requires balancing directional coverage and variance.}
As a case study, we compare different teachers under a fixed generation temperature of $0.6$ (\Cref{fig:comparison_singletemp} in appendix). \gnorm{} clearly separates Qwen-Instruct from LLaMA-Instruct teachers but fails to distinguish between Qwen, Phi-Instruct, and Qwen-Math-Instruct, suggesting that a low gradient variance alone is insufficient to identify the best teacher.
Although \gvendi{} provides better separation among teachers with low \gnorm{}, it also assigns higher scores to sub-optimal teachers, indicating that directional coverage by itself is also inadequate.
In contrast, \gradcv{} achieves the strongest correlation ($92\%$) and correctly identifies Qwen-3B-Instruct as the optimal teacher. 

\subsection{Guiding distillation practice with \gradcv{}}
\label{sec:elucidate} 

\gradcv{} can go beyond identifying the best teacher and inform distillation practices.
Below we discuss how \gradcv{} provides guidance under common scenarios.

\begin{figure}[h]
    \centering
    \includegraphics[width=\linewidth]{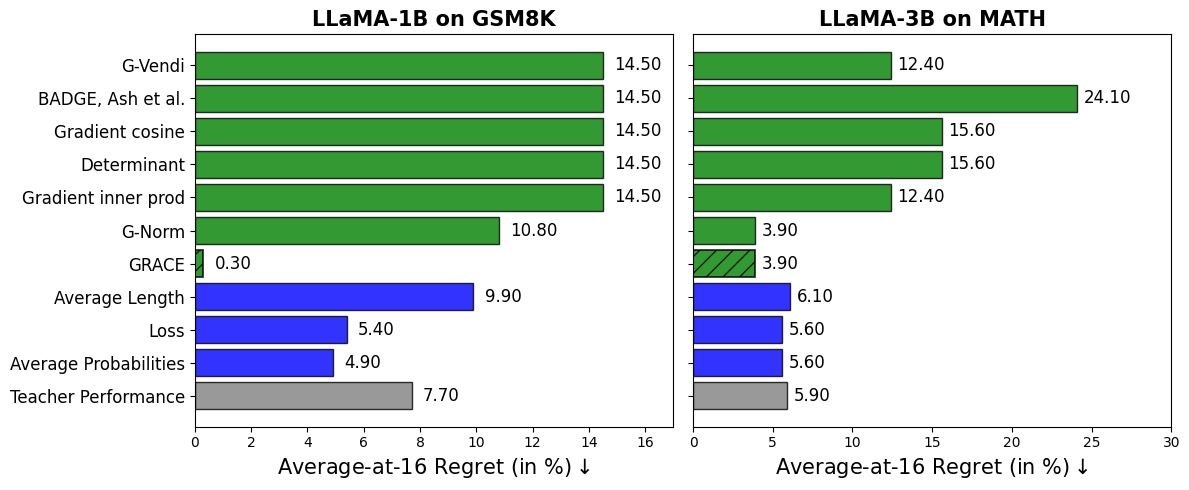}
    \caption{ \textbf{\gradcv{} achieves the minimum regret}, among all scores for \llamaS{} on \gsm{} and \llamaM{} on \math{}.
    Naively selecting the teacher with the best performance shows a regret of at least $7.7\%$ and $5.9\%$ in average-at-16 performance of student training on \gsm{} and \math{} respectively. On the other hand, \gradcv{} achieves a regret of $0.3\%$ and $3.9\%$ on \gsm{} and \math{} respectively. 
    In the left plot, scores that show $14.5$ regret  
     select the same teacher, resulting in identical regret values.
    }
    \label{fig:comp_to_baselines_regret}
\end{figure}

\begin{figure}[tb]
    \centering
    \includegraphics[width=\linewidth]{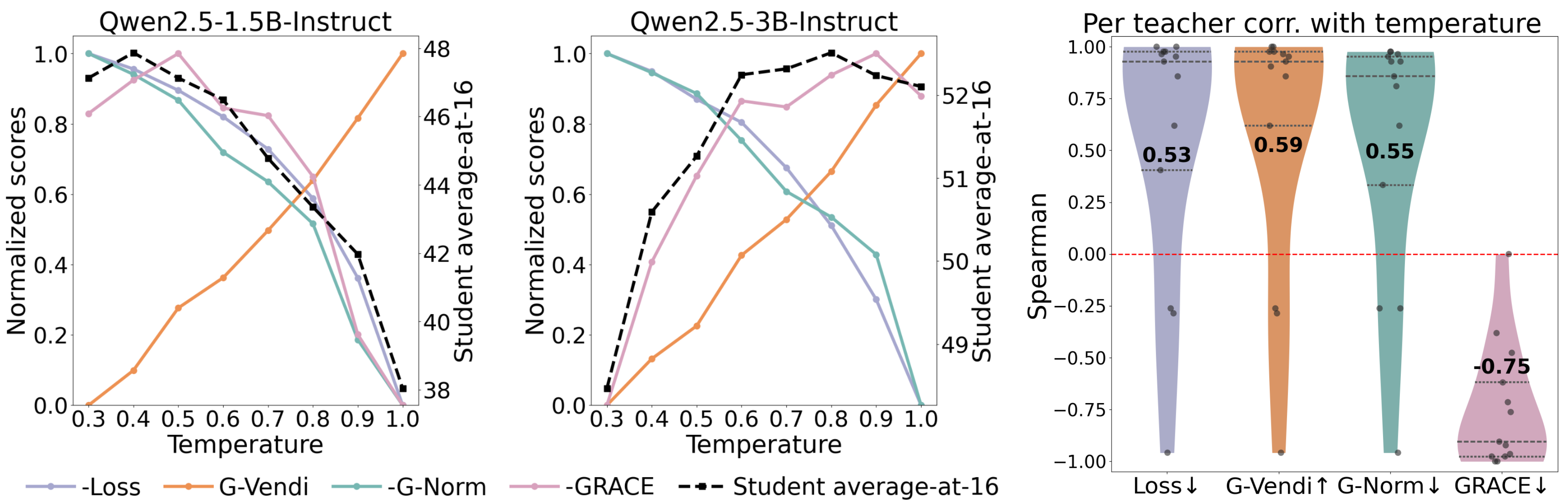}

    \caption{\textbf{\gradcv{} identifies a good teacher generation temperature.}
    (Left) Results are shown for \llamaS{} trained with Qwen-2.5-1.5B-Instruct and Qwen-2.5-3B-Instruct teachers on \gsm{}.
    \gradcv{} correctly identifies that (1) a lower temperature is optimal for Qwen-2.5-1.5B-Instruct, and (2) a higher temperature is effective for Qwen-2.5-3B-Instruct.
    In contrast, \gnorm{} can only identify (1) and \gvendi{} can only identify (2).  For clarity, all scores are normalized to  [0, 1]. The signs of Loss, \gnorm{}, and \gradcv{} are inverted so that all scores become higher-is-better for better visualization.  (Right) Average correlations of student performance with each score when the generation temperature is varied from a teacher. \gradcv{} achieves $75\%$ correlation, higher than \gvendi{} ($59\%$) and other scores.
    } 
    \label{fig:gsm8k-comparative}
\end{figure}

\textbf{Selecting generation temperature.}
The temperature $\tau$ used to rescale the teacher's logits when generating responses is known to have a strong influence on student performance after distillation~\citep{zheng2024knowledge,peng2024pretraining}.
However, there hasn't been a principled approach to choose the temperature.
We show in \Cref{fig:gsm8k-comparative} that \gradcv{} can identify such a good generation temperature for two Qwen teachers: it closely predicts the optimal generation temperature for \llamaS{} training, which are $0.8$ (vs. predicted $0.9$) with the 3B teacher and $0.4$ (vs. predicted \ $0.5$) with the 1.5B teacher.
In comparison, \gnorm{} and \gvendi{} tend to increase monotonically with the temperature, even though the student's performance shows an inverse U-shape in temperature.
Aggregating across all teachers in \cref{fig:gsm8k-comparative} (right), \gradcv{} achieves a 75\% correlation with student performance as temperature varies, substantially outperforming \gvendi{} (59\%). Interestingly, \gnorm{} is negatively correlated ($-53\%$) with performance in this setting.

\begin{figure}
    \centering
    \begin{minipage}[t]{0.48\linewidth}
        \includegraphics[width=\linewidth]{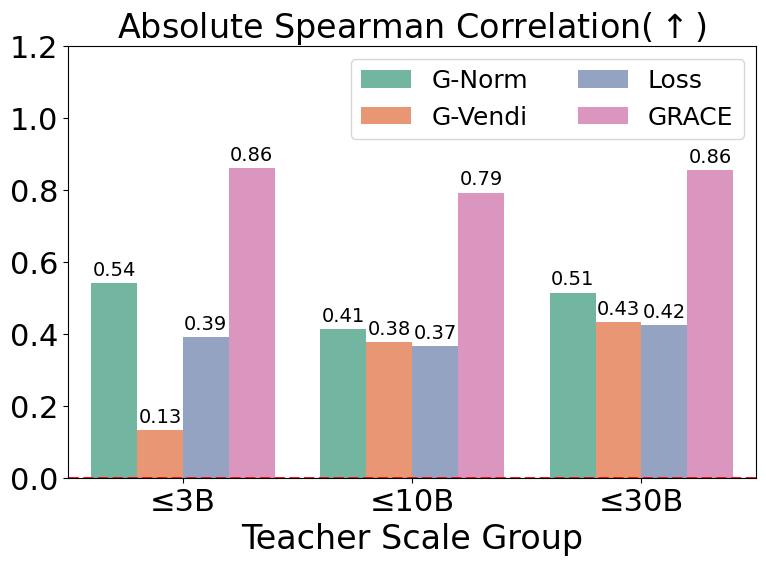}
    \end{minipage}%
    \begin{minipage}[t]{0.45\linewidth}
        \includegraphics[width=\linewidth]{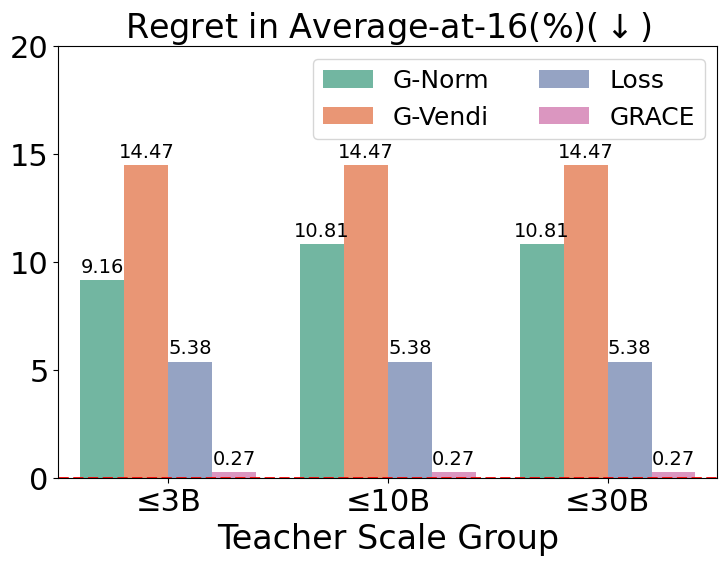}
    \end{minipage}
    %
    \caption{\textbf{\gradcv{} effectively predicts student performance under scale-constrained teachers.} 
    Results are for \llamaS{} on \gsm{}.
    (Left) Across all teacher scale constraints, \gradcv{} has high correlation (at least $79\%$) with the student performance after training with the teacher.
    (Right) Across all teacher scale constraints, \gradcv{}-selected teacher has minimum regret ($<1\%$) when compared against other scores.
    In the right plot, \gvendi{} and Loss consistently select the same teacher across all scale constraints, resulting in identical regret values in each case.
    }
    \label{fig:comparison_Across_scale_and_temperature}
\end{figure}


\textbf{Selecting a teacher within a size budget}.
In practice, one common resource constraint for distillation is the compute required to locally host open-source teachers.
Motivated by this, we test whether \gradcv{} can be used to select a teacher under a given size.
Specifically, we evaluate three scale constraints: (1) $3$B and below, (2) $10$B and below, and (3) $30$B and below.
\Cref{fig:comparison_Across_scale_and_temperature} shows that \gradcv{} achieves over $79\%$ correlation with student performance across all scale constraints. Moreover, within each scale constraint, using \gradcv{} to select a teacher yields a regret below $0.3\%$, compared to at least $9\%$ regret with \gnorm{} and \gvendi{}.




\looseness-1\textbf{Selecting teachers within a model family.}
Another practical limitation is the family of models that one can access, motivating us to test \gradcv{} against models within each model family.
We split the teacher models by model family and consider all generation temperatures.
Since some families include only a small number of teachers, the Spearman correlations can be unreliable. We hence report the regret of \gradcv{} for each teacher family.
As shown in \cref{fig:family_level_gaps} in appendix, \gradcv{} achieves an average regret of just $1\%$ when selecting a teacher within each family, while other metrics yield at least $3\%$ regret on average. Interestingly, the best teacher is not always from the same family as the student. For instance, a \llamaS{} student learns better from a Qwen-Instruct teacher than from any LLaMA-Instruct teacher.

\subsection{Ablations and additional experiments}
\label{sec:ablation}

\textbf{Effect of hyperparameters.} We test the effect of various hyperparameters used in the \gradcv{} computation.
We vary the number of prompts ($\numprompts$), the number of generations per prompt ($\numgenerations$), and the dimension of the gradient random projection ($\gradientdim$).
For the \llamaS{} student on \gsm{}, we find that \gradcv{} is generally robust to these hyperparameter choices, and the default values ($\numgenerations = \gradientdim = 512$, $\numgenerations=4$) work well (see details in \Cref{app:ablation_hyperparams}).
We also vary the number of cross-validation splits used in \gradcv{}.
For both \gsm{} and \math{}, the correlation with student performance remains fairly stable once $\numCV >= 6$ (\Cref{fig:varyC}), so we set $\numCV=10$ for our experiments.

\textbf{Robustness of the scores.} To test the robustness with respect to teacher selection, we evaluate correlations on random subsets of teachers.
In addition to the case studies in \Cref{sec:elucidate}, we repeatedly compute scores over random subsets of teachers.
As shown in \Cref{fig:ablation_robust_gsm},
\gradcv{} consistently maintains high correlations across these subsets (see details in \Cref{app:ablation_robust}).

\looseness-1\textbf{Changing Average-at-k to other performance measures.} 
We further examine how correlations change when replacing Average-at-k with other evaluation metrics. For \gsm{}, we find that Spearman correlation drops when switching from Average-at-k to either greedy or best-of-k accuracy, but \gradcv{} still returns the lowest regret in the corresponding performance measure ($<2.5\%$) when selecting the best teacher(\cref{tab:greedy_and_bestof16_gradient_temp} in appendix). Greedy reflects performance from a single generation at temperature $0.0$, and best-of-k measures whether the student answers correctly at least once over $k$ responses at generation temperature $1.0$.
A deeper investigation into the discrepancy between Average-at-k and these discrete performance metrics is left to future work.

\looseness-1\textbf{Out-of-distribution evaluation:} We evaluate \llamaS{} trained on \gsm{} on GSM-Symbolic \citep{mirzadeh2024gsm} (\cref{fig:gsm_symbolic_llama}), and \llamaM{} trained on \math{} on \math{}$^2$ \citep{shah2024ai} and \math{} Perturb \citep{huang2025math} (\cref{tab:math_perturb_llama}). On \gsm{}, student performance is strongly correlated between \gsm{} and GSM-Symbolic, and \gradcv{} continues to be highly predictive under this OOD shift.

\looseness-1For \math{}, we observe a nuanced pattern. On the simpler OOD task \math{} Perturb (Simple), \gnorm{} is a stronger predictor of student performance across teachers than \gradcv{}. However, on the more challenging OOD settings, \math{}$^2$ and \math{} Perturb (Hard), \gradcv{} achieves similar or higher spearman correlation and lower regret than \gnorm{}, indicating that \gradcv{} is a reliable score for identifying teachers for which the trained student generalizes well on difficult OOD tasks.

\section{Related work}
\paragraph{Knowledge distillation.} 
Knowledge distillation is a classic method used to improve the optimization and generalization of a small model~\citep{hinton2015distilling}.
A counterintuitive finding is that a better-performing model is not necessarily a better teacher,
which has been observed in both classic classification or regression settings~\citep{mirzadeh2019improved,jafari2021annealing,harutyunyan23supervision} and more recently in language models~\citep{zhang2023towards,zhang2023lifting,xu2024stronger,panigrahi2025progressive}.
For language models, one can distill from either the logits of the teacher or the generated texts.
\footnote{We consider generations following standard next-token distributions, as opposed to antidistillation sampling~\citep{savani2025antidistillation}.}
While the former can lead to better student performance, it is more computationally costly, requires higher access, and is less flexible due to tokenizer choices.
We hence focus on distilling from generated texts~\citep{eldan2023tinystories,li2023textbooks,busbridge2025distillation}. Recent work by \cite{guha2025openthoughts} supports our findings: they demonstrate that a weaker teacher can yield a stronger distilled model, that distillation benefits from increased sample size, and that filtering has little impact on the resulting student’s performance. 





\paragraph{Data selection.}
For text-based distillation, selecting the best teacher can be considered as the problem of choosing the most useful subset of samples from the generations of all teachers.
This aligns with the broad task of \textit{data selection}, which aims to identify subsets of data that maximize certain utility~\citep{sorscher22beyond,albalak24select}.
Many approaches leverage gradient information~\citep{mirzasoleiman2019coresets,killamsetty2020glister,pruthi2020estimating,xia24less},
including some that directly rely on notions of coverage~\citep{ash2019deep,jung2025prismatic}.
Directional coverage also ties to the notion of coverage in reinforcement learning.
Specifically, autoregressive training on teacher generations can be viewed as a form of behavior cloning, for which increasing the coverage is provably beneficial~\citep{song2024importance,huang2025bon,rohatgi2025computational}.
Despite these similarities, distillation differs from standard data selection in that it allows generating new data and offers a richer design space~\citep{peng2024pretraining}. An effective teacher-selection score should therefore be versatile and broadly applicable across scenarios, a property that \gradcv{} demonstrates as shown in \Cref{sec:elucidate}.

\citet{zhang2025best} study teacher selection at the granularity of individual prompts, proposing to score a pool of responses from different teachers by the student’s loss and train on the best-scoring completion for each prompt. However, as also observed by the contemporary work \citep{just2025distilling} and confirmed in our experiments, such loss-based scoring is only a coarse proxy for response quality. \citet{just2025distilling} propose using loss on short reasoning segments to mitigate this. They show that such scoring can be used to select the best global teacher, however their evaluation is limited to only three teachers. 
By contrast, we evaluate \gradcv{} across fifteen teachers and eight temperature settings each. 
Extending \gradcv{} to per-prompt teacher selection is a promising direction for future work.


\section{Discussion}
\label{sec:conclusion}

This work proposes the \gradcv{} score for identifying the most suitable distillation teacher.
Motivated from optimization and connected to CMI,
\gradcv{} leverages two distributional properties of the student's gradients: the directional coverage of the (normalized) gradients, and the gradient variance.
Experiments on \gsm{} and \math{} establish that \gradcv{} strongly correlates with the student's performance after distillation,
and enables principled comparison across teachers and offers actionable insights into practical scenarios.
Our results highlight \gradcv{}'s potential as a practical and general-purpose tool for guiding distillation practices.

Below we discuss promising avenues for future work.

First, \gradcv{} is not guaranteed to be robust in adversarial setups.
One example is when a teacher repeatedly emits partial reasoning steps or redundant text before eventually arriving at the correct answer.
Due to the in-context ability of language models, growing the number of repetitions will make the prediction closer to deterministic, and hence having an increasingly small gradient.
In this case, the \gradcv{} score may be deceptively low even though the student can perform arbitrarily poorly.
Fortunately, such behaviors are rare among well-performing teachers and can be mitigated via post-processing.
Fully characterizing the failure cases of \gradcv{} remains an open question.

Moreover, 
while \gradcv{} already captures two important distributional properties of the student's gradients, 
incorporating other properties of the teacher and characteristics of the data can be fruitful.
Although \gradcv{}'s design intentionally avoids requiring teacher logits, selectively incorporating logit-level information where available may lead to further performance gains.
It will also be interesting to investigate \gradcv{}'s utility in adaptive distillation strategies, where teacher choice may vary dynamically across training stages or subsets of data, possibly in an online fashion akin to reinforcement learning.


Finally, while this work focuses on mathematical reasoning due to the abundance of teacher choices and the ease of verification, \gradcv{} is not defined specifically for math. Extending \gradcv{} to more general domains would be an interesting direction for future work.
As a preliminary study, we extend our analysis to the ARC dataset~\citep{allenai:arc} in \cref{tab:values_on_arc}.
\gradcv{} achieves the highest Spearman correlation with student performance among all scores that do not rely on a verifier, and achieves the lower regret among all scores.
Interestingly on this task, we observe cases where teachers with extremely high \gradcv{} scores can still yield strong students.
Understanding the broader conditions under which \gradcv{} reliably predicts student performance is a promising future direction.



\subsection*{Acknowledgements}

We thank Jordan Ash, Alex Damian, Tianyu Gao, Jonathan Geuter, Sara Kangaslahti, Nihal Nayak, Xingyu Zhu, Simon Park, and Noam Razin for helpful discussions and suggestions.
Abhishek Panigrahi acknowledges support from Apple AIML and Siebel Scholarships.
This work was enabled in part by a gift from the Chan Zuckerberg Initiative Foundation to establish the Kempner Institute for the Study of Natural and Artificial Intelligence.

\bibliography{iclr2026_conference}

@article{panigrahi2025progressive,
  title     = {Progressive distillation induces an implicit curriculum},
  author    = {Abhishek Panigrahi and Bing Liu and Sadhika Malladi and Andrej Risteski and Surbhi Goel},
  journal   = {International Conference on Learning Representations},
  year      = {2025},
  bibSource = {Semantic Scholar https://www.semanticscholar.org/paper/61362e5da0c097209a86098e8b5fec1719bbaab3}
}

@article{xu2024stronger,
  title   = {Stronger Models are NOT Stronger Teachers for Instruction Tuning},
  author  = {Zhangchen Xu and Fengqing Jiang and Luyao Niu and Bill Yuchen Lin and Radha Poovendran},
  year    = {2024},
  journal = {arXiv preprint arXiv: 2411.07133}
}

@article{peng2024pretraining,
  title   = {Pre-training Distillation for Large Language Models: A Design Space Exploration},
  author  = {Hao Peng and Xin Lv and Yushi Bai and Zijun Yao and Jiajie Zhang and Lei Hou and Juanzi Li},
  year    = {2024},
  journal = {arXiv preprint arXiv: 2410.16215}
}

@article{zheng2024knowledge,
  title   = {Knowledge Distillation Based on Transformed Teacher Matching},
  author  = {Kaixiang Zheng and En-Hui Yang},
  year    = {2024},
  journal = {arXiv preprint arXiv: 2402.11148}
}

@article{zhang2023lifting,
  title     = {Lifting the Curse of Capacity Gap in Distilling Language Models},
  author    = {Chen Zhang and Yang Yang and Jiahao Liu and Jingang Wang and Yunsen Xian and Benyou Wang and Dawei Song},
  journal   = {Annual Meeting of the Association for Computational Linguistics},
  year      = {2023},
  doi       = {10.48550/arXiv.2305.12129},
  bibSource = {Semantic Scholar https://www.semanticscholar.org/paper/b2ec81b572fd5f0a5f5de843e3c62985b7d9c5a1}
}

@article{zhang2023towards,
  title   = {Towards the Law of Capacity Gap in Distilling Language Models},
  author  = {Chen Zhang and Dawei Song and Zheyu Ye and Yan Gao},
  year    = {2023},
  journal = {arXiv preprint arXiv: 2311.07052}
}

@article{mirzadeh2019improved,
  title     = {Improved Knowledge Distillation via Teacher Assistant},
  author    = {Seyed Iman Mirzadeh and Mehrdad Farajtabar and Ang Li and Nir Levine and Akihiro Matsukawa and H. Ghasemzadeh},
  journal   = {AAAI Conference on Artificial Intelligence},
  year      = {2019},
  doi       = {10.1609/AAAI.V34I04.5963},
  bibSource = {Semantic Scholar https://www.semanticscholar.org/paper/bc6dfc6bda2d929fec91042dce1831fd07999b39}
}

@article{jafari2021annealing,
  title   = {Annealing Knowledge Distillation},
  author  = {Aref Jafari and Mehdi Rezagholizadeh and Pranav Sharma and Ali Ghodsi},
  year    = {2021},
  journal = {arXiv preprint arXiv: 2104.07163}
}

@inproceedings{harutyunyan23supervision,
  author    = {Hrayr Harutyunyan and Ankit Singh Rawat and Aditya Krishna Menon and Seungyeon Kim and Sanjiv Kumar},
  title     = {Supervision Complexity and its Role in Knowledge Distillation},
  booktitle = {The Eleventh International Conference on Learning Representations, {ICLR} 2023, Kigali, Rwanda, May 1-5, 2023},
  publisher = {OpenReview.net},
  year      = {2023},
  url       = {https://openreview.net/forum?id=8jU7wy7N7mA},
  bibsource = {dblp computer science bibliography, https://dblp.org}
}

@article{cobbe2021gsm8k,
  title   = {Training Verifiers to Solve Math Word Problems},
  author  = {Karl Cobbe and Vineet Kosaraju and Mohammad Bavarian and Mark Chen and Heewoo Jun and Lukasz Kaiser and Matthias Plappert and Jerry Tworek and Jacob Hilton and Reiichiro Nakano and Christopher Hesse and John Schulman},
  year    = {2021},
  journal = {arXiv preprint arXiv: 2110.14168}
}

@article{hendrycks2021MATH,
  title     = {Measuring Mathematical Problem Solving With the MATH Dataset},
  author    = {Dan Hendrycks and Collin Burns and Saurav Kadavath and Akul Arora and Steven Basart and Eric Tang and D. Song and J. Steinhardt},
  journal   = {NeurIPS Datasets and Benchmarks},
  year      = {2021},
  bibSource = {Semantic Scholar https://www.semanticscholar.org/paper/57d1e7ac339e783898f2c3b1af55737cbeee9fc5}
}

@article{jung2025prismatic,
  title={Prismatic Synthesis: Gradient-based Data Diversification Boosts Generalization in LLM Reasoning},
  author={Jung, Jaehun and Han, Seungju and Lu, Ximing and Hallinan, Skyler and Acuna, David and Prabhumoye, Shrimai and Patwary, Mostafa and Shoeybi, Mohammad and Catanzaro, Bryan and Choi, Yejin},
  journal={arXiv preprint arXiv:2505.20161},
  year={2025}
}

@article{park2023trak,
  title={Trak: Attributing model behavior at scale},
  author={Park, Sung Min and Georgiev, Kristian and Ilyas, Andrew and Leclerc, Guillaume and Madry, Aleksander},
  journal={arXiv preprint arXiv:2303.14186},
  year={2023}
}

@article{razin2025makes,
  title   = {What Makes a Reward Model a Good Teacher? An Optimization Perspective},
  author  = {Noam Razin and Zixuan Wang and Hubert Strauss and Stanley Wei and Jason D. Lee and Sanjeev Arora},
  year    = {2025},
  journal = {arXiv preprint arXiv: 2503.15477}
}

@article{ash2019deep,
  title     = {Deep Batch Active Learning by Diverse, Uncertain Gradient Lower Bounds},
  author    = {J. Ash and Chicheng Zhang and A. Krishnamurthy and J. Langford and Alekh Agarwal},
  journal   = {International Conference on Learning Representations},
  year      = {2019},
  bibSource = {Semantic Scholar https://www.semanticscholar.org/paper/cf5a21684aefb1b8db6e0490167636d245396095}
}

@article{busbridge2025distillation,
  title   = {Distillation Scaling Laws},
  author  = {Dan Busbridge and Amitis Shidani and Floris Weers and Jason Ramapuram and Etai Littwin and Russ Webb},
  year    = {2025},
  journal = {arXiv preprint arXiv: 2502.08606}
}

@article{huang2025bon,
  title   = {Is Best-of-N the Best of Them? Coverage, Scaling, and Optimality in Inference-Time Alignment},
  author  = {Audrey Huang and Adam Block and Qinghua Liu and Nan Jiang and Akshay Krishnamurthy and Dylan J. Foster},
  year    = {2025},
  journal = {arXiv preprint arXiv: 2503.21878}
}

@article{rammal2022leave1one1out,
  title   = {On Leave-One-Out Conditional Mutual Information For Generalization},
  author  = {Mohamad Rida Rammal and Alessandro Achille and Aditya Golatkar and Suhas Diggavi and Stefano Soatto},
  year    = {2022},
  journal = {arXiv preprint arXiv: 2207.00581}
}

@article{savani2025antidistillation,
  title   = {Antidistillation Sampling},
  author  = {Yash Savani and Asher Trockman and Zhili Feng and Avi Schwarzschild and Alexander Robey and Marc Finzi and J. Zico Kolter},
  year    = {2025},
  journal = {arXiv preprint arXiv: 2504.13146}
}

@article{song2024importance,
  title   = {The Importance of Online Data: Understanding Preference Fine-tuning via Coverage},
  author  = {Yuda Song and Gokul Swamy and Aarti Singh and J. Andrew Bagnell and Wen Sun},
  year    = {2024},
  journal = {arXiv preprint arXiv: 2406.01462}
}

@article{eldan2023tinystories,
  title   = {TinyStories: How Small Can Language Models Be and Still Speak Coherent English?},
  author  = {Ronen Eldan and Yuanzhi Li},
  year    = {2023},
  journal = {arXiv preprint arXiv: 2305.07759}
}

@article{abdin2024phi4,
  title   = {Phi-4 Technical Report},
  author  = {Marah Abdin and Jyoti Aneja and Harkirat Behl and Sébastien Bubeck and Ronen Eldan and Suriya Gunasekar and Michael Harrison and Russell J. Hewett and Mojan Javaheripi and Piero Kauffmann and James R. Lee and Yin Tat Lee and Yuanzhi Li and Weishung Liu and Caio C. T. Mendes and Anh Nguyen and Eric Price and Gustavo de Rosa and Olli Saarikivi and Adil Salim and Shital Shah and Xin Wang and Rachel Ward and Yue Wu and Dingli Yu and Cyril Zhang and Yi Zhang},
  year    = {2024},
  journal = {arXiv preprint arXiv: 2412.08905}
}

@article{li2023textbooks,
  title   = {Textbooks Are All You Need II: phi-1.5 technical report},
  author  = {Yuanzhi Li and Sébastien Bubeck and Ronen Eldan and Allie Del Giorno and Suriya Gunasekar and Yin Tat Lee},
  year    = {2023},
  journal = {arXiv preprint arXiv: 2309.05463}
}

@inproceedings{sorscher22beyond,
  author    = {Ben Sorscher and Robert Geirhos and Shashank Shekhar and Surya Ganguli and Ari Morcos},
  editor    = {Sanmi Koyejo and S. Mohamed and A. Agarwal and Danielle Belgrave and K. Cho and A. Oh},
  title     = {Beyond neural scaling laws: beating power law scaling via data pruning},
  booktitle = {Advances in Neural Information Processing Systems 35: Annual Conference on Neural Information Processing Systems 2022, NeurIPS 2022, New Orleans, LA, USA, November 28 - December 9, 2022},
  year      = {2022},
  url       = {http://papers.nips.cc/paper\_files/paper/2022/hash/7b75da9b61eda40fa35453ee5d077df6-Abstract-Conference.html},
  bibsource = {dblp computer science bibliography, https://dblp.org}
}

@article{albalak24select,
  author    = {Alon Albalak and Yanai Elazar and Sang Michael Xie and Shayne Longpre and Nathan Lambert and Xinyi Wang and Niklas Muennighoff and Bairu Hou and Liangming Pan and Haewon Jeong and Colin Raffel and Shiyu Chang and Tatsunori Hashimoto and William Yang Wang},
  title     = {A Survey on Data Selection for Language Models},
  journal   = {Trans. Mach. Learn. Res.},
  volume    = {2024},
  year      = {2024},
  url       = {https://openreview.net/forum?id=XfHWcNTSHp},
  bibsource = {dblp computer science bibliography, https://dblp.org}
}

@article{hinton2015distilling,
  title   = {Distilling the Knowledge in a Neural Network},
  author  = {Geoffrey Hinton and Oriol Vinyals and Jeff Dean},
  year    = {2015},
  journal = {arXiv preprint arXiv: 1503.02531}
}

@article{rohatgi2025computational,
  title   = {Computational-Statistical Tradeoffs at the Next-Token Prediction Barrier: Autoregressive and Imitation Learning under Misspecification},
  author  = {Dhruv Rohatgi and Adam Block and Audrey Huang and Akshay Krishnamurthy and Dylan J. Foster},
  year    = {2025},
  journal = {arXiv preprint arXiv: 2502.12465}
}

@inproceedings{xia24less,
  author    = {Mengzhou Xia and Sadhika Malladi and Suchin Gururangan and Sanjeev Arora and Danqi Chen},
  title     = {{LESS:} Selecting Influential Data for Targeted Instruction Tuning},
  booktitle = {Forty-first International Conference on Machine Learning, {ICML} 2024, Vienna, Austria, July 21-27, 2024},
  publisher = {OpenReview.net},
  year      = {2024},
  url       = {https://openreview.net/forum?id=PG5fV50maR},
  bibsource = {dblp computer science bibliography, https://dblp.org}
}

@article{guha2025openthoughts,
  title   = {OpenThoughts: Data Recipes for Reasoning Models},
  author  = {Etash Guha and Ryan Marten and Sedrick Keh and Negin Raoof and Georgios Smyrnis and Hritik Bansal and Marianna Nezhurina and Jean Mercat and Trung Vu and Zayne Sprague and Ashima Suvarna and Benjamin Feuer and Liangyu Chen and Zaid Khan and Eric Frankel and Sachin Grover and Caroline Choi and Niklas Muennighoff and Shiye Su and Wanjia Zhao and John Yang and Shreyas Pimpalgaonkar and Kartik Sharma and Charlie Cheng-Jie Ji and Yichuan Deng and Sarah Pratt and Vivek Ramanujan and Jon Saad-Falcon and Jeffrey Li and Achal Dave and Alon Albalak and Kushal Arora and Blake Wulfe and Chinmay Hegde and Greg Durrett and Sewoong Oh and Mohit Bansal and Saadia Gabriel and Aditya Grover and Kai-Wei Chang and Vaishaal Shankar and Aaron Gokaslan and Mike A. Merrill and Tatsunori Hashimoto and Yejin Choi and Jenia Jitsev and Reinhard Heckel and Maheswaran Sathiamoorthy and Alexandros G. Dimakis and Ludwig Schmidt},
  year    = {2025},
  journal = {arXiv preprint arXiv: 2506.04178}
}

@article{team2024gemma,
  title   = {Gemma 2: Improving Open Language Models at a Practical Size},
  author  = {Gemma Team},
  year    = {2024},
  journal = {arXiv preprint arXiv: 2408.00118}
}

@article{mirzasoleiman2019coresets,
  title   = {Coresets for Data-efficient Training of Machine Learning Models},
  author  = {Baharan Mirzasoleiman and Jeff Bilmes and Jure Leskovec},
  year    = {2019},
  journal = {arXiv preprint arXiv: 1906.01827}
}

@article{li2025small,
  title     = {Small Models Struggle to Learn from Strong Reasoners},
  author    = {Yuetai Li and Xiang Yue and Zhangchen Xu and Fengqing Jiang and Luyao Niu and Bill Yuchen Lin and Bhaskar Ramasubramanian and Radha Poovendran},
  journal   = {Annual Meeting of the Association for Computational Linguistics},
  year      = {2025},
  doi       = {10.48550/arXiv.2502.12143},
  bibSource = {Semantic Scholar https://www.semanticscholar.org/paper/181b4a18a17105c6c270ec533b96b519c9e28c42}
}

@article{killamsetty2020glister,
  title   = {GLISTER: Generalization based Data Subset Selection for Efficient and Robust Learning},
  author  = {Krishnateja Killamsetty and Durga Sivasubramanian and Ganesh Ramakrishnan and Rishabh Iyer},
  year    = {2020},
  journal = {arXiv preprint arXiv: 2012.10630}
}

@article{steinke2020reasoning,
  title   = {Reasoning About Generalization via Conditional Mutual Information},
  author  = {Thomas Steinke and Lydia Zakynthinou},
  year    = {2020},
  journal = {arXiv preprint arXiv: 2001.09122}
}

@article{olmo20242,
  title   = {2 OLMo 2 Furious},
  author  = {Team OLMo},
  year    = {2024},
  journal = {arXiv preprint arXiv: 2501.00656}
}

@article{grattafiori2024llama,
  title   = {The Llama 3 Herd of Models},
  author  = {The Llama Team},
  year    = {2024},
  journal = {arXiv preprint arXiv: 2407.21783}
}

@article{duchi2011adaptive,
  title={Adaptive subgradient methods for online learning and stochastic optimization.},
  author={Duchi, John and Hazan, Elad and Singer, Yoram},
  journal={Journal of machine learning research},
  volume={12},
  number={7},
  year={2011}
}

@article{xu2017information,
  title={Information-theoretic analysis of generalization capability of learning algorithms},
  author={Xu, Aolin and Raginsky, Maxim},
  journal={Advances in neural information processing systems},
  volume={30},
  year={2017}
}

@article{loshchilov2017decoupled,
  title={Decoupled weight decay regularization},
  author={Loshchilov, Ilya and Hutter, Frank},
  journal={arXiv preprint arXiv:1711.05101},
  year={2017}
}

@article{kingma2014adam,
  title={Adam: A method for stochastic optimization},
  author={Kingma, Diederik P},
  journal={arXiv preprint arXiv:1412.6980},
  year={2014}
}

@article{pruthi2020estimating,
  title={Estimating training data influence by tracing gradient descent},
  author={Pruthi, Garima and Liu, Frederick and Kale, Satyen and Sundararajan, Mukund},
  journal={Advances in Neural Information Processing Systems},
  volume={33},
  pages={19920--19930},
  year={2020}
}

@article{dubey2024llama,
  title={The llama 3 herd of models},
  author={Dubey, Abhimanyu and Jauhri, Abhinav and Pandey, Abhinav and Kadian, Abhishek and Al-Dahle, Ahmad and Letman, Aiesha and Mathur, Akhil and Schelten, Alan and Yang, Amy and Fan, Angela and others},
  journal={arXiv e-prints},
  pages={arXiv--2407},
  year={2024}
}

@article{yang2024qwen2,
  title={Qwen2. 5-math technical report: Toward mathematical expert model via self-improvement},
  author={Yang, An and Zhang, Beichen and Hui, Binyuan and Gao, Bofei and Yu, Bowen and Li, Chengpeng and Liu, Dayiheng and Tu, Jianhong and Zhou, Jingren and Lin, Junyang and others},
  journal={arXiv preprint arXiv:2409.12122},
  year={2024}
}

@misc{qwen25technicalreport,
      title={Qwen2.5 Technical Report}, 
      author={Qwen and An Yang and Baosong Yang and Beichen Zhang and Binyuan Hui and Bo Zheng and Bowen Yu and Chengyuan Li and Dayiheng Liu and Fei Huang and Haoran Wei and Huan Lin and Jian Yang and Jianhong Tu and Jianwei Zhang and Jianxin Yang and Jiaxi Yang and Jingren Zhou and Junyang Lin and Kai Dang and Keming Lu and Keqin Bao and Kexin Yang and Le Yu and Mei Li and Mingfeng Xue and Pei Zhang and Qin Zhu and Rui Men and Runji Lin and Tianhao Li and Tianyi Tang and Tingyu Xia and Xingzhang Ren and Xuancheng Ren and Yang Fan and Yang Su and Yichang Zhang and Yu Wan and Yuqiong Liu and Zeyu Cui and Zhenru Zhang and Zihan Qiu},
      year={2025},
      eprint={2412.15115},
      archivePrefix={arXiv},
      primaryClass={cs.CL},
      url={https://arxiv.org/abs/2412.15115}, 
}

@misc{gemmateam2024gemma2improvingopen,
      title={Gemma 2: Improving Open Language Models at a Practical Size}, 
      author={Gemma Team},
      year={2024},
      eprint={2408.00118},
      archivePrefix={arXiv},
      primaryClass={cs.CL},
      url={https://arxiv.org/abs/2408.00118}, 
}

@article{ba2016layer,
  title={Layer normalization},
  author={Ba, Jimmy Lei and Kiros, Jamie Ryan and Hinton, Geoffrey E},
  journal={arXiv preprint arXiv:1607.06450},
  year={2016}
}

@inproceedings{li2022robust,
  title={Robust training of neural networks using scale invariant architectures},
  author={Li, Zhiyuan and Bhojanapalli, Srinadh and Zaheer, Manzil and Reddi, Sashank and Kumar, Sanjiv},
  booktitle={International Conference on Machine Learning},
  pages={12656--12684},
  year={2022},
  organization={PMLR}
}

@inproceedings{kwon2023efficient,
  title={Efficient Memory Management for Large Language Model Serving with PagedAttention},
  author={Woosuk Kwon and Zhuohan Li and Siyuan Zhuang and Ying Sheng and Lianmin Zheng and Cody Hao Yu and Joseph E. Gonzalez and Hao Zhang and Ion Stoica},
  booktitle={Proceedings of the ACM SIGOPS 29th Symposium on Operating Systems Principles},
  year={2023}
}

@article{zhang2025best,
  title={The best instruction-tuning data are those that fit},
  author={Zhang, Dylan and Dai, Qirun and Peng, Hao},
  journal={arXiv preprint arXiv:2502.04194},
  year={2025}
}

@article{arora2022exposure,
  title={Why exposure bias matters: An imitation learning perspective of error accumulation in language generation},
  author={Arora, Kushal and Asri, Layla El and Bahuleyan, Hareesh and Cheung, Jackie Chi Kit},
  journal={arXiv preprint arXiv:2204.01171},
  year={2022}
}

@article{fang2024wrong,
  title={What is Wrong with Perplexity for Long-context Language Modeling?},
  author={Fang, Lizhe and Wang, Yifei and Liu, Zhaoyang and Zhang, Chenheng and Jegelka, Stefanie and Gao, Jinyang and Ding, Bolin and Wang, Yisen},
  journal={arXiv preprint arXiv:2410.23771},
  year={2024}
}

@article{allenai:arc,
      author    = {Peter Clark  and Isaac Cowhey and Oren Etzioni and Tushar Khot and
                    Ashish Sabharwal and Carissa Schoenick and Oyvind Tafjord},
      title     = {Think you have Solved Question Answering? Try ARC, the AI2 Reasoning Challenge},
      journal   = {arXiv:1803.05457v1},
      year      = {2018},
}

@article{engstrom2024dsdm,
  title={Dsdm: Model-aware dataset selection with datamodels},
  author={Engstrom, Logan and Feldmann, Axel and Madry, Aleksander},
  journal={arXiv preprint arXiv:2401.12926},
  year={2024}
}

@article{just2025distilling,
  title={Distilling Reasoning into Student LLMs: Local Naturalness for Selecting Teacher Data},
  author={Just, Hoang Anh and Ko, Myeongseob and Jia, Ruoxi},
  journal={arXiv preprint arXiv:2510.03988},
  year={2025}
}

@misc{lozhkov2024fineweb-edu,
    author       = { Lozhkov, Anton and Ben Allal, Loubna and von Werra, Leandro and Wolf, Thomas },  
    title        = { FineWeb-Edu: the Finest Collection of Educational Content }, 
    year         = 2024,  
    url          = { https://huggingface.co/datasets/HuggingFaceFW/fineweb-edu },  
    doi          = { 10.57967/hf/2497 },
    publisher    = { Hugging Face }
}

@article{mirzadeh2024gsm,
  title={Gsm-symbolic: Understanding the limitations of mathematical reasoning in large language models},
  author={Mirzadeh, Iman and Alizadeh, Keivan and Shahrokhi, Hooman and Tuzel, Oncel and Bengio, Samy and Farajtabar, Mehrdad},
  journal={arXiv preprint arXiv:2410.05229},
  year={2024}
}

@article{shah2024ai,
  title={Ai-assisted generation of difficult math questions},
  author={Shah, Vedant and Yu, Dingli and Lyu, Kaifeng and Park, Simon and Yu, Jiatong and He, Yinghui and Ke, Nan Rosemary and Mozer, Michael and Bengio, Yoshua and Arora, Sanjeev and others},
  journal={arXiv preprint arXiv:2407.21009},
  year={2024}
}

@article{huang2025math,
  title={MATH-Perturb: Benchmarking LLMs' Math Reasoning Abilities against Hard Perturbations},
  author={Huang, Kaixuan and Guo, Jiacheng and Li, Zihao and Ji, Xiang and Ge, Jiawei and Li, Wenzhe and Guo, Yingqing and Cai, Tianle and Yuan, Hui and Wang, Runzhe and others},
  journal={arXiv preprint arXiv:2502.06453},
  year={2025}
}

@article{guo2025deepseek,
  title={Deepseek-r1: Incentivizing reasoning capability in llms via reinforcement learning},
  author={Guo, Daya and Yang, Dejian and Zhang, Haowei and Song, Junxiao and Zhang, Ruoyu and Xu, Runxin and Zhu, Qihao and Ma, Shirong and Wang, Peiyi and Bi, Xiao and others},
  journal={arXiv preprint arXiv:2501.12948},
  year={2025}
}
\bibliographystyle{colm2025_conference}

\newpage
\appendix
\tableofcontents


\section{Outline of the appendix}
\label{sec:outline}

Below, we outline the structure of the appendix.

\cref{sec:baselines} details the baseline methods against which we compare \gradcv{}. \cref{sec:gradient_influence} introduces gradient influence–based scores and explains why they are ineffective for identifying the best teacher. Additional scoring methods are discussed in \cref{app:other_metrics}.

\cref{app:theory} formally defines conditional mutual information (CMI), which provides the theoretical basis for \gradcv{}. \cref{sec:connect_GRACE_to_Loss} highlights key limitations of using CMI to explain the success of \gradcv{} and outlines open questions to bridge this gap.

\cref{app:add_expts} shows additional experimental results: additional models on \gsm{} and \math{} (\cref{app:add_results}); results within individual teacher families (\cref{app:family}); the effect of teacher performance on \gradcv{} (\cref{app:teacher_performance}); correlations for metrics beyond Average-at-16 (\cref{app:other_student_performance}); evaluation on non-math datasets (\cref{app:ARC}); and computational complexity analysis (\cref{app:compute}). Finally, \cref{sec:ablation} presents ablations studying key design choices in \gradcv{}.





\section{Discussion on the baselines}
\label{sec:baselines}

\subsection{Using gradient influence for teacher selection}
\label{sec:gradient_influence}


Existing data selection methods estimate the importance of each data point based on its influence on a validation set. Among these, gradient-based approaches define this influence through the relationship between the gradient of an individual training example and the gradients computed on the validation set.

A straightforward extension of these data selection strategies is to measure the quality of the gradient distribution on $\empD$ w.r.t. a heldout validation set $\empD'$. Adapting the gradient influence metric from \cite{pruthi2020estimating}, we measure the influence of the gradients from training set $\empD$ on the validation set $\empD'$. Mathematically, this is defined as 
$$
\sum_{(\prompt, \gen) \in \empD} \sum_{(\prompt', \gen') \in \empD'}  \frac{\cos(\vh(\prompt, \gen), \vh(\prompt', \gen'))}{|\empD| \cdot |\empD'|} 
$$
Following \cite{xia24less}, we can additionally improve this definition by taking Adam optimization into account. This requires an initial warmup training of the model on an independent copy of $\empD$;
more details are provided in \cref{app:LESS}.

\begin{wrapfigure}{r}{0.5\linewidth}
    \centering
    \includegraphics[width=\linewidth]{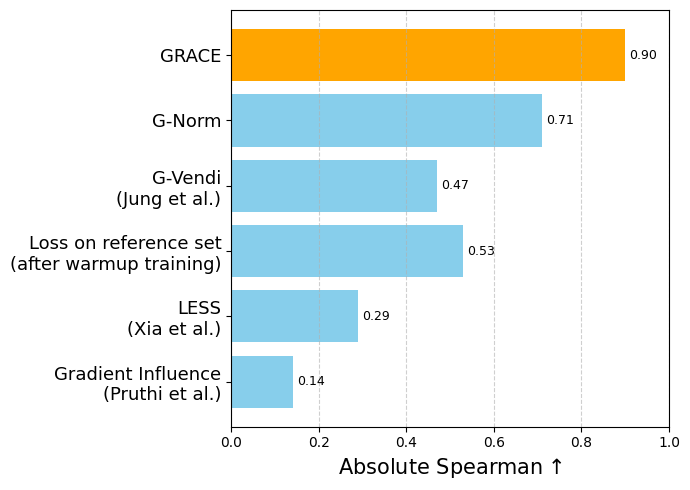}
    \caption{\looseness-1Influence-based measures that rely on a validation set are ineffective for selecting good teachers.
    We report \llamaS{} performance on \gsm{} across different teacher choices at a generation temperature of $0.6$.
    The validation split from \cite{cobbe2021gsm8k} serves as the held-out set for computing Gradient Influence and LESS \citep{pruthi2020estimating,xia24less}.}
    \label{fig:comparison_w_reference_based_baselines}
    \vspace{-30pt}
\end{wrapfigure}

We compare teachers for \llamaS{} trained on \gsm{}, with $\empD'$ being the validation split consisting of human-written solutions.
We also report the loss on the reference set $\empD'$ of the trained models after the warmup phase following \cite{xia24less}.
As shown in \cref{fig:comparison_w_reference_based_baselines}, influence-based metrics show low correlation with the  student performance after training.



\subsubsection{Details on LESS \citep{xia24less}} \label{app:LESS}

For discussion in this section, we will use a few additional notations: we will use $\Theta_{\cal S} \in \mathbb{R}^{D}$ to generally denote the parameters of the student model. $\Theta_{\cal S}^{(t)}$ will denote the parameters of the model at time $t$.

Suppose, we train our model with Adam  \citep{kingma2014adam} with hyperparameters $\eta, \beta_1, \beta_2, \epsilon$, that tracks two gradient statistics: $\mm_{t}$ and $\vv_{t}$ (initialized at 0s) at each step of training.
The gradient update step at any step $t > 0$ with a gradient vector $\mg \in \mathbb{R}^{D}$ is given by
\begin{align*}
    &\Theta_{\cal S}^{(t+1)} \gets \Theta_{\cal S}^{(t)} - \eta_t \Gamma_{t+1} (\mg) , \\&
    \text{where} \quad
    \Gamma_{t+1} (\mg) =  
    \mm_{t+1} \odot \frac{1}{\sqrt{\mv_{t+1}}  + \epsilon}, \\&
    \mm_{t+1} \gets \frac{1}{1-\beta_1^{t+1}} \left( (1-\beta_1) \mm_t + \beta_1 \mg \right),  \\&
    \vv_{t+1} \gets \frac{1}{1-\beta_2^{t+1}} \left( (1-\beta_2) \vv_t + \beta_2 \left(\mg \odot \mg\right) \right),
\end{align*}
where $\eta_t$ denotes the learning rate at time step $t$, that can be a function of $\eta$ and time-step $t$ depending on the learning rate schedule. Here $\Gamma$ denotes gradient pre-conditioning function that uses the gradient statistics $\mm$ and $\vv$.

LESS \citep{xia24less} considers the gradient update under Adam to compute the influence of the gradients from training dataset $\empD$ on validation set $\empD'$. Suppose in addition to $\empD$, we have a warmup dataset $\empD_{warmup}$. Then, the model is trained for $T$ steps ($4$ epochs) on the warmup dataset $\empD_{warmup}$. 

Let $\Gamma_{T/4}, \Gamma_{T/2}, \Gamma_{T/4}, $ and $\Gamma_{T}$ be the pre-conditioning functions at step $T/4$, $T/2$, $3T/4$, and $T$ respectively. Furthermore, let $\Bar{\eta}_{T/4}$ denote the average learning rate between timestep $0$ and $T/4$ (similarly, define  $\Bar{\eta}_{T/2}$ as the average learning rate between timestep $T/4$ and $T/2$, and so on).
Then, the influence of each input $(\prompt, \gen) \in \empD$, denoted by $\mathbf{Inf}(\prompt, \gen)$, is given by
\begin{align}
   \mathbf{Inf}(\prompt, \gen) := \frac{1}{4|\empD'|} \sum_{(\prompt', \gen') \in \empD'} \sum_{i=1}^{4} \Bar{\eta}_{iT/4} \text{cosine}  \left( \Gamma_{iT/4} (\vg (\prompt, \gen)),   \vg (\prompt', \gen') \right). \label{eq:LESS_influence}
\end{align}
For computational efficiency, we use a random projection $\Pi$:
\begin{align*}
   \mathbf{Inf}(\prompt, \gen) := \frac{1}{4|\empD'|} \sum_{(\prompt', \gen') \in \empD'} \sum_{i=1}^{4} \Bar{\eta}_{iT/4} \text{cosine}  \left( \Pi \Gamma_{iT/4} (\vg (\prompt, \gen)),  \Pi \vg (\prompt', \gen') \right). 
\end{align*}

In LESS, $\mathbf{Inf}(\prompt, \gen)$ was used to pick the most influential data points in $\empD$. We report the performance of LESS for teacher selection by using an average influence function on the dataset $\empD$ to measure its overall quality:
\begin{align*}
    \mathbf{Inf}(\empD) = \frac{1}{|\empD|} \sum_{(\prompt, \gen) \in \empD}  \mathbf{Inf}(\prompt, \gen).
\end{align*}

\subsection{Issues with influence based metrics for teacher selection} \label{app:first_order_estimate}

We hypothesize that a key limitation of these influence-based metrics is the choice of $\empD'$: 
it should both reflect a common ground truth and effectively differentiate models without bias toward any particular one, which can be challenging to construct.

Gradient influence metrics like LESS approximate the decrease in validation loss when taking a gradient optimization step. On an update $\Theta_{\cal S}^{(t+1)}$ from $\Theta_{\cal S}^{(t)}$ using a gradient vector $\vg$, with the corresponding student given by $\student^{(t+1)}$ and $\student^{(t)}$ respectively, we can show with a first order taylor expansion:

\begin{align}
        \E_{(\prompt', \gen') \in \empD'} \left( \loss ( \gen'|\prompt'; \student^{(t+1)})  -  \loss ( \gen' |\prompt'; \student^{(t)})  \right) &\propto \E_{(\prompt', \gen') \in \empD'} \left\langle \Theta_{\cal S}^{(t+1)} - \Theta_{\cal S}^{(t)} , \vg ( \gen', \prompt' ) \right\rangle \nonumber\\&
        = \E_{(\prompt', \gen') \in \empD'} \eta_t \left \langle \Gamma_{t+1} (\mg),  \vg ( \gen', \prompt' ) \right \rangle.
    \label{eq:influence}
\end{align}

However, there are two primary components in the solution $\gen'$: chain-of-thought and the final answer. If we denote each response $\gen'$ as composition of $(\text{cot}', a')$, then the above formulation can be re-formulated as
\begin{align*}
        \E_{(\prompt', \gen') \in \empD': \gen'=(\text{cot}', a')} \left( \loss ( \text{cot}', a'|\prompt'; \student^{(t+1)})  -  \loss ( \text{cot}', a' |\prompt'; \student^{(t)})  \right). 
\end{align*}

For a given prompt $\prompt'$, the final answer will be the same across different responses but the chain-of-thought $\text{cot}'$ can differ. 
For a pool of teachers $\{{\cal T}_1, \cdots, {\cal T}_p\}$ with training generations $\empD_{{\cal T}_i}$ for each teacher, the chain-of-thought patterns differ across the teachers.
\looseness-1Therefore, the validation set $\empD'$ should ideally be constructed such that the student's loss or gradient metrics evaluated on it can effectively distinguish between the diverse chain-of-thought behaviors across all the teachers’ training generations.

In \cref{fig:comparison_w_reference_based_baselines}, we show that using the validation set from the GSM8k dataset, which has human-written answers to each question, as $\empD'$ doesn't help effective teacher selection. In fact, the loss on the set $\empD'$ itself doesn't help to effectively distinguish between different teachers. We keep principled selection of a common validation set $\empD'$ for selecting the right teacher to future work.

\subsubsection{Can we use teacher generated validation sets for each teacher?}


An alternative is to use separate validation sets for different teachers, each containing responses generated by a specific teacher on a shared set of questions. That is, on a given validation set of prompts $\empD'$, we simply collect the response of a teacher ${\cal T}$ to each prompt $\prompt' \in \empD'$ and create validation set $\empD'_{{\cal T}}$. Then, for each teacher ${\cal T}$, we modify the LESS influence metric on an example $(\prompt, \gen) \in \empD_{{\cal T}}$ from \cref{eq:LESS_influence} as
\begin{align*}
        \mathbf{Inf}(\prompt, \gen) :=
        \frac{1}{4|\textcolor{red}{\empD'_{{\cal T}}}|}
        \sum_{(\prompt', \gen') \in \textcolor{red}{\empD'_{{\cal T}}}}
        \sum_{i=1}^{4} \Bar{\eta}_{iT/4}
        \text{cosine}\!\left(
          \Pi \Gamma_{iT/4}(\vg(\prompt, \gen)),
          \Pi \vg(\prompt', \gen')
        \right),
\end{align*}
where $\Pi$ denotes a random projection matrix.

\begin{figure}
    \centering
    \includegraphics[width=\linewidth]{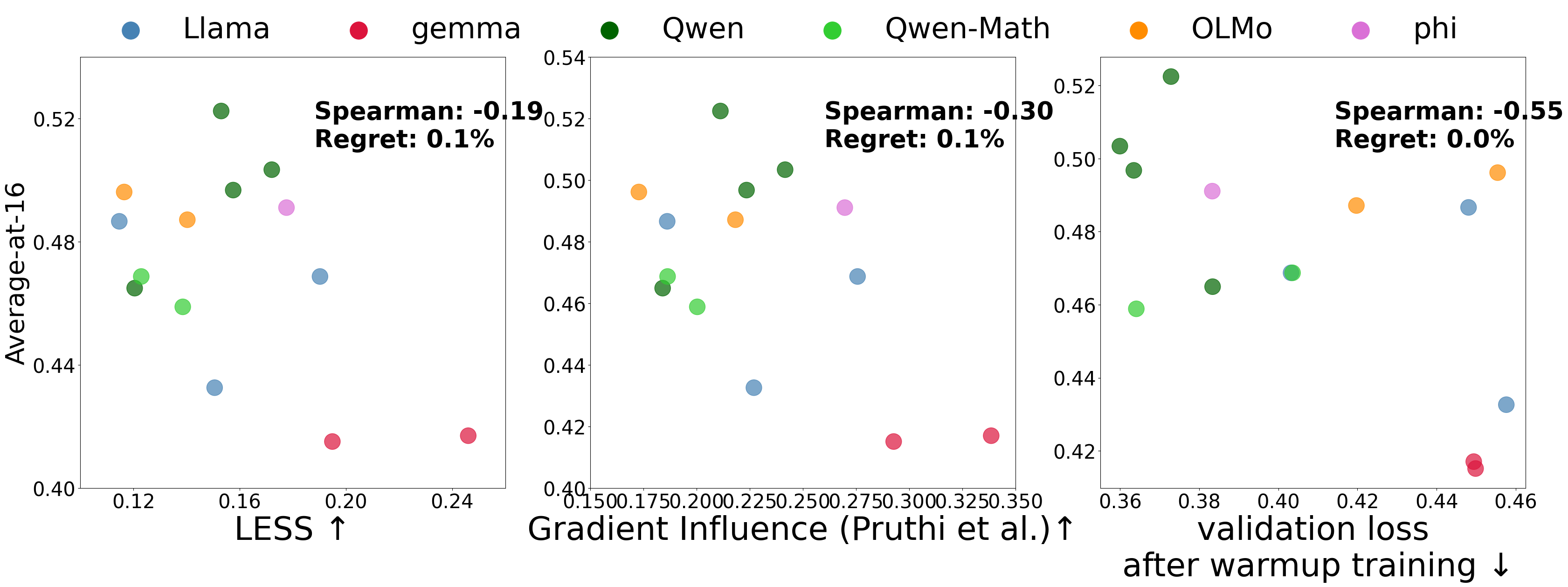}
    \caption{\looseness-1We repeat experiments from \cref{fig:comparison_w_reference_based_baselines}. But instead of using a common validation set with human written answers from \cite{cobbe2021gsm8k} to measure the gradient influence metrics across the teachers, we use a separate validation set for each teacher, where we use the respective teacher's response on the validation set of questions.
   We then report the performance of \llamaS{} on \gsm{} across different teacher choices at a generation temperature of $0.6$.
    We find that the teacher-selection performance of the influence-based metrics does not improve under this setup.
    For comparison, refer to \cref{fig:comparison_singletemp}, where gradient-based metrics achieve notably higher Spearman correlation with student performance.
    }
    \label{fig:comparison_w_reference_based_baselines_wteachereval}
\end{figure}

We report the performance of gradient influence metrics under this setup in \cref{fig:comparison_w_reference_based_baselines_wteachereval}. We find that the performance of the gradient influence metrics for teacher selection do not substantially change under this setup, compared to using a common validation set in \cref{fig:comparison_w_reference_based_baselines}.

One primary concern is that this approach breaks comparability across students trained on different teachers, as gradient estimates from different validation sets cannot be used to compare across them. That is, if $(\empD_{{\cal T}_1}, \empD'_{{\cal T}_1})$ and $(\empD_{{\cal T}_2}, \empD'_{{\cal T}_2})$ represent the training and validation sets from two teachers, influence estimate scores with $(\empD_{{\cal T}_1}, \empD'_{{\cal T}_1})$  and $(\empD_{{\cal T}_2}, \empD'_{{\cal T}_2})$ can't be used to directly differentiate between the generations of the teachers ${\cal T}_1$ and ${\cal T}_2$. 



\subsection{Descriptions of other baselines}
\label{app:other_metrics}

We consider the following additional baselines:
\begin{enumerate}
    \item \textbf{Student Loss on the teacher's generations:} This is defined by the average cross entropy loss of the student model on the teacher's generations before training.
    \item \textbf{\gnorm{}}: Defined in \Cref{eq:mean_grad_norm}, this measures the variance of the gradients w.r.t. the mean gradient.
    \item \textbf{\gvendi{}}: Defined in \Cref{eq:g-vendi}, this measures the directional entropy in the gradients.
    \item \textbf{Determinant}: This is defined by the determinant of the gradient matrix $\Tilde{\mG}(\empD)$ and serves as notion of diversity of the gradients.
    \item \textbf{Determinant $\times$ gradient norm}, corresponding to BADGE~\citep{ash2019deep}: This is defined by the determinant of the gradient matrix ${\mG}(\empD)$ and serves as notion of diversity of the gradients, that also takes their norm into  account.
    
    \item \textbf{Gradient cosine}, which is another way to capture gradient diversity: Given gradients from the training set $\empD$, we compute pairwise inner product between the normalized gradients of generations for the same prompt:
    \begin{align*}
        &\mathbb{E}_{\prompt} \mathbb{E}_{(\prompt, \gen_1), (\prompt, \gen_2) \sim \empD } \left[\frac{\mathbf{g_1}}{\norm{\mathbf{g}_1}}\right]^{\top} \frac{\mathbf{g_2}}{\norm{\mathbf{g}_2}},\\
        \text{where } &\mathbf{g}_1 = \mathbf{h} (\prompt, \gen_1), \\
        & \mathbf{g}_2 = \mathbf{h} (\prompt, \gen_2).
    \end{align*}
    
    \item \textbf{Gradient inner product}: we compute pairwise inner product between the gradients of generations from the same prompt.
    \begin{align*}
        &\mathbb{E}_{\prompt} \mathbb{E}_{(\prompt, \gen_1), (\prompt, \gen_2) \sim \empD } \text{ } \mathbf{g_1}^{\top} \mathbf{g_2},\\
        \text{where } &\mathbf{g}_1 = \mathbf{h} (\prompt, \gen_1), \\
        & \mathbf{g}_2 = \mathbf{h} (\prompt, \gen_2).
    \end{align*}
    This can be considered as considering gradient magnitude in addition to the gradient cosine.
    
    \item \textbf{Average Probabilities} (per token): this computes the average probability per token of the student on the teacher's generations, averaged over all generations and all prompts. 

    \item \textbf{Average length:} this computes the average length of the teacher's generations.
\end{enumerate}

As mentioned in \Cref{sec:expr}, naive metrics are not useful for identifying the best teachers.


\section{Connecting \gradcv{} to Leave-one-out Conditional Mutual Information (LOO-CMI)}\label{app:theory}

\subsection{Background on LOO-CMI}

We start with a general description of the LOO-CMI framework from \cite{steinke2020reasoning,rammal2022leave1one1out}. In this setup, we have a dataset $\empD = \{Z_1, Z_2, \cdots, Z_n\}$ of $\numprompts$ i.i.d. samples drawn from the parent distribution $\mathcal{P}$, where $Z_i$ refers to a general datapoint containing the data-label pair $(\prompt, \gen)$.

Consider a training algorithm $\mathcal{A}: \empD \to \Theta$ that takes a training set $\empD$ 
and returns a weight parameter $\Theta$.
Let $\idx$ be an index uniformly drawn from $\{1, \cdots, \numprompts\}$,
and denote by $\empD_{-\idx} := \empD \setminus \{Z_{\idx}\}$ the size-$(\numprompts-1)$ dataset with the $\idx$-th sample removed.
Denote the output of the training algorithm on $\empD_{-\idx}$ by $\Theta_{-\idx} = \mathcal{A}(\empD_{-\idx})$.
Then, the \emph{leave-one-out conditional mutual information (LOO-CMI)} of $\mathcal{A}$ is defined as
\begin{equation}
\begin{split}
    I_{\mathrm{LOO}}(\gA; \empD)
    :=& I(\Theta_{-U}; U \mid \empD)
    = \E_{\idx\sim\text{Unif}[\numprompts]} \left[\mathrm{KL}(P_{\Theta_{-u}|u,\empD} \| P_{\Theta_{-u} | \empD} )\right]
\end{split}
\label{eq:loo_cmi_def}
\end{equation}
where $I(\cdot;\cdot\mid\cdot)$ denotes the conditional mutual information, $\mathrm{KL}(\cdot\|\cdot)$ the Kullback-Leibler divergence and $P_{\Theta}$ denotes the distribution over parameters $\Theta$, where the randomness comes from gradient noise.
The LOO-CMI quantifies how much information the trained model parameters $\Theta$ reveal
about which sample index $U$ was removed when training on the remaining data.
A small value of $I_{\mathrm{LOO}}(\Theta; \empD)$ indicates that the algorithm’s output is insensitive to the removal of a single training example, which corresponds to higher algorithmic stability and better generalization.

\paragraph{Generalization bound.}
With a slight abuse of notation, let $\ell: \Theta \times \empD \to [0, 1]$ denote the loss on a dataset $\empD$,
and let $\ell: \Theta \to [0,1]$ denote the population loss, i.e. taking expectation over $\gP$.
The generalization gap of a randomized algorithm $\gA$ that has trained on a dataset $\empD$ is measured as
\begin{align*}
    \mathrm{gen}(\mathcal{A})
    := \left|\mathbb{E}_{\substack{
    \mathcal{A}, \empD \sim \mathcal{P},
    |\empD| = \numprompts - 1
    }}
    \left[\ell (\mathcal{A}(\empD)) - \ell (\mathcal{A}(\empD), \empD)\right] \right|.
\end{align*}

\cite{rammal2022leave1one1out} show that the expected generalization gap of $\mathcal{A}$ satisfies
\begin{equation}
    \mathrm{gen}(\mathcal{A})
    \;\le\;
    \frac{\numprompts}{\sqrt{2}(\numprompts-1)}
    \mathbb{E}_{\empD \sim \mathcal{P}} \sqrt{
            I_{\mathrm{LOO}}(\mathcal{A}; \empD)
    }.
    \label{eq:loo_cmi_gen_bound}
\end{equation}


 \newpage
\subsection{Connecting \gradcv{} to LOO-CMI}

All our discussions focus on the original gradient without pre-processing.

\paragraph{Notations.}
Let $\Theta_{\cal S} \in \mathbb{R}^{D}$ denote the parameters of the student model before training.
For the theoretical analysis, we use $\numgenerations=1$ teacher response per prompt on the $\numprompts$ prompts in the training set $\empD$,
though our results can be generalized to $\numgenerations > 1$.
From now on, we will use the uniformly random prompt-generation pair $(\prompt, \gen) \in \empD$ in place of the scale random variable $\idx \sim \text{Unif}([\numprompts])$ where $|\empD| = \numprompts$.
For example, we will use $\empD_{-(\prompt, \gen)}$ to denote the dataset with $(\prompt, \gen)$ removed from $\empD$. For simplicity, we will drop $\Theta _{\cal S}$ from all our notations involving $\precond$ and $\mathbf{g}$.

We will use $\Hat{\mathbb{E}}$ to denote the empirical expectation over $\empD$. 
For each $(\prompt, \gen)$, we perform a single gradient update with a preconditioner matrix $\precond$ that can depend on the training set $\empD_{-(\prompt, \gen)}$:
\begin{align}
    \Theta_{-(\prompt, \gen)} \;\gets\; \Theta_{\cal S} - \eta \, \Hat{\E}_{(\Bar{\prompt},\Bar{\gen})\in \empD_{-(\prompt, \gen)}} 
    \big[ \precond(\empD_{-(\prompt, \gen)}) \mathbf{g}(\Bar{\prompt}, \Bar{\gen}) \big] + \epsilon, \label{eq:pre-condition-update}
\end{align}
where $\epsilon \sim \mathcal{N}(0, \sigma^2 \mathbf{I})$ denotes a Gaussian noise, which represents the randomness in our training algorithm.

For any dataset $\Hat{\empD}$, we define the following notations:
\begin{align*}
    \mu(\Hat{\empD}) &= \Hat{\mathbb{E}}_{ (\prompt, \gen) \in \Hat{\empD} } [\mathbf{g}(\prompt, \gen)],
    \quad
    \Tilde{\Sigma}( \Hat{\empD} ) = \frac{1}{|\Hat{\empD}|} \Tilde{\mathbf{G}} (\Hat{\empD}) ^{\top} \Tilde{\mathbf{G}} (\Hat{\empD}),
    \quad
    \Hat{\Sigma}( \Hat{\empD} ) = \Tilde{\Sigma}( \Hat{\empD} ) + \frac{\nu}{\gradientdim} \mI,
\end{align*}
where $\nu>0$ is added for numerical stability.

We will further use the following shorthands:
\begin{align*}
    \mu_{- (\prompt, \gen)} := \mu \left(\empD_{-(\prompt, \gen)}\right),
    \quad
    &\precond_{- (\prompt, \gen)} :=  \precond \left(\empD_{-(\prompt, \gen)}\right).
\end{align*}

Then, the CMI bounds for the preconditioned update step in \cref{eq:pre-condition-update} are given by:






\noindent\fcolorbox{blue}{blue!5}{
\parbox{\dimexpr\linewidth-2\fboxsep-2\fboxrule}{
\begin{lemma}[Bounds for Pre-conditioned Gradient Descent]\label{lem:formal}
    Suppose $\exists \beta, B > 0$ such that $\norm{\precond_{-(\prompt, \gen)} - \precond_{-(\Bar{\prompt}, \Bar{\gen})}} \leq \frac{\beta}{\numprompts}$ for all $(\Bar{\prompt}, \Bar{\gen}), ({\prompt}, {\gen}) \in \empD$,
    and $\|\mathbf{g}(\prompt, \gen)\| \leq B$ for all $(\prompt, \gen) \in \empD$.
    Under the one-step update rule on the parameter $\Theta$ with learning rate $\eta$ (\cref{eq:pre-condition-update}) as algorithm $\mathcal{A}$, the LOO-CMI can be bounded as
    \begin{align}
        I_{\mathrm{LOO}}(\gA; \empD)
        \leq &\frac{3\eta^2}{\sigma^2 (\numprompts-1)^2 } \quad 
        \Hat{\mathbb{E}}_{(\prompt, \gen) \in \empD}  \quad  \norm{ \precond_{- (\prompt, \gen) } \mathbf{g}_{\mu}(\prompt, \gen) }^2 \nonumber \\
        & + \frac{3\eta^2}{\sigma^2 } \Hat{\mathbb{E}}_{(\prompt, \gen) \in \empD} \norm{ \left(\precond_{- (\prompt, \gen) } - \Hat{\mathbb{E}}_{(\Hat{\prompt}, \Hat{\gen}) \in \empD} \precond_{- (\Hat{\prompt}, \Hat{\gen}) } \right) \mu(\empD) }^2  + \mathcal{O}(\beta/\numprompts^4), \label{eq:formal_eq_cmi}\\
        & \text{where } \mathbf{g}_{\mu}(\prompt, \gen) = \mathbf{g} (\prompt, \gen) - \mu(\empD) \nonumber. 
    \end{align}
\end{lemma}
}}
\addtocounter{lemma}{-1}


\textbf{Special cases:} There are two notable special cases: the first is when $\precond = \mathbf{I}$, corresponding to update step in the standard gradient descent update algorithm. The second is when $\precond = \Tilde{\Sigma}^{-1/2}$, which denotes the preconditioner matrix used to define \gradcv{}.

\noindent\fcolorbox{orange}{orange!5}{
\parbox{\dimexpr\linewidth-2\fboxsep-2\fboxrule}{
\begin{corollary}[Connecting CMI to \gnorm{}]\label{cor:identity}
    When $\precond$ is set as identity matrix $\mathbf{I}$ for all $(\prompt, \gen) \in \empD$, then LOO-CMI for the one-step update rule on the parameter $\Theta$ with learning rate $\eta$ given by
    \begin{align*}
        I_{\mathrm{LOO}}(\gA; \empD)
        &\leq \frac{3\eta^2}{\sigma^2 (\numprompts-1)^2}  \Hat{\mathbb{E}}_{(\prompt, \gen) \in \empD}  \quad  \norm{ \mathbf{g}_{\mu}(\prompt, \gen) }^2  \\
        & \text{where } \mathbf{g}_{\mu}(\prompt, \gen) = \mathbf{g} (\prompt, \gen) - \mu(\empD)
    \end{align*}
    This connects CMI to the \gnorm{} (\cref{eq:mean_grad_norm}) metric.
\end{corollary}
}}

\noindent\fcolorbox{orange}{orange!5}{
\parbox{\dimexpr\linewidth-2\fboxsep-2\fboxrule}{
\begin{corollary}[Connecting CMI to \gradcv{}]\label{cor:precond}
     $\precond(\Hat{\empD})$ is set as $\Hat{\Sigma}(\Hat{\empD})^{-1/2}$ for any training set $\Hat{\empD}$. Denote $\Hat{\Sigma}_{-\{({\prompt}, {\gen})\}} := \Hat{\Sigma}(\empD_{-\{({\prompt}, {\gen})\}})$ for any $(\prompt, \gen)$ pair for simplicity. 
     Suppose there exists a constant $0 < \alpha < 1$, 
     s.t. for any $(\prompt, \gen) \in \empD$, 
     \begin{align}
          \norm{ \left(\Hat{\Sigma}^{-1/2}_{- (\prompt, \gen) } - \Hat{\mathbb{E}}_{(\Hat{\prompt}, \Hat{\gen}) \in \empD} \Hat{\Sigma}^{-1/2}_{- (\Hat{\prompt}, \Hat{\gen}) } \right) \mu(\empD) }^2 < \frac{\alpha}{\numprompts^2} \norm{ \Hat{\Sigma}^{-1/2}_{-(\prompt, \gen) } \mathbf{g}_{\mu}(\prompt, \gen) }^2. \label{eq:ass_precond}
     \end{align}

    Then LOO-CMI for the one-step update rule on the parameter $\Theta$ with learning rate $\eta$ is bounded as
    \begin{align*}
        I_{\mathrm{LOO}}(\gA; \empD)
        &\leq \frac{3 (1+\alpha) \eta^2}{\sigma^2 (\numprompts-1)^2 } \quad 
        \Hat{\mathbb{E}}_{(\prompt, \gen) \in \empD}  \quad  \norm{ \Hat{\Sigma}^{-1/2}_{- (\prompt, \gen) } \mathbf{g}_{\mu}(\prompt, \gen) }^2 + \mathcal{O}(\beta/\numprompts^4) \\
        & := \frac{3\eta^2 (1+\alpha)}{\sigma^2 (\numprompts-1)^2} \text{\gradcv{}-Variance}(\empD) + \mathcal{O}(\beta/\numprompts^4) \\
        & \leq \frac{3\eta^2 (1+\alpha)}{\sigma^2 (\numprompts-1)^2} \text{\gradcv{}}(\empD) + \mathcal{O}(\beta/\numprompts^4) \\
        & \text{where } \mathbf{g}_{\mu}(\prompt, \gen) = \mathbf{g} (\prompt, \gen) - \mu (\empD).  
    \end{align*}
    This connects CMI to the \gradcv{} (\cref{eq:gradcv}) score with $\numprompts$-fold cross-validation.
\end{corollary}
}}



\looseness-1\Cref{cor:precond} indicates that \gradcv{} evaluates the stability of a one-step gradient update when a few prompts are removed from the batch.
Importantly, the outcome of this update depends on the optimization method, since gradient descent and preconditioned updates can behave differently.
In our setting, the preconditioner matrix is closely related to the one used in AdaGrad \citep{duchi2011adaptive}. Since adaptive optimizers are the de facto choice for training language models, it is essential to incorporate this preconditioning effect in our analysis. 

\looseness-1However, the pre-conditioner matrix that we use doesn't exactly match the optimization states of the Adam optimization algorithm that we use in practice, leaving space for further improving the definition of \gradcv{}. This might require a short warm-up training phase of the student model and setting $\precond$ as a function of the optimizer states during the warm-up training, akin to \cite{xia24less}. We leave a more thorough exploration of this direction to future work.


\textbf{Note on theoretical limitations:} Our current analysis  establishes a connection between \gradcv{} and an upper bound on the LOO-CMI.
While LOO-CMI has been shown to upper bound the generalization gap in terms of the loss~\citet{rammal2022leave1one1out},
our result has two limitations.
First, \Cref{lem:formal} is stated only for a single gradient step, while in practice we compare the student's performance across different teachers after extensive training.
Second, the loss may not reflect the performance metric of interest;
in particular, the student model's test loss after training does not correlate with the student performance when compared across different teachers.
Furthermore, our empirical results show that \gradcv{} reliably predicts of student performance but correlate poorly with loss-based quantities.
\footnote{We will extensively discuss the comparison of \gradcv{} and loss in \cref{sec:connect_GRACE_to_Loss}.}
Such theory-practice gap highlights the need for further theoretical understanding of \gradcv{}, which we leave to future work.

\begin{figure}
    \centering
    \includegraphics[width=0.6\linewidth]{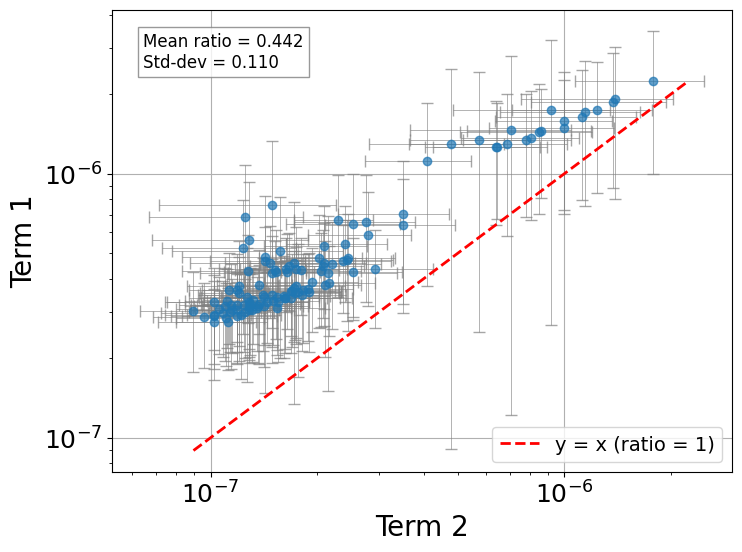}
    \caption{Validating assumption \cref{eq:ass_precond} in \cref{cor:precond} using teachers on \llamaS{} and \gsm{} using $\numprompts=512$. 
    Here Term 1 is given by $\frac{1}{\numprompts^2} \norm{ \Hat{\Sigma}^{-1/2}_{- (\prompt, \gen) } \mathbf{g}_{\mu}(\prompt, \gen) }$, Term 2 is given by $\norm{ \left(\Hat{\Sigma}^{-1/2}_{- (\prompt, \gen) } - \Hat{\mathbb{E}}_{(\Hat{\prompt}, \Hat{\gen}) \in \empD} \Hat{\Sigma}^{-1/2}_{- (\Hat{\prompt}, \Hat{\gen}) } \right) \mu(\empD) }^2$. $\alpha$ is defined as Term 2 / Term 1. \textbf{Main takeaway:} Term 1 dominates Term 2 across all teachers, with mean $\alpha=0.442$, validating the assumption in \cref{cor:precond}. }
    \label{fig:alpha}
\end{figure}

\textbf{Validating assumption in \cref{eq:ass_precond}:} 
For simplicity, denote the two terms of interest as
\begin{align*}
    \text{Term 1:}& \frac{1}{\numprompts^2} \norm{ \Hat{\Sigma}^{-1/2}_{- (\prompt, \gen) } \mathbf{g}_{\mu}(\prompt, \gen) } \\
    \text{Term 2:}& \norm{ \left(\Hat{\Sigma}^{-1/2}_{-(\prompt, \gen)} - \Hat{\mathbb{E}}_{(\Hat{\prompt}, \Hat{\gen}) \in \empD} \Hat{\Sigma}^{-1/2}_{- (\Hat{\prompt}, \Hat{\gen}) } \right) \mu(\empD) }^2.
\end{align*}
Expectation of Term 1 gives \gradcv{} with $\numCV=\numprompts$ and is the main term of interest. Term 2 measures the stability of the pre-conditioning matrix $\Tilde{\Sigma}$ when a prompt is dropped. The ratio of Term 1 and Term 2 measures how dominant is Term 1 compared to Term 2 in \cref{eq:formal_eq_cmi}, formally referred to as $\alpha$ in \cref{eq:ass_precond}. If $\alpha < 1$, this implies the bound on the conditional mutual information is primarily driven by our formulation of \gradcv{}.
Empirically, on \llamaS{} training on \gsm{}, we show that $\alpha$ has an average value between $0.442$ across all teachers in \cref{fig:alpha}.


\subsubsection{Proof of \cref{lem:formal}}

\begin{proof}

    For any $(\prompt, \gen)$ pair, denote the mean parameter update on the training set $\empD_{-(\prompt, \gen)}$ as 
    $\delta_{-(\prompt, \gen)} := - \eta \precond_{- (\prompt, \gen)} \mu_{- (\prompt, \gen)}$.
    Starting from student parameters $\Theta_{{\cal S}}$,
    the update rule for any set $\empD \setminus \{(\prompt, \gen)\}$ is given by 
    \begin{align*}
        \Theta_{-(\prompt, \gen)} \gets \Theta_{{\cal S}} + \delta_{-(\prompt, \gen)} + \epsilon,
    \end{align*}
    where $\epsilon \sim \gN(0, \sigma^2 I)$ is a Gaussian noise and captures the randomness in the algorithm.
    Hence $\Theta_{-(\prompt, \gen)}$ follows the distribution
    \begin{align*}
        \Theta_{-(\prompt, \gen)} \sim \mathcal{N}\left(\Theta_{{\cal S}} + \delta_{-(\prompt, \gen)}, \sigma^2 \mathbf{I} \right).
    \end{align*}

    Then, using the properties of Gaussian distribution;
\begingroup
\allowdisplaybreaks
\begin{align}
&
I_{\mathrm{LOO}}(\gA; \empD)
= \Hat{\mathbb{E}}_{(\prompt, \gen) \in \empD} D_{\mathrm{KL}} \!\left( 
    P_{\Theta_{-(\prompt, \gen)} } \,\bigg\|\, 
    \Hat{\mathbb{E}}_{(\Hat{\prompt}, \Hat{\gen}) \in \empD} P_{\Theta_{-(\Hat{\prompt}, \Hat{\gen})} } 
\right) \nonumber\\[6pt]
=& \Hat{\mathbb{E}}_{(\prompt, \gen) \in \empD} 
   \mathbb{E}_{\zeta \sim \mathcal{N}\!\left( \delta_{-(\prompt, \gen)}, \sigma^2 \mathbf{I} \right)}\left(
    \log \!\left( \tfrac{1}{Z} e^{-\norm{\zeta - \delta_{-(\prompt, \gen)}}^2 / 2\sigma^2} \right) 
     - \log { \Hat{\mathbb{E}}_{(\Hat{\prompt}, \Hat{\gen}) \in \empD} \!\left( \tfrac{1}{Z} e^{-\norm{\zeta - \delta_{-(\Hat{\prompt}, \Hat{\gen})}}^2 / 2\sigma^2} \right) } \right) \label{eq:simplify_cmi_eq1} \\
    & \leq
    \Hat{\mathbb{E}}_{(\prompt, \gen) \in \empD} 
   \mathbb{E}_{\zeta \sim \mathcal{N}\!\left( \delta_{-(\prompt, \gen)}, \sigma^2 \mathbf{I} \right)}
    \left( \log \!\left( \tfrac{1}{Z} e^{-\norm{\zeta - \delta_{-(\prompt, \gen)}}^2 / 2\sigma^2} \right) 
     - \Hat{\mathbb{E}}_{(\Hat{\prompt}, \Hat{\gen}) \in \empD} \log {  \!\left( \tfrac{1}{Z} e^{-\norm{\zeta - \delta_{-(\Hat{\prompt}, \Hat{\gen})}}^2 / 2\sigma^2} \right) }  \right) \label{eq:simplify_cmi_eq2}\\&
     = \frac{1}{2\sigma^2}\Hat{\mathbb{E}}_{(\prompt, \gen) \in \empD} 
   \mathbb{E}_{\zeta \sim \mathcal{N}\!\left( \delta_{-(\prompt, \gen)}, \sigma^2 \mathbf{I} \right)}
    \left( -\norm{\zeta - \delta_{-(\prompt, \gen)}}^2 + \Hat{\mathbb{E}}_{(\Hat{\prompt}, \Hat{\gen}) \in \empD} \norm{\zeta - \delta_{-(\Hat{\prompt}, \Hat{\gen})}}^2  \right) \nonumber\\&
    = \frac{1}{2\sigma^2}\Hat{\mathbb{E}}_{(\prompt, \gen) \in \empD} \Hat{\mathbb{E}}_{(\Hat{\prompt}, \Hat{\gen}) \in \empD} \mathbb{E}_{\zeta \sim \mathcal{N}\!\left( \delta_{-(\prompt, \gen)}, \sigma^2 \mathbf{I} \right)} \left( -\norm{\zeta - \delta_{-(\prompt, \gen)}}^2 + \norm{\zeta - \delta_{-(\Hat{\prompt}, \Hat{\gen})}}^2  \right) \nonumber\\&
    = \frac{1}{2\sigma^2}\Hat{\mathbb{E}}_{(\prompt, \gen) \in \empD} \Hat{\mathbb{E}}_{(\Hat{\prompt}, \Hat{\gen}) \in \empD}  \norm{ \delta_{-(\Hat{\prompt}, \Hat{\gen})} - \delta_{-(\prompt, \gen)}}^2 \label{eq:simplify_cmi_eq3}\\
    &= \frac{1}{\sigma^2} \Hat{\mathbb{E}}_{(\prompt, \gen) \in \empD} \norm{ \delta_{-(\prompt, \gen)}  - \Hat{\mathbb{E}}_{(\Hat{\prompt}, \Hat{\gen}) \in \empD} \delta_{-(\Hat{\prompt}, \Hat{\gen})} }^2 \label{eq:simplify_cmi_eq4}
\end{align}
\endgroup
In the second step (\cref{eq:simplify_cmi_eq1}), we simply use the CDF formulation of gaussian distribution, where $Z = (2\pi e)^{-D}.$ The third step applies a jensen's inequality (\cref{eq:simplify_cmi_eq2}). In \cref{eq:simplify_cmi_eq3}, we use 
for any two vectors $\mathbf{a}, \mathbf{b}$: 
\begin{align*}
    \mathbb{E}_{\zeta \sim \mathcal{N}\!\left( \mathbf{a}, \sigma^2 \mathbf{I} \right)} \left( -\norm{\zeta - \mathbf{a}}^2 + \norm{\zeta - \mathbf{b}}^2\right) &= \mathbb{E}_{\zeta \sim \mathcal{N}\!\left( \mathbf{a}, \sigma^2 \mathbf{I} \right)} 2(\mathbf{a} - \mathbf{b})^{\top} \zeta - \norm{\mathbf{a}}^2 + \norm{\mathbf{b}}^2
    \\&
    = 2(\mathbf{a} - \mathbf{b})^{\top} \mathbf{a} - \norm{\mathbf{a}}^2 + \norm{\mathbf{b}}^2 \\&
    = \norm{\mathbf{b} - \mathbf{a}}^2.
\end{align*}
Finally, \cref{eq:simplify_cmi_eq4} follows because both  $ \delta_{-(\prompt, \gen)}$  and  $\delta_{-(\Hat{\prompt}, \Hat{\gen})}$ come from the identical set.


Using the definition of $\delta$, we have from \cref{eq:simplify_cmi_eq4}:
\begin{align*}
    \mathbb{E}_{u \sim \text{Unif}([\numprompts])} I(\Theta_{-u}; u \mid \empD) \leq \frac{1}{\sigma^2} \Hat{\mathbb{E}}_{(\prompt, \gen) \in \empD} \norm{ \precond_{- (\prompt, \gen) } \mu_{- (\prompt, \gen)}  - \Hat{\mathbb{E}}_{(\Hat{\prompt}, \Hat{\gen}) \in \empD} \precond_{- (\Hat{\prompt}, \Hat{\gen}) } \mu_{- (\Hat{\prompt}, \Hat{\gen})} }^2
\end{align*}

\textbf{Warmup: When the pre-conditioner is identity matrix} Then for any $(\prompt, \gen)$ pair, we have $\precond_{- (\prompt, \gen) } = \mathbf{I}$. Then, the formulation simplifies to
\begin{align}
    &\mathbb{E}_{u \sim \text{Unif}([\numprompts])} I(\Theta_{-u}; u \mid \empD) \nonumber\\&\leq \frac{\eta^2}{\sigma^2} \Hat{\mathbb{E}}_{(\prompt, \gen) \in \empD} \norm{  \mu_{- (\prompt, \gen)}  - \Hat{\mathbb{E}}_{(\Hat{\prompt}, \Hat{\gen}) \in \empD} \mu_{- (\Hat{\prompt}, \Hat{\gen})} }^2
    \nonumber\\& = \frac{\eta^2}{\sigma^2} \Hat{\mathbb{E}}_{(\prompt, \gen) \in \empD} \norm{ \frac{\numprompts}{\numprompts-1} \mu(\empD) - \frac{1}{\numprompts-1} \mathbf{g} (\prompt, \gen)  - \Hat{\mathbb{E}}_{(\Hat{\prompt}, \Hat{\gen}) \in \empD} \left(\frac{\numprompts}{\numprompts-1} \mu(\empD) - \frac{1}{\numprompts-1} \mathbf{g} (\Hat{\prompt}, \Hat{\gen}) \right) }^2 \label{eq:identity_cmi_eq1}\\&
    = \frac{\eta^2}{\sigma^2 (\numprompts-1)^2} \Hat{\mathbb{E}}_{(\prompt, \gen) \in \empD}  \norm{ \mathbf{g} (\prompt, \gen) - \Hat{\mathbb{E}}_{(\Hat{\prompt}, \Hat{\gen}) \in \empD} \mathbf{g} (\Hat{\prompt}, \Hat{\gen})  }^2 \nonumber \\&
    := \frac{\eta^2}{\sigma^2 (\numprompts-1)^2} \Hat{\mathbb{E}}_{(\prompt, \gen) \in \empD}  \norm{ \mathbf{g} (\prompt, \gen) - \mu(\empD)  }^2 := \frac{\eta^2}{\sigma^2 (\numprompts-1)^2} \Hat{\mathbb{E}}_{(\prompt, \gen) \in \empD}  \norm{ \mathbf{g}_{\mu} (\prompt, \gen)   }^2\nonumber
\end{align}
The first step (\cref{eq:identity_cmi_eq1}) follows from the fact that $\mu(\empD) = \Hat{\mathbb{E}}_{(\Hat{\prompt}, \Hat{\gen}) \in \empD} \mathbf{g}( \Hat{\prompt}, \Hat{\gen} ) := \frac{1}{\numprompts} \sum_{ (\Hat{\prompt}, \Hat{\gen}) \in \empD } \mathbf{g}( \Hat{\prompt}, \Hat{\gen} )$ and $\mu_{- (\prompt, \gen)} =  \frac{1}{\numprompts-1} \sum_{ ({\prompt'}, {\gen'}) \in \empD_{-(\prompt, \gen)} } \mathbf{g}( \prompt', \gen' )$. 

\textbf{General pre-conditioner $\precond$:} We follow similar steps as above:
\begingroup
\allowdisplaybreaks
\begin{align}
    &\mathbb{E}_{u \sim \text{Unif}([\numprompts])} I(\Theta_{-u}; u \mid \empD) \nonumber 
    \\&\leq \frac{\eta^2}{\sigma^2} \Hat{\mathbb{E}}_{(\prompt, \gen) \in \empD} \norm{  \precond_{- (\prompt, \gen)} \mu_{- (\prompt, \gen)}  - \Hat{\mathbb{E}}_{(\Hat{\prompt}, \Hat{\gen}) \in \empD} \precond_{- (\Hat{\prompt}, \Hat{\gen})} \mu_{- (\Hat{\prompt}, \Hat{\gen})} }^2
    \nonumber
    \\& = \frac{\eta^2}{\sigma^2} \Hat{\mathbb{E}}_{(\prompt, \gen) \in \empD} \Big\| \frac{\numprompts}{\numprompts-1} \precond_{- (\prompt, \gen)}  \mu(\empD) - \frac{1}{\numprompts-1} \precond_{- (\prompt, \gen)} \mathbf{g} (\prompt, \gen)  \nonumber
    \\& \quad\quad\quad - \Hat{\mathbb{E}}_{(\Hat{\prompt}, \Hat{\gen}) \in \empD} \left(\frac{\numprompts}{\numprompts-1}  \precond_{- (\Hat{\prompt}, \Hat{\gen})} \mu(\empD) - \frac{1}{\numprompts-1} \precond_{- (\Hat{\prompt}, \Hat{\gen})} \mathbf{g} (\Hat{\prompt}, \Hat{\gen}) \right) \Big\|^2 \tag*{\text{(similar to \cref{eq:identity_cmi_eq1})}} \nonumber
    \\&
    = \frac{\eta^2}{\sigma^2} \Hat{\mathbb{E}}_{(\prompt, \gen) \in \empD} \Big\| \frac{\numprompts}{\numprompts-1} \left( \precond_{- (\prompt, \gen)} - \Hat{\mathbb{E}}_{(\Hat{\prompt}, \Hat{\gen}) \in \empD} \precond_{- (\Hat{\prompt}, \Hat{\gen})} \right) \mu(\empD) \nonumber
    \\&
    \quad \quad \quad \quad \quad - \frac{1}{\numprompts-1} \left( \precond_{- (\prompt, \gen)} \mathbf{g} (\prompt, \gen) - \Hat{\mathbb{E}}_{(\Hat{\prompt}, \Hat{\gen}) \in \empD} \left( \precond_{- (\Hat{\prompt}, \Hat{\gen})} \mathbf{g} (\Hat{\prompt}, \Hat{\gen}) \right) \right) \Big\|^2 \nonumber
    \\&
    = \frac{\eta^2}{\sigma^2} \Hat{\mathbb{E}}_{(\prompt, \gen) \in \empD} \Big\| \frac{\numprompts}{\numprompts-1} \left( \precond_{- (\prompt, \gen)} - \Hat{\mathbb{E}}_{(\Hat{\prompt}, \Hat{\gen}) \in \empD} \precond_{- (\Hat{\prompt}, \Hat{\gen})} \right) \mu(\empD) \nonumber
    \\&
    \quad \quad \quad \quad \quad \quad \quad \quad  - \frac{1}{\numprompts-1}  \precond_{- (\prompt, \gen)} \left( \mathbf{g} (\prompt, \gen) - \Hat{\mathbb{E}}_{(\Hat{\prompt}, \Hat{\gen}) \in \empD}  \mathbf{g} (\Hat{\prompt}, \Hat{\gen}) \right) \nonumber
    \\&
    \quad \quad \quad \quad \quad \quad \quad \quad   - \frac{1}{\numprompts-1}  \Hat{\mathbb{E}}_{(\Hat{\prompt}, \Hat{\gen}) \in \empD} \left( \precond_{- (\prompt, \gen)} - \precond_{- (\Hat{\prompt}, \Hat{\gen})} \right)   \mathbf{g} (\Hat{\prompt}, \Hat{\gen}) \Big\|^2 \nonumber
    \\&
    \leq \frac{3\eta^2}{\sigma^2} \Hat{\mathbb{E}}_{(\prompt, \gen) \in \empD} \left(\frac{\numprompts}{\numprompts-1}\right)^2 \Hat{\mathbb{E}}_{(\prompt, \gen) \in \empD} \norm{  \left( \precond_{- (\prompt, \gen)} - \Hat{\mathbb{E}}_{(\Hat{\prompt}, \Hat{\gen}) \in \empD} \precond_{- (\Hat{\prompt}, \Hat{\gen})} \right) \mu(\empD) }^2 \nonumber
    \\&
    \quad \quad + \frac{3\eta^2}{\sigma^2} \frac{1}{(\numprompts-1)^2} \Hat{\mathbb{E}}_{(\prompt, \gen) \in \empD} \norm{ \Hat{\mathbb{E}}_{(\Hat{\prompt}, \Hat{\gen}) \in \empD} \left( \precond_{- (\prompt, \gen)} - \precond_{- (\Hat{\prompt}, \Hat{\gen})} \right)   \mathbf{g} (\Hat{\prompt}, \Hat{\gen}) }^2 \label{eq:preconditioner_cmi_eq0}
    \\&
    \quad \quad + \frac{3\eta^2}{\sigma^2} \frac{1}{(\numprompts-1)^2}  \Hat{\mathbb{E}}_{(\prompt, \gen) \in \empD} \norm{ \precond_{- (\prompt, \gen)} \left( \mathbf{g} (\prompt, \gen) - \Hat{\mathbb{E}}_{(\Hat{\prompt}, \Hat{\gen}) \in \empD}  \mathbf{g} (\Hat{\prompt}, \Hat{\gen}) \right) }^2 \nonumber
    \\&
     \leq \frac{3\eta^2}{\sigma^2} \Hat{\mathbb{E}}_{(\prompt, \gen) \in \empD} \left(\frac{\numprompts}{\numprompts-1}\right)^2 \Hat{\mathbb{E}}_{(\prompt, \gen) \in \empD} \norm{  \left( \precond_{- (\prompt, \gen)} - \Hat{\mathbb{E}}_{(\Hat{\prompt}, \Hat{\gen}) \in \empD} \precond_{- (\Hat{\prompt}, \Hat{\gen})} \right) \mu(\empD) }^2  \nonumber
     \\&
    \quad \quad + \frac{3\eta^2}{\sigma^2} \frac{1}{(\numprompts-1)^2}  \Hat{\mathbb{E}}_{(\prompt, \gen) \in \empD} \norm{ \precond_{- (\prompt, \gen)} \left( \mathbf{g} (\prompt, \gen) - \Hat{\mathbb{E}}_{(\Hat{\prompt}, \Hat{\gen}) \in \empD}  \mathbf{g} (\Hat{\prompt}, \Hat{\gen}) \right) }^2 + \mathcal{O}\left(\beta/\numprompts^4\right) \label{eq:preconditioner_cmi_eq1} 
\end{align}
\endgroup

\cref{eq:preconditioner_cmi_eq1} uses assumptions of bounded gradient norms and stability of the pre-conditioner matrix to bound the term in \cref{eq:preconditioner_cmi_eq0}.\end{proof}



 \subsection{Can we connect \gradcv{}'s predictive behavior of student performance to student's validation loss after training?}
 \label{sec:connect_GRACE_to_Loss}

In \cref{lem:cmi_result}, we establish a connection between leave-one-out conditional mutual information (LOO-CMI) and \gradcv{}. Because LOO-CMI has been used to derive generalization bounds for training algorithms, primarily measured via the model’s loss after training, we ask whether \gradcv{} correlates with the student’s loss after being trained on different teachers. To do so, we investigate two key questions: (a) does validation loss of the student after training correlate with student performance? and (b) does \gradcv{} correlate with the validation loss of the student after training?

\subsubsection{Does validation loss after training correlate to student performance?}

For each teacher, we create validation set using the teacher's responses on the validation set of prompts. We use the following loss functions of the student after training on a teacher's responses on a training set:
\begin{itemize}
    \item \textbf{Validation loss} is computed as the cross entropy loss of the student on the teacher's responses on the validation set. 
    \item \textbf{Validation - Train loss} is computed as the difference in the cross entropy loss of the student  on the teacher's responses on the training set and the  validation set.
    \item \textbf{Forward KL} is computed as the difference in the cross entropy loss of the student and the teacher  on the teacher's responses on the training set.
    \item \textbf{Backward KL} is computed as the difference in the cross entropy loss of the teacher and the student   on the student's responses on the training set.
    \item \textbf{Loss on pre-training} is computed as the  cross entropy loss of the student on the pre-training data. Because we don't have access to pre-training data of \llamaS{}, we use examples from fineweb-edu as a surrogate \citep{lozhkov2024fineweb-edu}.
\end{itemize}
We measure correlations between each loss score and the student average-at-16 performance after training across all teachers.

\begin{table}[h]
    \centering
    \begin{tabular}{c|c|c}
       Performance score (after training)  &  Absolute Spearman $(\uparrow)$ & Regret  $(\downarrow)$\\
       \toprule
       Validation Loss  & 0.15 & 4.1 $\%$ \\
       Validation Forward KL & 0.44 & 4.6 $\%$ \\
       Validation Backward KL & 0.14 & 12.9 $\%$ \\
       Validation - Train Loss &  0.17 & 4.1 $\%$\\
       \midrule
       Loss on pre-training & 0.07  & 6.5 $\%$ \\
       \bottomrule
    \end{tabular}
    \caption{Setting: \llamaS{} training on \gsm{}. \textbf{Observation: None of the loss variants of the student after training correlate with the student performance.} The performance scores have been computed on the student after training on each teacher generations.}
    \label{tab:loss_prediction}
\end{table}

\textbf{Primary observations:} In \cref{tab:loss_prediction}, we observe that none of the loss-based scores correlate well with student performance.
Several factors may explain this. First, the validation sets differ across teachers due to variations in their chain-of-thought reasoning styles, making the resulting losses across teachers non-comparable.
More critically, there exists an inherent mismatch between the cross-entropy loss and the auto-regressive generation objective: cross-entropy loss measures the average token-level prediction error on a fixed reference set, whereas auto-regressive evaluation tests whether the model can generate the correct solution through sequential token generation. Thus, a student can have low loss on the teacher's generations after training, but can struggle with getting correct auto-regressive generations.
This discrepancy has been extensively discussed in prior works~\citep{arora2022exposure,fang2024wrong}.

\subsubsection{Does \gradcv{} correlate with validation loss after training?}


Here, we study whether \gradcv{} correlates with validation loss of the student after training. To do so, we consider the correlations between the loss-based scores of the student after training, described in the previous section, and the scores \gradcv{}, \gvendi{}, and \gnorm{} computed on the gradients of the student before training.

\begin{table}[h]
    \centering
    \begin{tabular}{c|c|c|c}
       Performance score (after training)  & \gnorm{} & \gvendi{} & \gradcv{} \\
       \toprule
       Validation Loss  & 0.34 & 0.27 &  0.09  \\
       Validation Forward KL &  0.10 & 0.28 &  0.15 \\
       Validation Backward KL & 0.09 & 0.40 &  0.01 \\
       Validation - Train Loss & 0.0 & 0.08 & 0.23 \\
       \midrule
       Loss on pre-training data & 0.03 & -0.07 & 0.11  \\
       \bottomrule
    \end{tabular}
    \caption{Setting: \llamaS{} training on \gsm{}. Here, we measure absolute spearman correlations between each loss-based scores with each of  \gvendi{}, \gradcv{}, and \gnorm{}.
    \textbf{Observation: \gradcv{} doesn't correlate well with any of the loss variants of the student after training.} \gradcv{}, \gvendi{}, \gnorm{} have been computed using the gradients of the student before training, and the performance scores have been computed on the student after training. } 
    \label{tab:loss_to_score_correlation}
\end{table}

\textbf{Primary observations:} As shown in \cref{tab:loss_to_score_correlation}, \gradcv{} exhibits only a weak correlation with the student’s loss across teachers. Interestingly, \gvendi{} and \gnorm{} show even stronger correlation with the student’s loss-based scores. This suggests that the predictive effectiveness of \gradcv{} cannot be directly attributed to the student’s validation loss after training.

\paragraph{Conclusion:} Together, these gaps weaken the theoretical link between LOO-CMI and \gradcv{}, leaving open the question of why \gradcv{} correlates more strongly with student performance. Furthermore, since LOO-CMI only offers an upper bound on the generalization error, it cannot serve as a direct comparative metric. A deeper theoretical understanding of how \gradcv{} relates to student performance after training remains an important avenue for future investigation.

\section{More experimental results and comparisons}
\label{app:add_expts}

\subsection{Additional results }
\label{app:add_results}

\begin{figure}[htbp]
    \centering
     \includegraphics[width=\linewidth]{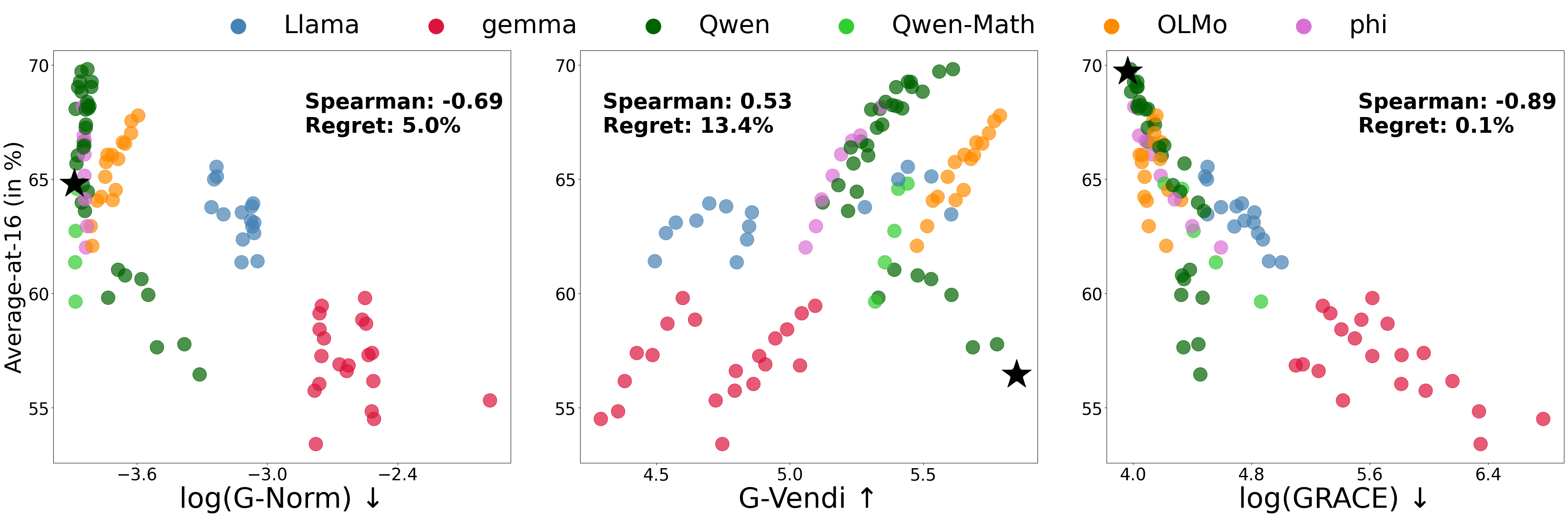}
    \caption{ \textbf{\gradcv{} achieves $89\%$ correlation to Gemma-2B performance after training on \gsm{}, across all teacher, generation temperature combinations. }
    \gnorm{} and \gvendi{} can  achieve 69\% and 53\% correlation respectively. 
    Here, ${\numprompts}=512, \gradientdim=512$ are used to compute all scores. When evaluated by regret, \gradcv{} selects a teacher that trains the student to within $0.1\%$ of the best achievable performance, outperforming \gnorm{} and \gvendi{}, which incur regrets of $5\%$ and $13.4\%$, respectively.
    } 
    \label{fig:across_all_temps_gsm_gemma}
\end{figure}

\paragraph{\gsm{}:} First, we show the behavior of Gemma-2B, when trained with different teachers in \cref{fig:across_all_temps_gsm_gemma}. We also observe that \gradcv{} has higher spearman correlation to the student performance, compared to \gnorm{} and \gvendi{}.

\begin{figure}[t]
    \centering
    \includegraphics[width=\linewidth]{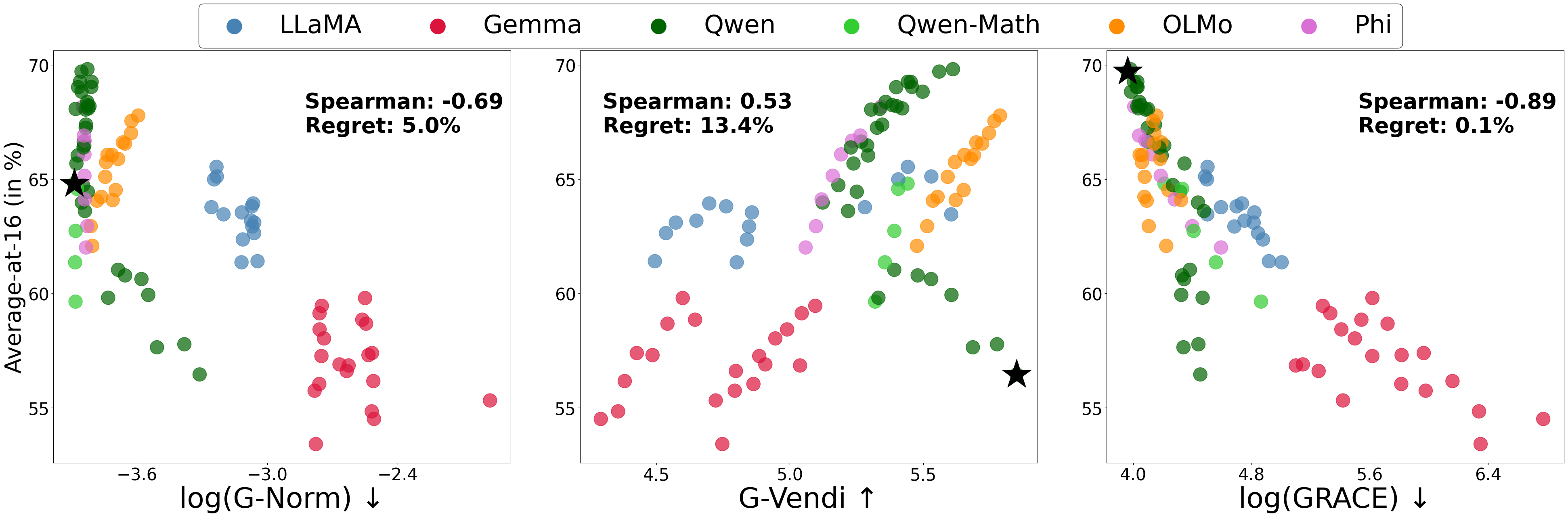}
    \caption{ Repeated experiment from \cref{fig:across_all_temps} but with $\gradientdim=1024$. \textbf{\gradcv{} achieves $90\%$ correlation to \llamaS{} performance after training on \gsm{}, across all teacher, generation temperature combinations. }
    \gnorm{} and \gvendi{} can only achieve 53\% and 47\% correlation respectively. When evaluated by regret, \gradcv{} selects a teacher that trains the student to within $0.3\%$ of the best achievable performance, outperforming \gnorm{} and \gvendi{}, which incur regrets of $10.8\%$ and $14.5\%$, respectively.
    } 
    \label{fig:across_all_temps_dim1024}
\end{figure}

\begin{figure}[t]
    \centering
    \includegraphics[width=\linewidth]{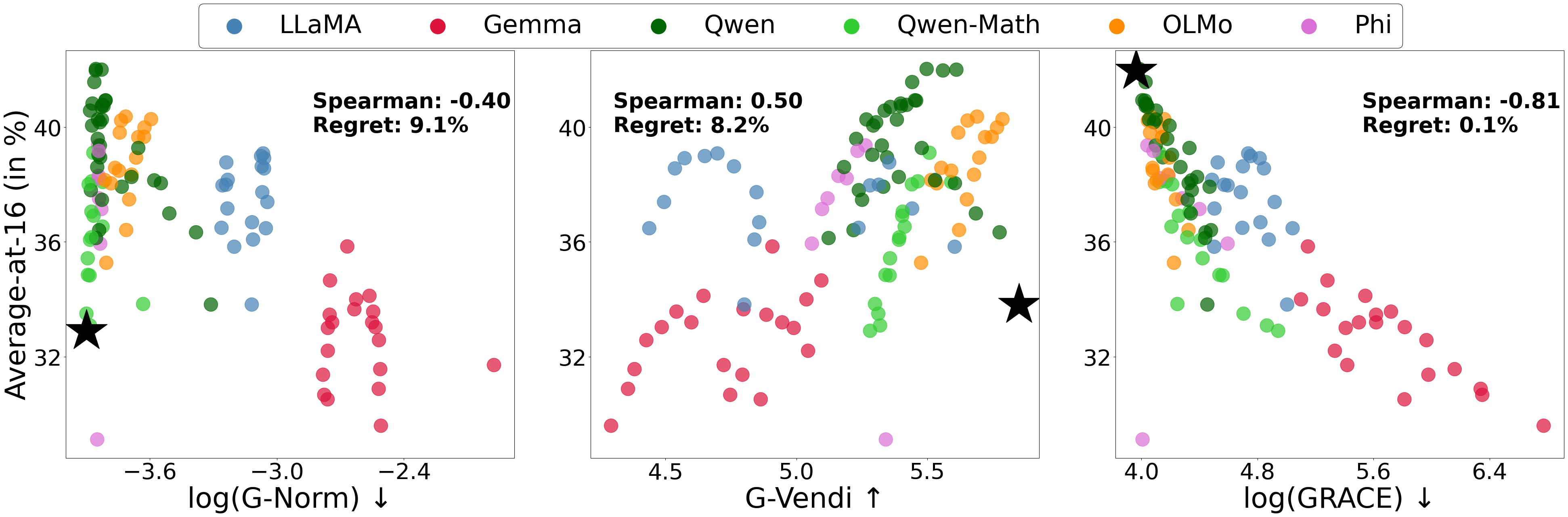}
    \caption{ Repeated experiment from \cref{fig:across_all_temps_olmo} but with $\gradientdim=1024$. \textbf{\gradcv{} achieves $81\%$ correlation to \olmoS{} performance after training on \gsm{}, across all teacher, generation temperature combinations. }
    \gnorm{} and \gvendi{} can only achieve 40\% and 50\% correlation respectively. When evaluated by regret, \gradcv{} selects a teacher that trains the student to within $0.1\%$ of the best achievable performance, outperforming \gnorm{} and \gvendi{}, which incur regrets of $9.1\%$ and $8.2\%$, respectively. 
    } 
    \label{fig:across_all_temps_olmo_dim1024}
\end{figure}

We also report the performance when we allow more computation for the computation of \gradcv{}. In \cref{fig:across_all_temps_olmo_dim1024,fig:across_all_temps_dim1024}, we show that the final spearman correlation improves by atleast $7\%$ and $4\%$ for \olmoS{} and \llamaS{} respectively on \gsm{}, when we increase the projected dimension of gradients from $512$ to $1024$.

\begin{figure}[htbp]
    \centering
    \includegraphics[width=\linewidth]{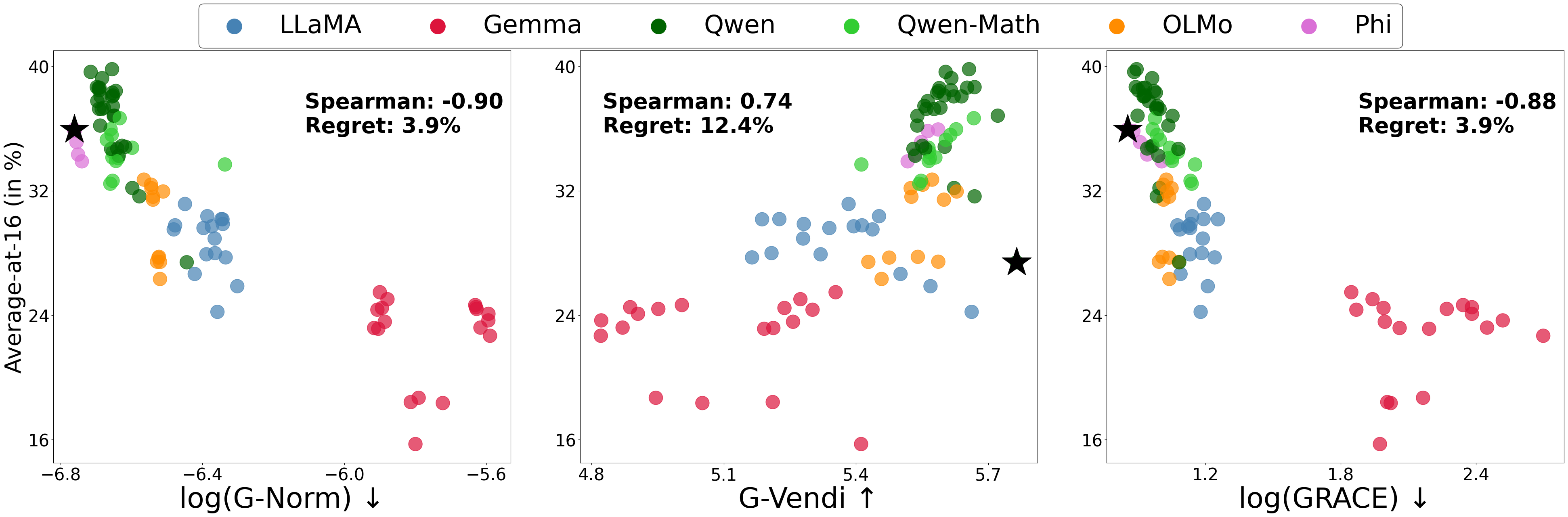}
    \caption{ \textbf{\gradcv{} achieves $88\%$ correlation to \llamaM{} performance after training on \math{}, across all teacher, generation temperature combinations. }
    \gnorm{} and \gvendi{} can  achieve 90\% and 74\% correlation respectively. 
    Here, ${\numprompts}=512, \gradientdim=512$ are used to compute all metrics. When evaluated by regret, \gradcv{} selects a teacher that trains the student to within $3.9\%$ of the best achievable performance (similar to \gnorm{}), outperforming \gvendi{}, which incurs regret of $12.4\%$.
    } 
    \label{fig:across_all_temps_math}
\end{figure}

\paragraph{\math{}:} In \cref{fig:across_all_temps_math}, we show the scatter plot of the performance of the model when trained with different teachers. We show that \gnorm{} and \gradcv{} achieve $89\%$ and $88\%$ spearman correlation to the student performance post distillation training, while \gvendi{} only achieves $74\%$ spearman correlation.

\subsection{Measuring effectiveness of \gradcv{} within each teacher family}
\label{app:family}

\begin{figure}[h]
    \centering
    \begin{minipage}[htbp]{0.8\linewidth}
        \includegraphics[width=\linewidth]{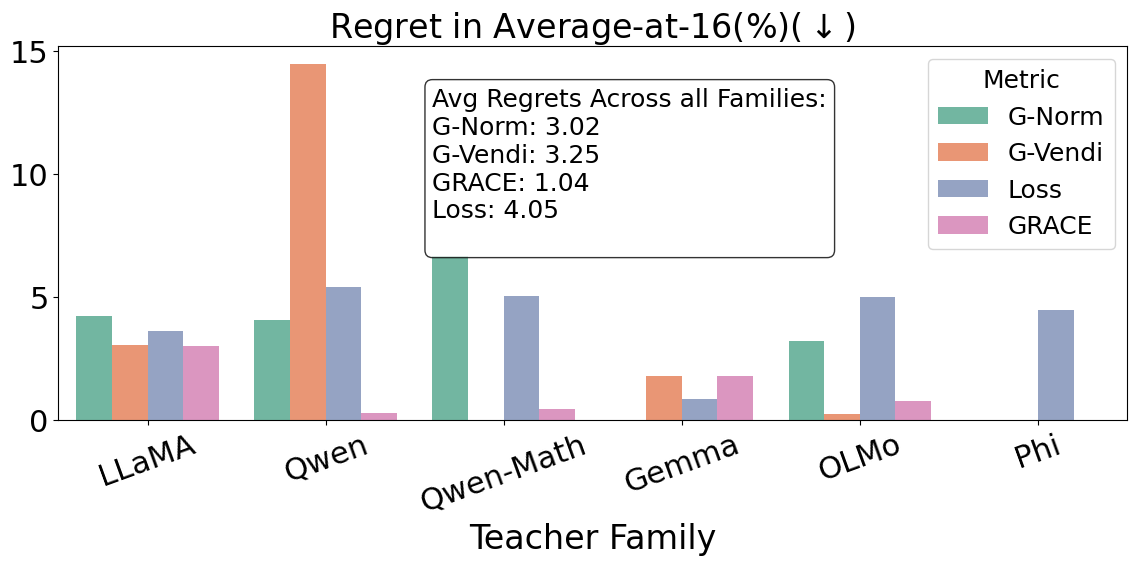}
    \end{minipage}%
    \hfill
    \caption{
    Here, we measure the regret incurred when a score is used to select a teacher within each teacher family. On average, \gradcv{} identifies a teacher that enables the student to achieve performance within $1.04\%$ of the best performance possible from distillation with a teacher in that family, whereas other scores incur an average regret of at least $3\%$.    }
    \label{fig:family_level_gaps}
\end{figure}

We measure the regret of teacher selection using each score within individual teacher families. For a given family, the regret of a score is defined as the performance gap between (i) the best student performance achieved after distilling from all teachers in that family and (ii) the student performance obtained when the teacher is chosen according to that score. On average, \gradcv{} yields only $1.04\%$ regret, while other metrics incur at least $3\%$. See \cref{fig:family_level_gaps} for details.

\subsection{Out-of-distribution evaluation}\label{app:ood}

\begin{figure}[h]
    \centering
    \includegraphics[width=\linewidth]{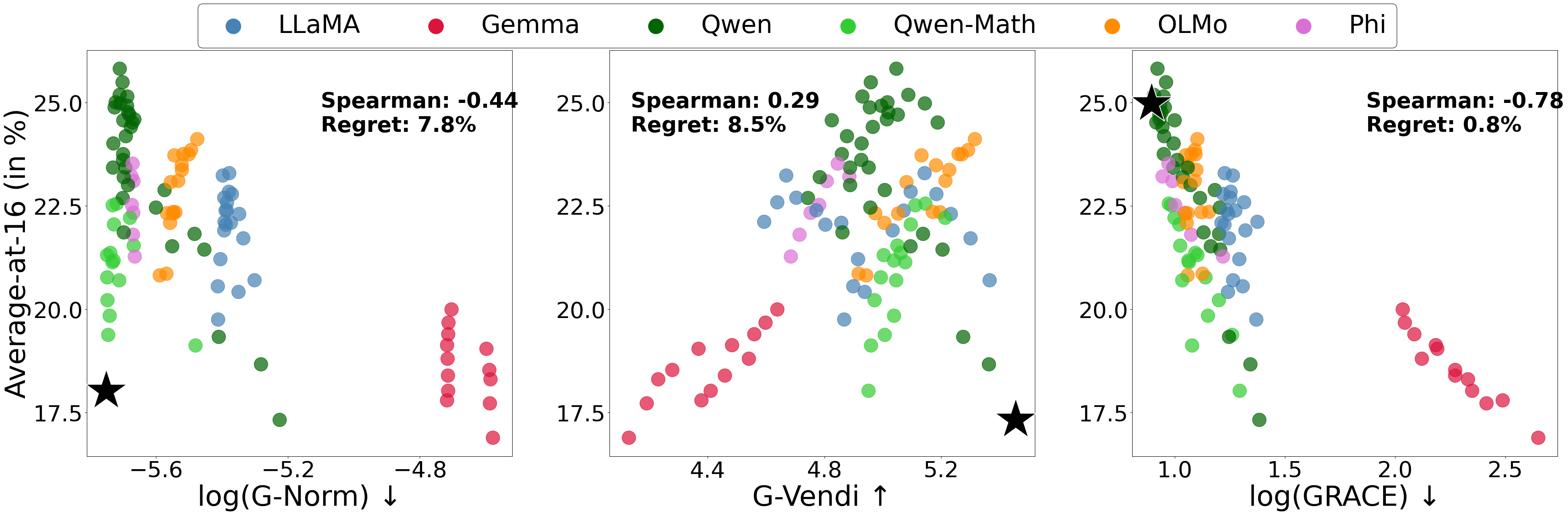}
    \caption{\textbf{Out-of-distribution results show consistent trends on \gsm{}.} We further evaluate the \llamaS{} students (trained on \gsm{}) on the GSM-Symbolic dataset \citep{mirzadeh2024gsm}, an out-of-distribution benchmark relative to \gsm{}. \gradcv{} achieves a Spearman correlation of $78\%$ with student performance, outperforming both \gnorm{} and \gvendi{}. As expected, this is slightly lower than the $86\%$ correlation observed in-distribution (\cref{fig:across_all_temps}). In terms of regret, \gradcv{} achieves the lowest regret ($0.8\%$), outperforming both \gnorm{} and \gvendi{} in this OOD setting.}
    \label{fig:gsm_symbolic_llama}
\end{figure}

One question is whether \gradcv{} can also predict the performance of the trained teacher in out-of-distribution settings. To check, we measure the performance of the trained \llamaS{} models (trained on \gsm{}) on GSM-Symbolic \citep{mirzadeh2024gsm} and measure the spearman correlations and regrets of \gnorm{}, \gvendi{}, and \gradcv{} with respect to the OOD performance (\cref{fig:gsm_symbolic_llama}). Similarly, we perform OOD evaluation of \llamaM{} models (trained on \math{}) on \math{}$^2$ \citep{shah2024ai} and \math{} Perturb (Simple \& Hard) \citep{huang2025math}, and measure the spearman correlations and regrets of \gnorm{}, \gvendi{}, and \gradcv{} with respect to the OOD performance (\cref{tab:math_perturb_llama}). We observe the following:
\begin{itemize}
    \item On \llamaS{}, \gradcv{} has both high spearman correlation ($78\%$) and low regret ($0.8\%$) compared to \gnorm{} and \gvendi{}. Thus, good in-distribution performance of the student also translates to good OOD performance.
    
    \item On \llamaM{}, we observe a more nuanced pattern. \gnorm{} attains higher Spearman correlation and lower regret than \gradcv{} on \math{} Perturb (Simple), which constitutes a relatively mild out-of-distribution shift. In contrast, \gradcv{} matches or surpasses \gnorm{} on the more challenging \math{}$^2$ and \math{} Perturb (Hard) settings. These results suggest that \gradcv{} provides a reliable indicator of student performance under substantial distribution shifts.
    
\end{itemize}

\begin{table}[h]
    \centering
    \begin{tabular}{c|c|c|c|c}
        Evaluation dataset & Best Student  &  Score & Absolute  & Regret ($\downarrow$) \\
        & Average-at-16 & & Spearman ($\uparrow$) & \\
        \toprule
        \multirow{3}{*}{\math{} Perturb (Simple)} & \multirow{3}{*}{31.32$\%$} & \gnorm{} & \textbf{0.89} & \textbf{2.8} $\%$ \\
        & & \gvendi{} & 0.59 & 16.7\% \\
        & & \gradcv{} & 0.76 & \textbf{2.8}  $\%$ \\
        \midrule
        \multirow{3}{*}{\math{}$^2$} & \multirow{3}{*}{9.36$\%$}  & \gnorm{} & 0.87 & \textbf{0.1} $\%$ \\
        & & \gvendi{} &  0.66 & $3.9\%$ \\
        & & \gradcv{} & \textbf{0.88} & \textbf{0.1}$\%$ \\
        \midrule
        \multirow{3}{*}{\math{} Perturb (Hard)} & \multirow{3}{*}{13.72$\%$} & \gnorm{} &  0.80 & \textbf{2.1} $\%$ \\
        & & \gvendi{} & 0.75 & $5\%$ \\
        & & \gradcv{} & \textbf{0.88} & \textbf{2.1$\%$} \\
        \bottomrule
    \end{tabular}
    \caption{ \looseness-1\textbf{Out-of-distribution results show consistent trends on \math{}.} We further evaluate the \llamaM{} trained students (on \math{}) on \math{}$^2$ \citep{shah2024ai} and \math{} Perturb (Simple \& Hard) \citep{huang2025math}, out-of-distribution benchmarks relative to \math{}. While \gnorm{} yields similar or higher Spearman correlation and lower regret than \gradcv{} on \math{} Perturb (Simple), a mild OOD benchmark, \gradcv{} outperforms \gnorm{} on both \math{}$^2$ and \math{} Perturb (Hard), which represent substantially more challenging OOD shifts.}
    \label{tab:math_perturb_llama}
\end{table}

\subsection{How does knowing the teacher performance change the behavior of the different scores?}
\label{app:teacher_performance}

\looseness-1Our main results focus on the setting where no verifier is used to pre-filter the teachers. A key question is whether the teacher’s own performance influences the correlation between \gradcv{} and student performance, and if teacher performance were known, whether it could further improve the predictive power of various scores. To examine this, we analyze how the behavior of \gradcv{} varies when teachers are grouped into different performance bins.

To verify, we divide all our teachers into three equal bins based on their own performance. In each bin, we compute the spearman correlation of each metric to the student performance post distillation training. We also report the regret of a score within each bin as the performance gap between (i) the best student performance achieved after distilling from all teachers in that performance bin and (ii) the student performance obtained when the teacher is chosen according to that score. 

\begin{figure}[h]
    \centering
    \includegraphics[width=0.49\linewidth]{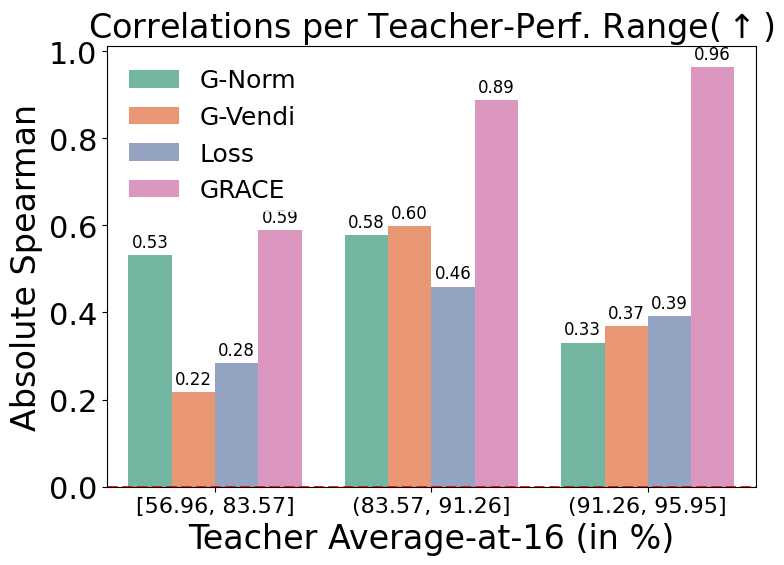}
    \includegraphics[width=0.49\linewidth]{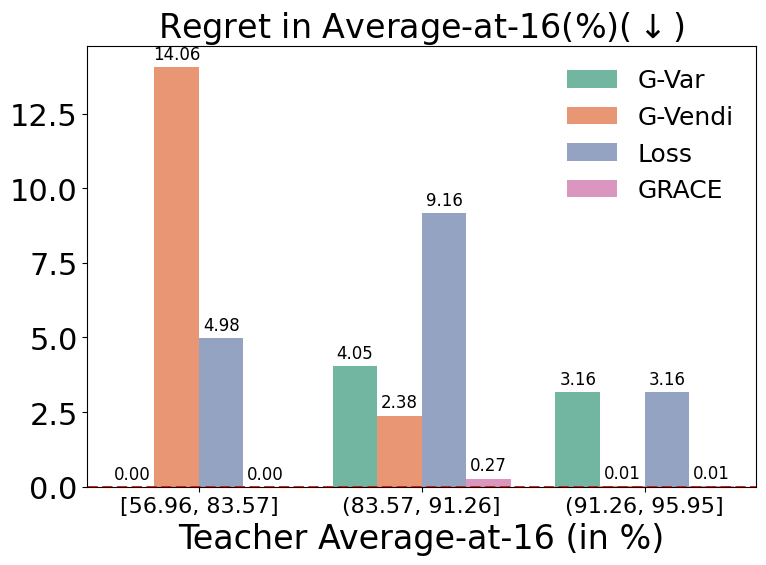}
    \caption{Setting: \llamaS{} training on \gsm{}. When we \textbf{divide teachers into three equal bins} based on their own performance, we compare both \textbf{Spearman correlation} and \textbf{Regret}. Across all bins, \gradcv{} shows a stronger Spearman correlation with student performance after distillation and consistently shows minimum regret. Its advantage is more pronounced when distinguishing among high-performing (top performance bin) teachers.}
    \label{fig:binning_w_teacherperf}
\end{figure}

\begin{figure}[h]
    \centering
    \includegraphics[width=0.49\linewidth]{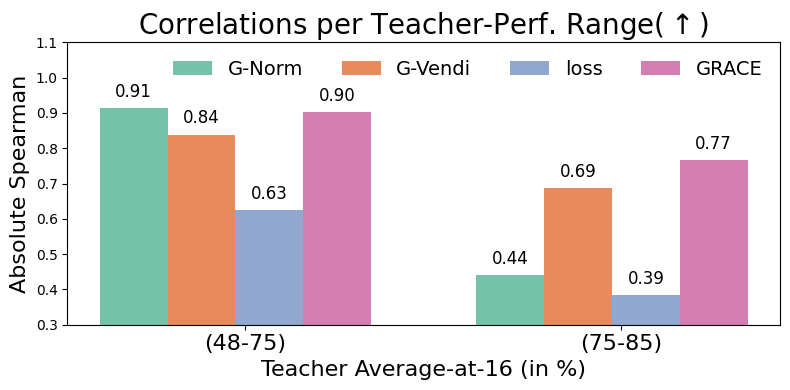}
    \includegraphics[width=0.49\linewidth]{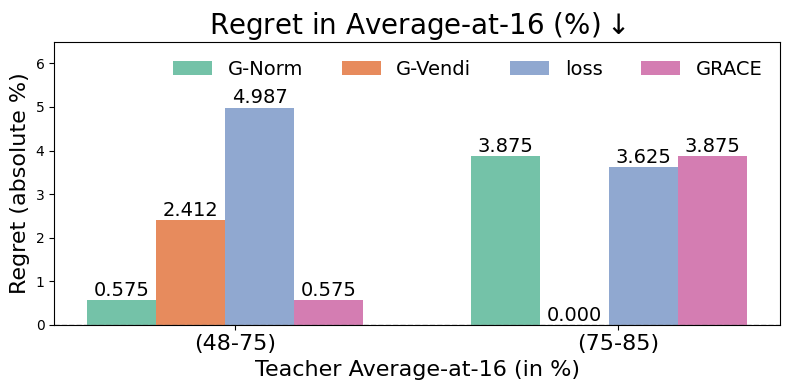}
    \caption{Setting: \llamaM{} training on \math{}. We split teachers into two equal bins according to their own performance and evaluate each bin using both \textbf{Spearman correlation} and \textbf{regret}. Across bins, \gradcv{} maintains a stronger correlation with the student’s post-distillation performance. However, in the top-performing bin, \gvendi{} achieves lower regret than \gradcv{}. }
    \label{fig:binning_w_teacherperf_math}
\end{figure}

\begin{itemize}
    \item In \cref{fig:binning_w_teacherperf}, we show that \gradcv{} shows better spearman correlation and also lower regret, across all the performance bins for \llamaS{} training on \gsm{}. Advantage of \gradcv{} are particularly more pronounced for distinguishing high performing teachers.
    \item In \cref{fig:binning_w_teacherperf_math}, for \llamaM{} training on \math{}, \gradcv{} maintains stronger Spearman correlation with student performance across all bins. However, \gvendi{} shows lower regret than \gradcv{} in the highest-performance teacher bin.
\end{itemize}

\subsection{Measuring correlations with other measures of student performance}
\label{app:other_student_performance}

We primarily measured the correlation of the metrics to Average-at-16 performance of the student post distillation, when we generate responses at temperature $1.0$. A key question is: how do the correlations change when the performance metric of the student is changed? To verify, we use two other commonly used performance metrics -- greedy student performance and Best-of-16 performance. Greedy performance measures correctness of single student's generation response per prompt at temperature $0.0$. Best-of-16 performance measures whether the student gets atleast one correct response, when generating $16$ responses per prompt at temperature $1.0$. In \cref{tab:greedy_and_bestof16_gradient_temp}, we observe that the spearman correlation of \gradcv{} to the student performance drops, when we switch performance measure to greedy accuracy or Best-of-16 performance. However, \gradcv{} still achieves small regret with the new performance metrics, compared to \gnorm{} and \gvendi{}.

\begin{table}[h]
    \centering
    \begin{tabular}{c|cc|cc|cc}
        Metric & \multicolumn{2}{c|}{Average-at-16 ($\tau=1.0$)} & \multicolumn{2}{c|}{Greedy accuracy} & \multicolumn{2}{c}{Best-of-16 ($\tau=1.0$)} \\
        \toprule
        & Abs. Spearman& Regret & Abs. Spearman & Regret   & Abs. Spearman  & Regret  \\
        &  ($\uparrow$) &  ($\downarrow$) &  ($\uparrow$) &  ($\downarrow$) &  ($\uparrow$) &  ($\downarrow$) \\
        \midrule
         \gvendi{} & 0.44 & 14.5 $\%$ & 0.65 & 4.5$\%$ & \textbf{0.68} & $4.5\%$ \\
         \gnorm{} &  0.55 & 4.9 $\%$ & 0.33 & 13.5$\%$ & 0.30 & $7.5\%$ \\
         \gradcv{} &  \textbf{0.86} & \textbf{0.3}$\%$ & \textbf{0.70} & \textbf{2.4}$\%$ & 0.64 & \textbf{0}$\%$ \\
        \bottomrule
    \end{tabular}
    \caption{Comparing \gvendi{}, \gnorm{}, and \gradcv{} when measuring average-at-16, greedy, and best-of-16 performance of the student post distillation. Here, we consider \llamaS{} training on \gsm{} (setting of \cref{fig:across_all_temps}). The spearman correlation for \gradcv{} sharply drops when measuring  greedy, and best-of-16 performance of the student post distillation. However, regret from using \gradcv{} to pick the best teacher remains smaller than the regret of \gnorm{} and \gvendi{}.
    }
    \label{tab:greedy_and_bestof16_gradient_temp}
\end{table}

\begin{figure}[h]
    \centering
    \includegraphics[width=\linewidth]{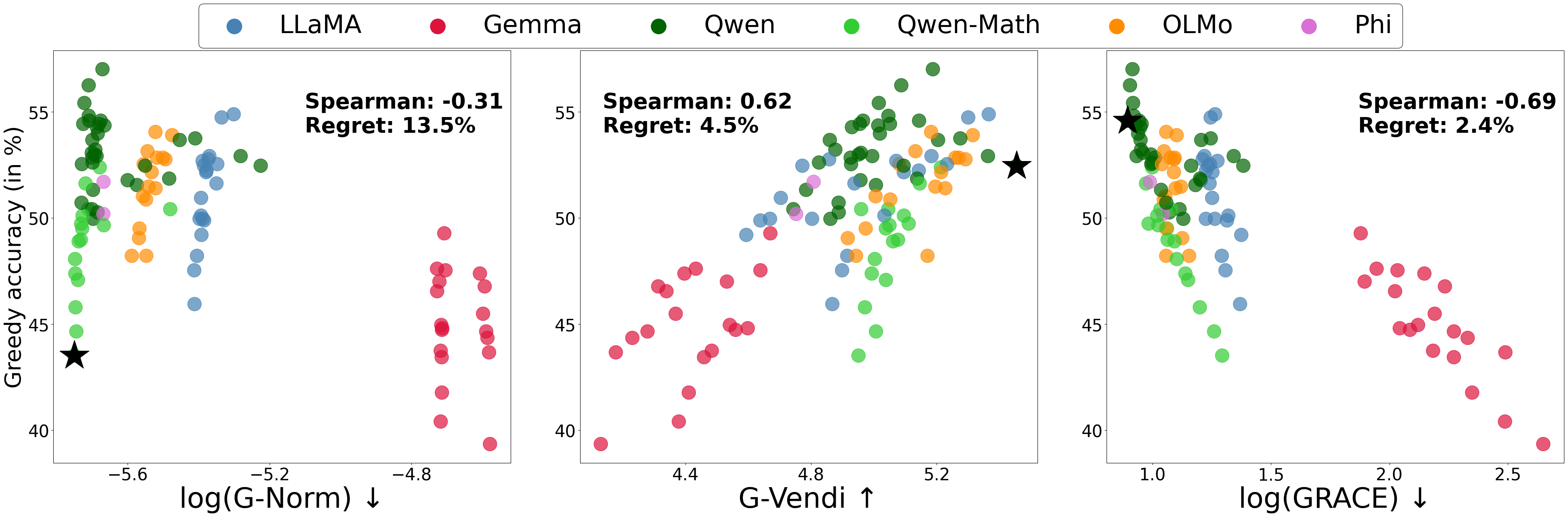}
    \caption{Repeated experiment from \cref{fig:across_all_temps} but greedy performance of trained student model. \textbf{\gradcv{} achieves only $70\%$ correlation to \llamaS{} performance after training on \gsm{}, across all teacher, generation temperature combinations.} This is a sharp reduction from $86\%$ correlation to Average-at-16. However, \gradcv{} still predicts close to the optimal teacher.
    }
    \label{fig:greedy_gradient_temp}
\end{figure}

\begin{figure}[h]
    \centering
    \includegraphics[width=\linewidth]{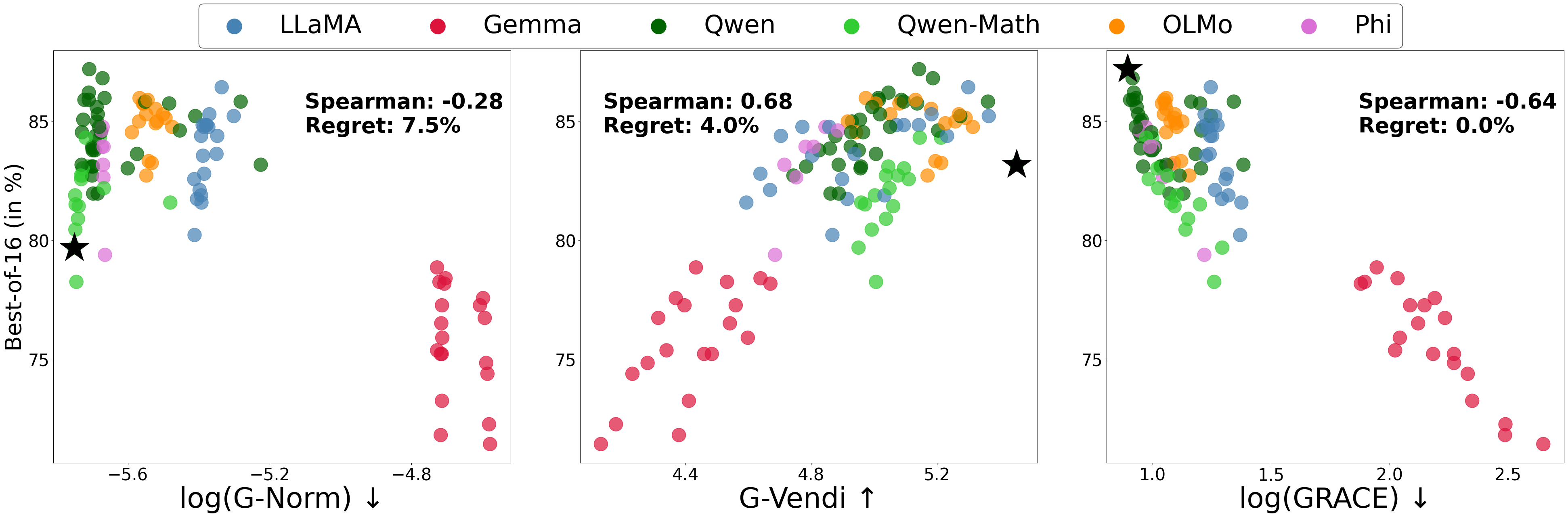}
    \caption{Repeated experiment from \cref{fig:across_all_temps} but best-of-16 performance of trained student model. \textbf{\gradcv{} achieves only $65\%$ correlation to \llamaS{} performance after training on \gsm{}, across all teacher, generation temperature combinations.} This is a sharp reduction from $86\%$ correlation to Average-at-16. However, \gradcv{} still predicts the optimal teacher.} 
    \label{fig:bestof16_gradient_temp}
\end{figure}


\subsection{Preliminary exploration: General NLP tasks beyond math}
\label{app:ARC}

One question is whether the predictive nature of \gradcv{} holds beyond math based tasks. We study ARC \citep{allenai:arc}. ARC is a general reasoning task, where the model is given a question and 4 choices to choose from to answer. An example of a question in ARC is as follows:

\fcolorbox{black}{gray!10}{%
  \parbox{\textwidth}{%
    A fold observed in layers of sedimentary rock most likely resulted from the:\\
    (a) \textit{cooling of flowing magma.}\\
    (b) \textit{converging of crustal plates.}\\
    (c) \textit{deposition of river sediments.}\\
    (d) \textit{solution of carbonate minerals.}
  }%
}

\begin{table}[h]
    \centering
    \begin{tabular}{c|c|c}
         Score & Absolute Spearman ($\uparrow$) & Regret ($\downarrow$)  \\
         & & (Best Average-at-16\\
         & &  student: $61\%$) \\
         \toprule
         \gvendi{} & 0.11 & 19.1 $\%$\\
         \gnorm{}  & 0.43 & \underline{2.4} $\%$ \\
         \gradcv{} & \underline{0.56} & \textbf{1.5}$\%$ \\
         Determinant & $0.09$ & 28.9 $\%$ \\
         BADGE & $0.19$ & 28.9 $\%$ \\
         Gradient inner product & $0.09$ & 6.4 $\%$\\
         Gradient cosine & $0.19$ & 28.9 $\%$ \\
         Student's loss on teacher's generations & 0.14 &  $12.9\%$  \\
         Average (token) probabilities & 0.09 & $11.5\%$ \\
         Average length & 0.06 & $6.4\%$ \\
         \midrule 
         Teacher average-at-16 performance & \textbf{0.82} & \underline{2.4} $\%$ \\
         \bottomrule
    \end{tabular}
    \caption{\llamaS{} training on ARC. \textbf{Among all scores that do not rely on a verifier for teacher filtering, \gradcv{} achieves the best Spearman correlation and regret in identifying the optimal teacher.}
    While teacher performance, that assumes access to a verifier, shows much stronger correlation with student performance after training, \gradcv{} still yields lower regret and remains the more effective criterion for teacher selection.}
    \label{tab:values_on_arc}
\end{table}

\textbf{Experiment Setting:} We train a \llamaS{} model and use the same set of teachers as in our \gsm{} and \math{} experiments, considering a subset of generation temperatures ${0.4, 0.6, 0.8, 1.0}$. Each teacher is prompted to provide the correct answer explicitly, rather than just the choice number. Nonetheless, some teachers still respond with the choice number only. To account for this, our evaluation accepts either the choice number or the full answer. For choice numbers, we apply exact match, while for textual answers, we compute the edit distance to each of the four options to determine correctness. We sample $16$ responses per prompt from each teacher to construct the training dataset and evaluate performance on the ARC validation split.

\textbf{Primary observations:} We observe that the teacher's performance itself is a strong indicator of the student performance after training. However, among all scores that do not assume access to such verifier based filtering on the teacher, \gradcv{} returns the best correlation to the student performance (however, lower than spearman correlation it could achieve on math based tasks). Among all scores (teacher performance included), \gradcv{} returns lower regret in selecting the best teacher for training the student. 

These observations indicate that \gradcv{} is not a perfectly predictive measure of student performance after training. However, it remains highly effective for selecting strong teachers, consistently yielding lower regret compared to other metrics. For instance, in our experiments, Gemma-2B teachers exhibited much higher (worse) \gradcv{} scores ($10^{-3}$) than Qwen-Math models ($10^{-4}$), yet students distilled from Gemma-2B teachers ($56.4\%$) outperformed those trained from Qwen-Math ($41.0\%$) . Conversely, directly choosing the highest-performing teacher (Llama-70B-instruct) led to a student with an average-at-16 score of $58.6\%$, whereas selecting based on \gradcv{} produced a student reaching $59.5\%$. 

\looseness-1This pattern stands in contrast to our findings on math-related tasks, where \gradcv{} was highly predictive of student performance after distillation and also could reliably identify close-to-optimal teacher. We keep extensive search beyond math-related tasks to future work.

\fcolorbox{black}{orange!10}{%
  \parbox{\textwidth}{%
Overall, \textbf{\gradcv{} remains a reliable score for teacher selection}, even though teachers with low \gradcv{} values may yield strong students—leading to slightly weaker correlations. Identifying the conditions under which \gradcv{} serves as a near-perfect predictor of student performance remains an important open question for future work.
  }%
}


\subsection{Computational complexity}
\label{app:compute}

\gradcv{} is computationally inexpensive to compute.
As shown in \Cref{tab:compute}, for $\numprompts = \gradientdim = 512$ and $\numgenerations = 4$, the gradients for each model takes around 10 minutes to compute and around 4.3MB to store. Computing the \gradcv{} score given the gradients of the student w.r.t. a teacher is inexpensive and takes roughly $10$ seconds.

\begin{table}[h]
    \centering
    \begin{tabular}{c|c|c}
    & Gradient Features Computation &  Metric Computation \\
    \midrule 
     Computation complexity & $\mathcal{O}(\numprompts \cdot \numgenerations \cdot \modelparams \cdot \gradientdim)$ & $\mathcal{O} (\numprompts \cdot \numgenerations \cdot \gradientdim^2 + \gradientdim^3)$ \\
     Running time & $\approx 10$ minutes & $<10$ seconds \\
     \midrule
     Storage Complexity & $\mathcal{O}(\numprompts \cdot \numgenerations \cdot \gradientdim)$ & - \\
     Actual storage & $4.3$ MB  & - \\
    \bottomrule
    \end{tabular}
    \caption{Time complexity to compute $\gradcv$. The running time and the actual storage have been computed on $\Tilde{\numprompts}=512$, $\numgenerations=4$, $\gradientdim=512$ for \llamaS{}  training on \gsm{}, and have been reported as a rough average across all settings. Wall-clock time has been reported on a single H100 (80 GB) GPU. For gradient computation, we use $32$ parallel CPU threads following \cite{park2023trak}. Here, $\modelparams$ denotes the number of parameters in the model.}
    \label{tab:compute}
\end{table}

\section{Ablations}

\begin{figure}[t]
    \centering
    \includegraphics[width=0.9\linewidth]{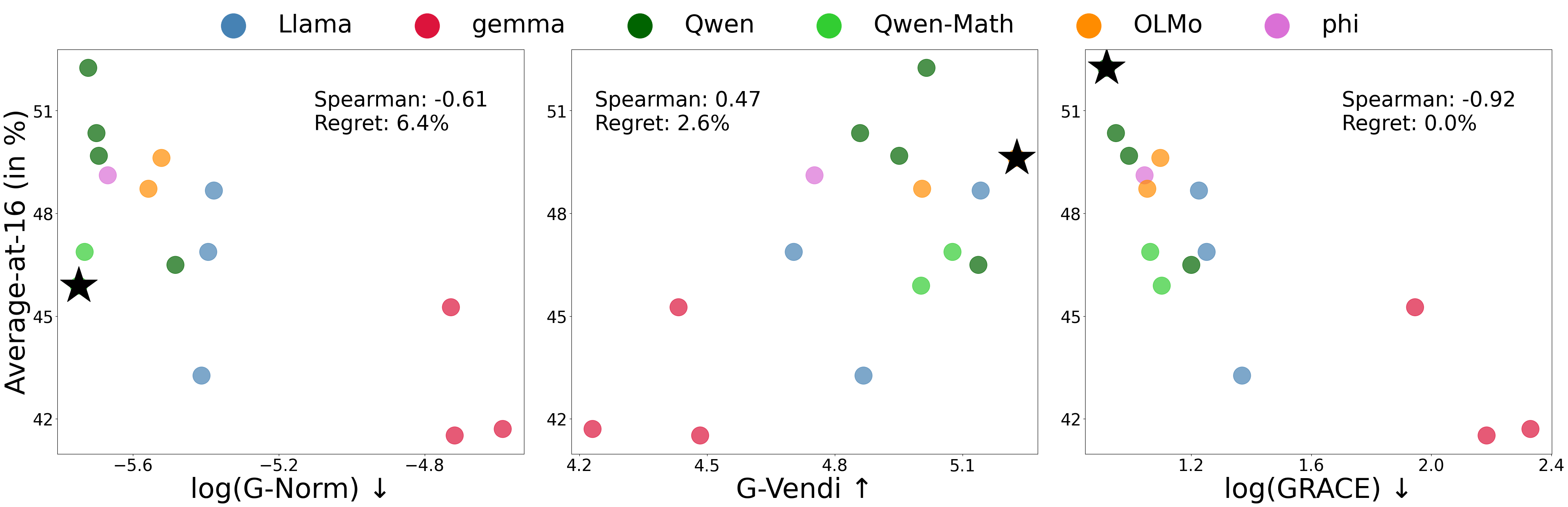}
    \caption{\textbf{Qualitative study: \gradcv{} can effectively correlate with student performance when compared across different teacher choices.}
    Here, we report \llamaS{} performance on \gsm{} across different teacher choices at a generation temperature $0.6$. \gnorm{} can capture differences between Llama and Qwen teacher families, however can't differentiate between Qwen, Qwen-Math and Phi teacher families.  \gvendi{} provides better separation among these families, but
    assigns higher scores to sub-optimal teachers. \gradcv{} achieves 90\% correlation with student performance after training, while also predicting Qwen-3B-Instruct to be the optimal teacher.
    } 
    \label{fig:comparison_singletemp}
\end{figure}

\subsection{Filtering v/s No filtering} \label{sec:filter_v_nofilter}
 In our experiments in the main paper, we perform no filtering of the responses from the teacher. Here, we compare to the case when we filter the teacher's responses by correctness. We sample $16$ responses from each teacher and remove the incorrect responses. Then, we sample with repetition to get a set of $16$ responses to train the model.

\begin{figure}[htbp]
     \centering
     \includegraphics[width=0.5\linewidth]{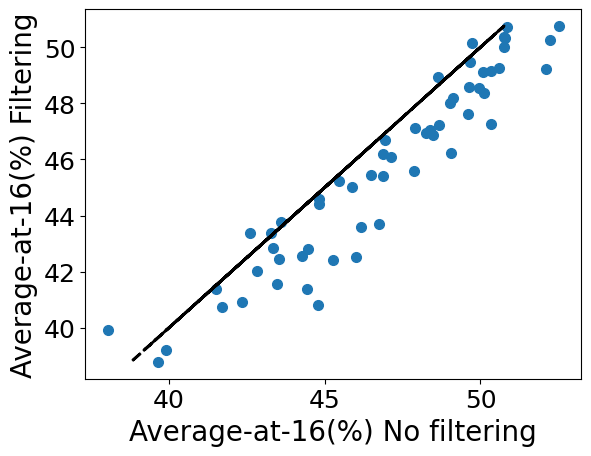}
     \caption{Comparing teachers, when we filter correct responses from the teacher v/s when we don't filter correct responses from the teacher. Here, we train \llamaS{} on \gsm{} with 15 teachers and generation temperatures 0.4, 0.6, 0.8, 1.0. We compare students trained from teacher without filtering (x-axis) with students trained from teacher with correct answer filtering  (y-axis). We find that students trained with no filtering outperforms models trained with filtering.}
     \label{fig:filter_v_nofilter}
 \end{figure}

\begin{figure}[htbp]
    \centering
    \begin{minipage}{\linewidth}
        \centering
        \includegraphics[width=\linewidth]{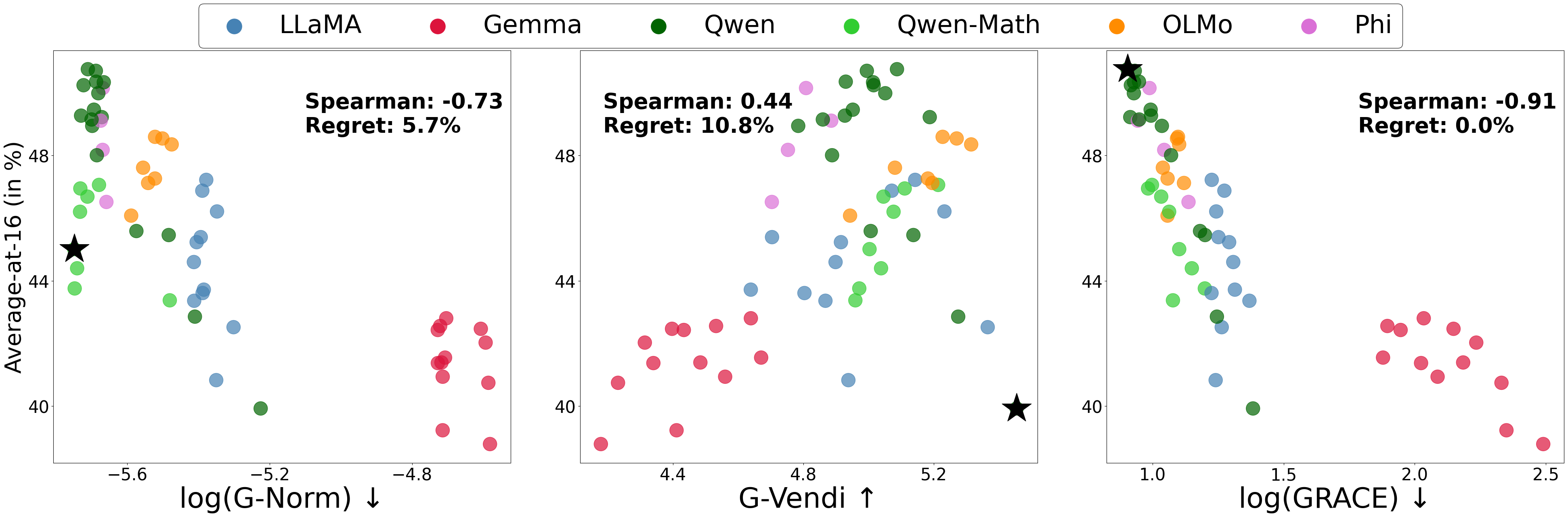}
    \end{minipage}\vfill
    \begin{minipage}{\linewidth}
        \centering
        \includegraphics[width=\linewidth]{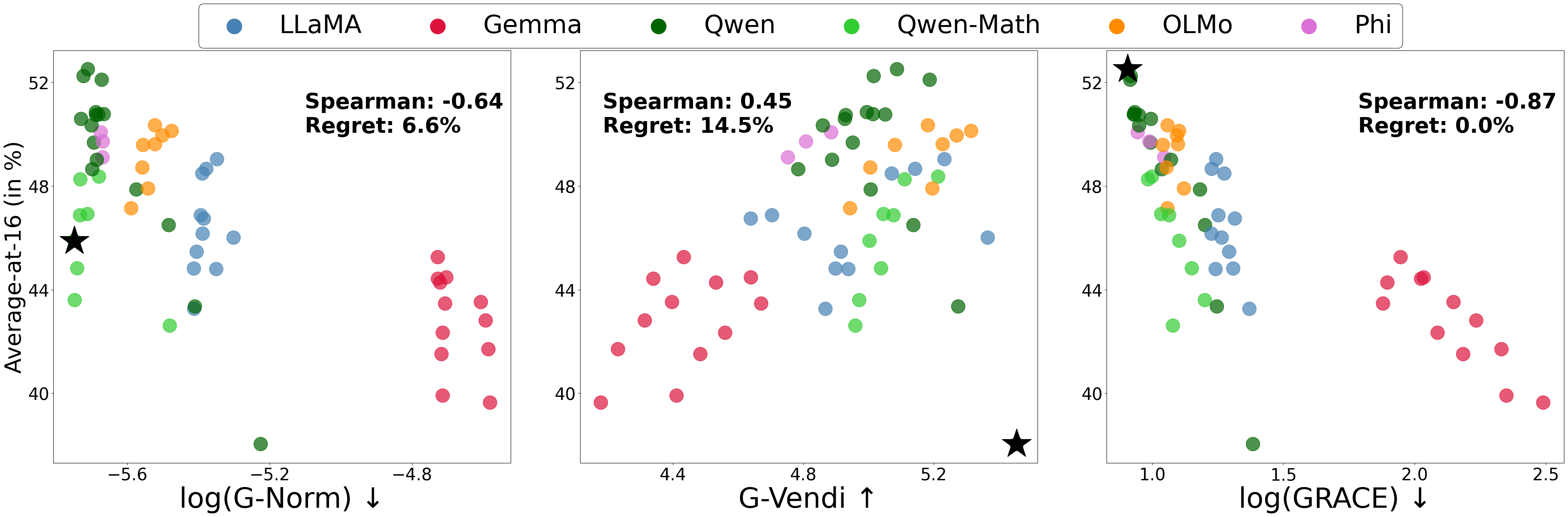}
    \end{minipage}
    \caption{Comparisons between the metrics and the student performance when we filter responses v/s we don't filter correct responses from the teacher. Here, we train \llamaS{} on \gsm{} with 15 teachers and generation temperatures 0.4, 0.6, 0.8, 1.0.  We find that our metrics have slightly higher spearman correlation to the student performance when we filter correct responses from the teacher and train only on them.}
    \label{fig:gsm8k_filter_vs_nofilter}
\end{figure}

 First, we find that the student gets worse performance with filtering of correct responses from the teacher (\cref{fig:filter_v_nofilter}). However, we find that when we compare our metrics to the student performance after training, we find that our metrics have slightly higher spearman correlation with the student performance when we train with filtering on teacher responses, compared to student trained with no filtering on the teacher responses (\cref{fig:gsm8k_filter_vs_nofilter}).

\subsection{Ablation on training hyperparameters}
\label{app:training_ablation}

\begin{figure}[htbp]
    \centering
    \includegraphics[width=0.7\linewidth]{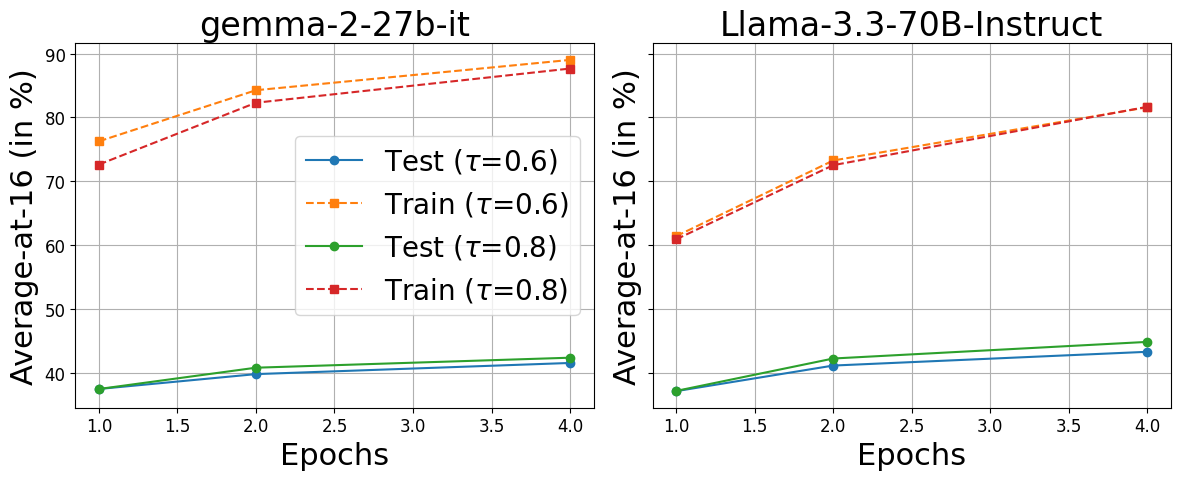}
    \caption{\llamaS{} training on \gsm{} with $16$ responses per prompt of gemma-27b-instruct and llama-70b instruct model. We vary the number of epochs and observe that both train and test performance improves with more epochs of training.  } 
    \label{fig:longer_runs_gsm}
\end{figure}

\begin{figure}[htbp]
    \centering
    \includegraphics[width=0.7\linewidth]{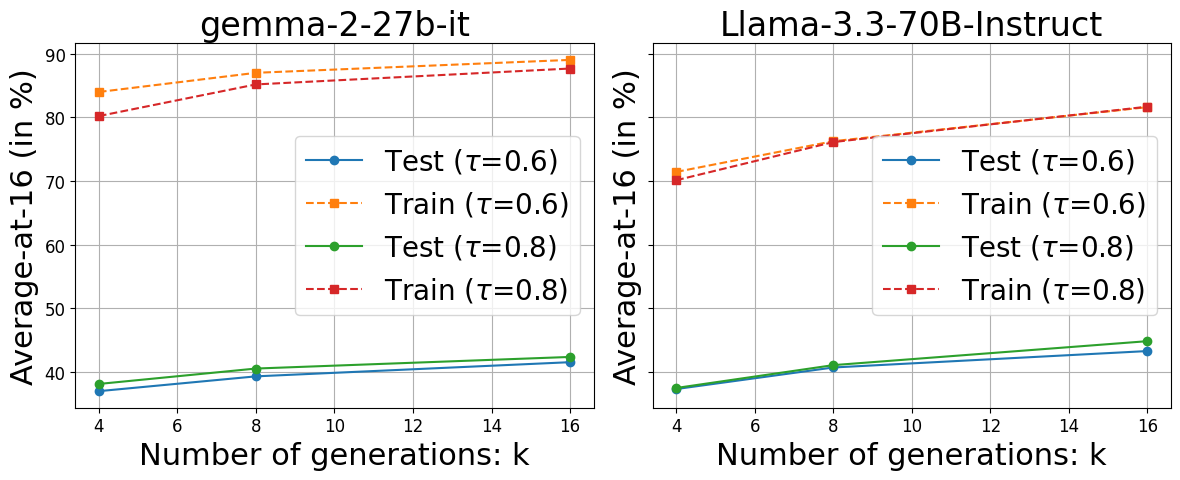}
    \caption{\llamaS{} training on \gsm{} with varying number of  responses per prompt of gemma-27b-instruct and llama-70b instruct model. We observe that both train and test performance improves with more training samples from the teacher.  } 
    \label{fig:more_gen_gsm}
\end{figure}

We observe that a Llama-1B model trained on generations of Llama-70B Instruct models and Gemma-2-27B Instruct models perform badly. We train with learning $1e0^{-5}$ on the $16$ generations per prompt of the teacher for $4$ epochs. One primary question is whether the small model is over-optimizing on the teacher's generations. To check this, we track the train and test performance of the trained model with varying number of generations (\cref{fig:longer_runs_gsm}) and epochs of training (\cref{fig:more_gen_gsm}). We observe that the performance of the trained student model improves with increasing number of epochs and number of generations, implying no over-optimization in our training setting.

\subsection{Ablations on the parameters of \gradcv{}}
\label{app:ablation_hyperparams}

\begin{figure}[h]
    \centering
    \includegraphics[width=0.4\linewidth]{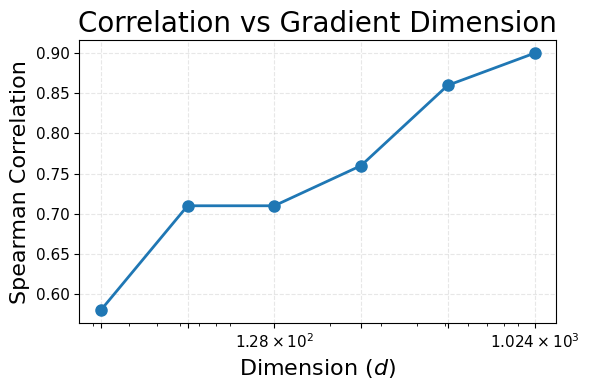}
    \includegraphics[width=0.4\linewidth]{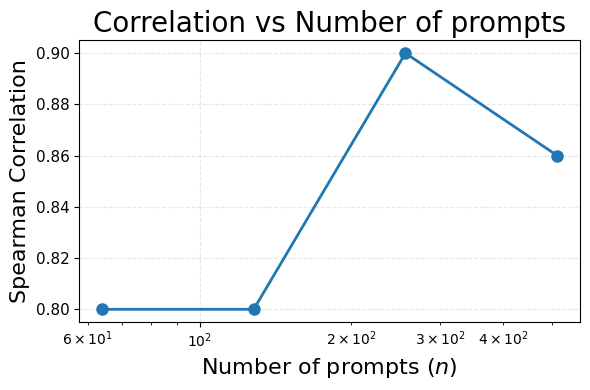}
    \caption{Varying hyperparameters for \gradcv{} on \llamaS{} training on \gsm{}. We use the base setup as $\numprompts=512$, and $\gradientdim=512$. We vary one of them, while fixing the other. Main takeaway: (a) \gradcv{} improves with increasing gradient dimension, (b) \gradcv{} generally increases with number of prompts that we consider but shows a small dip as we increase further. $\numprompts=\gradientdim/2$ gives the best results for spearman correlation.}
    \label{fig:ablations-fisher-gsm8k}
\end{figure}

In \cref{fig:ablations-fisher-gsm8k}, we use \llamaS{} trained on \gsm{} as a case study and ablate two key factors affecting \gradcv{}: the gradient projection dimension ($\gradientdim$) and the number of prompts ($\numprompts$) used to compute the score. With a fixed $\numprompts$, the correlation between \gradcv{} and student performance consistently improves as $\gradientdim$ increases. Conversely, when holding $\gradientdim$ fixed, correlation increases with $\numprompts$, peaking when $\numprompts=\gradientdim/2$.

We further vary the number of generations per prompt ($\numgenerations$) while keeping $\numprompts$ and $\gradientdim$ fixed. In our base configuration, using $\numgenerations = 2, 4, 8$ yields correlations of $0.70$, $0.86$, and $0.87$, respectively, suggesting that $\numgenerations = 4$ already provides a sufficiently strong estimate of \gradcv{}.

\begin{figure}
    \centering
    \includegraphics[width=0.48\linewidth]{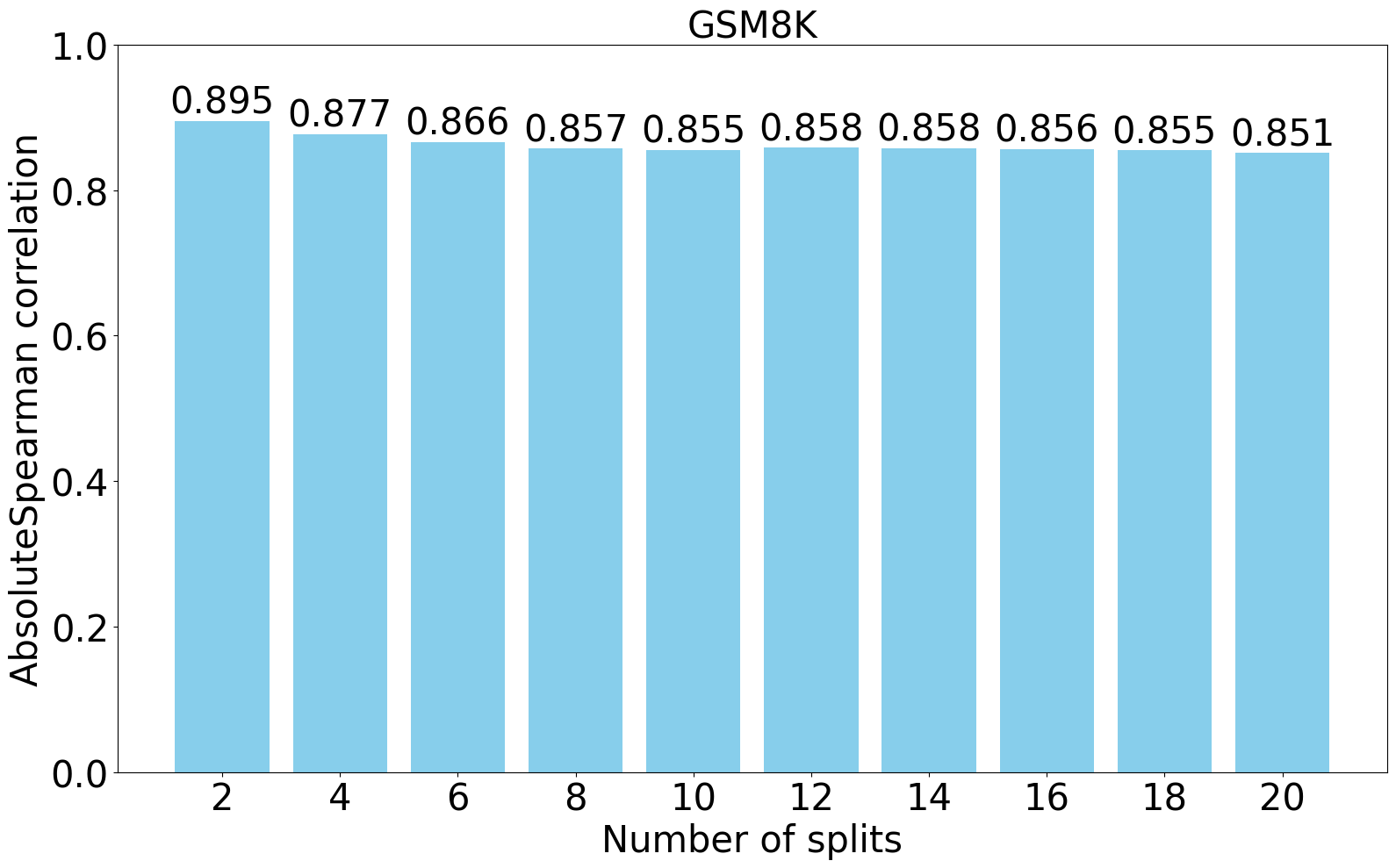}
    \includegraphics[width=0.48\linewidth]{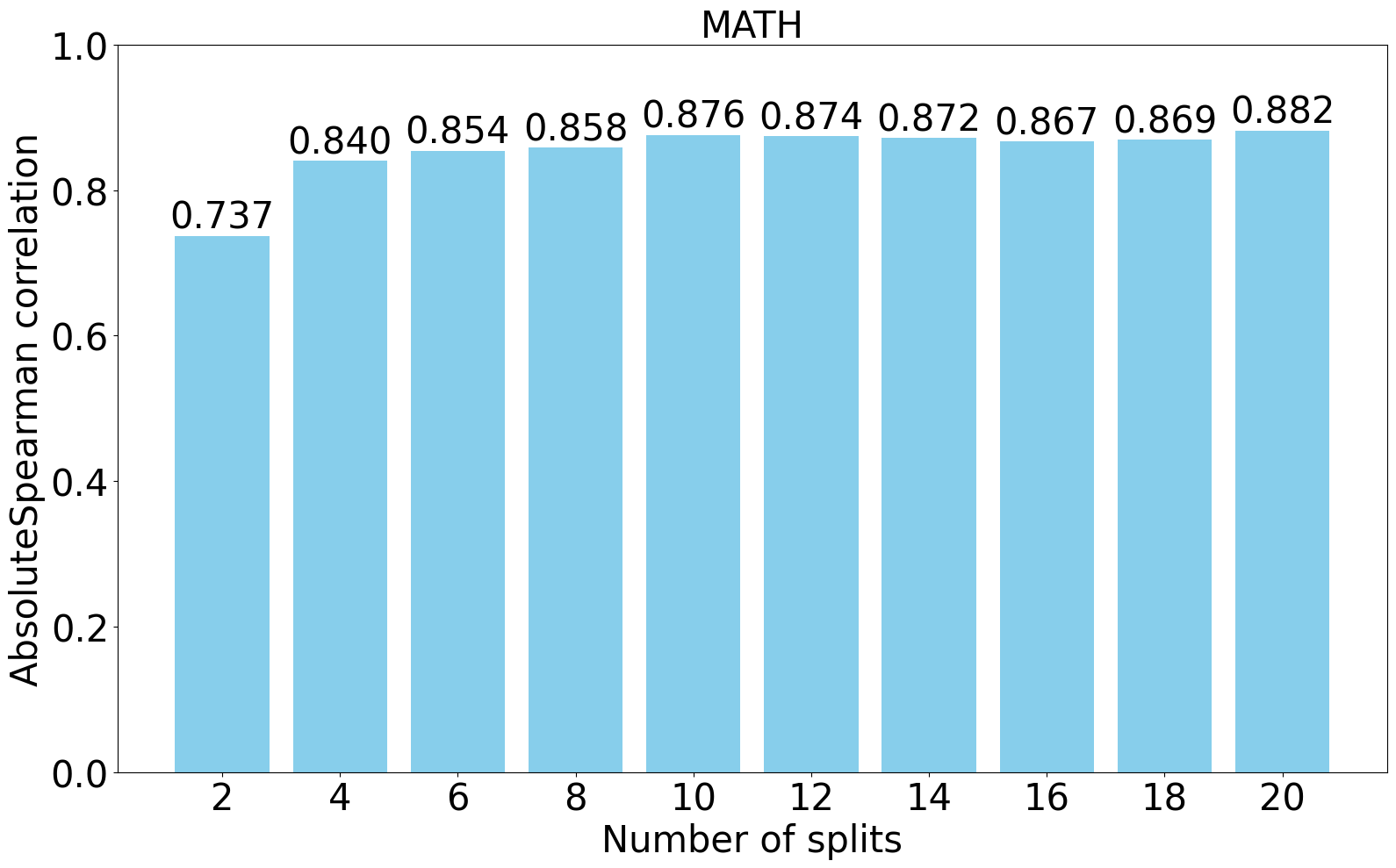}
    \caption{\textbf{Varying number of cross-validation splits} on GSM8K (left) and MATH (right).}
    \label{fig:varyC}
\end{figure}

We additionally vary the number of cross-validation splits used in \gradcv{}.
As shown in \Cref{fig:varyC}, the correlations to the student performance do not vary much for both GSM8k and MATH for more than $\numCV=6$ splits.
We take $\numCV=10$ as the default.

\subsection{Gradient norm's relation to length}
\label{app:grad_norm_length}

\begin{figure}
    \centering
    \includegraphics[width=0.75\linewidth]{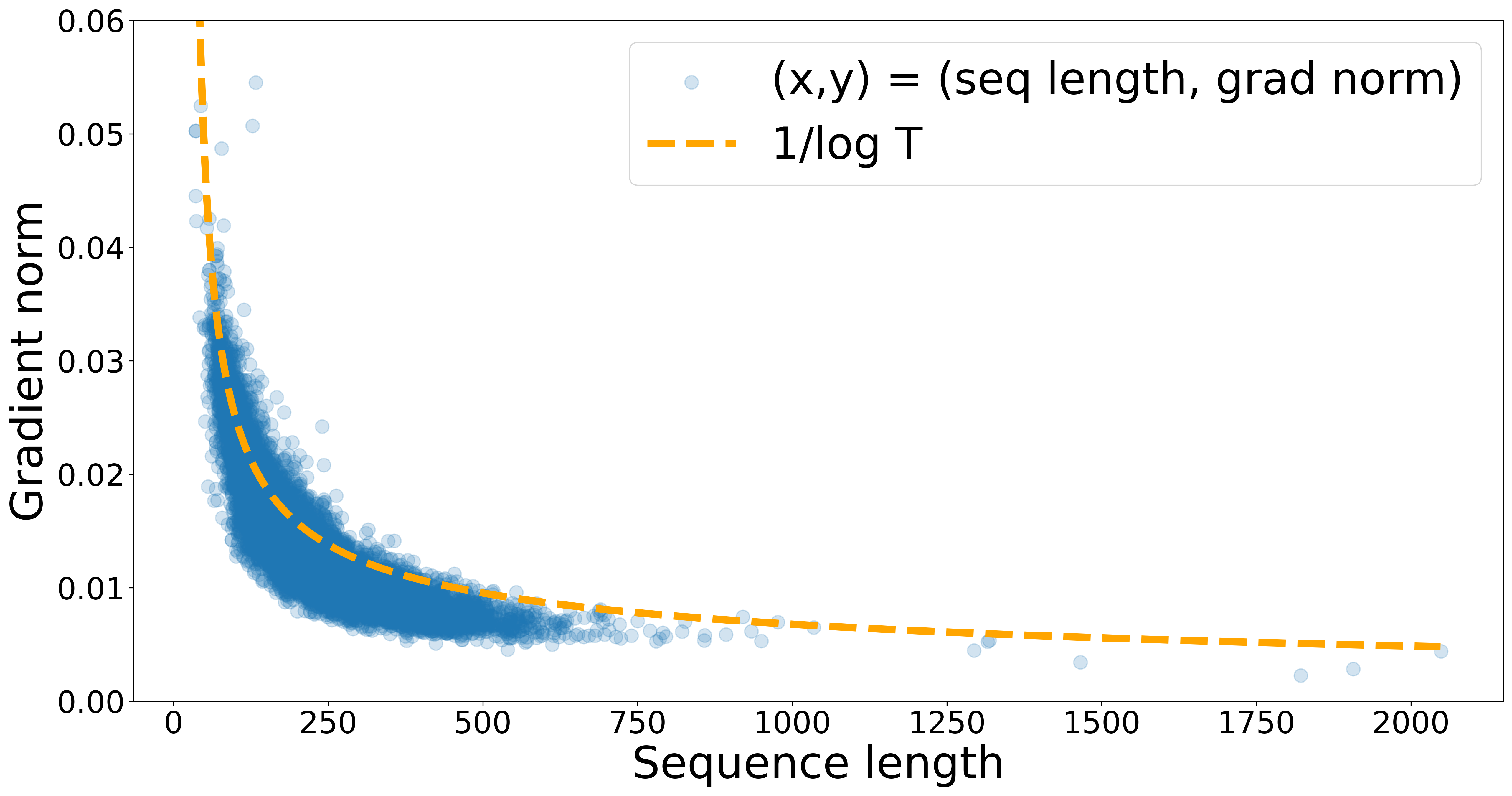}
    \caption{\textbf{Gradient norm decreases inversely with $\log T$}, where $T$ is the sequence length.
    This motivates the gradient scaling in \Cref{sec:grace}.}
    \label{fig:grad_norm_length}
\end{figure}

\Cref{fig:grad_norm_length} shows that the norm of the gradient (averaged over tokens) on a generation decreases as the generation length grows, roughly following a trend of $1/\log T$ for length-$T$ generations, consistent with observations in~\cite{xia24less}.
Intuitively, this is likely because longer generations tend to contain a larger fraction of less important tokens that do not contribute much to the overall gradient.
This observation motivates the $\log T$ scaling in \Cref{sec:grace}.

\subsection{Ablation on robustness of metrics}
\label{app:ablation_robust}

We check the robustness of each metric by reporting the distributions of the metric values computed over random subsets of teachers.
Specifically, we use 100 random draws of subsets consisting of 60\% of teachers.

We compare \gradcv{} against the baselines listed in~\Cref{app:other_metrics}, with an additional score of the average probabilities of the correct responses.
Among all candidate metrics, \gradcv{}~is the only one showing consistently strong correlations.

\begin{figure}[h]
    \centering
    \includegraphics[width=\linewidth]{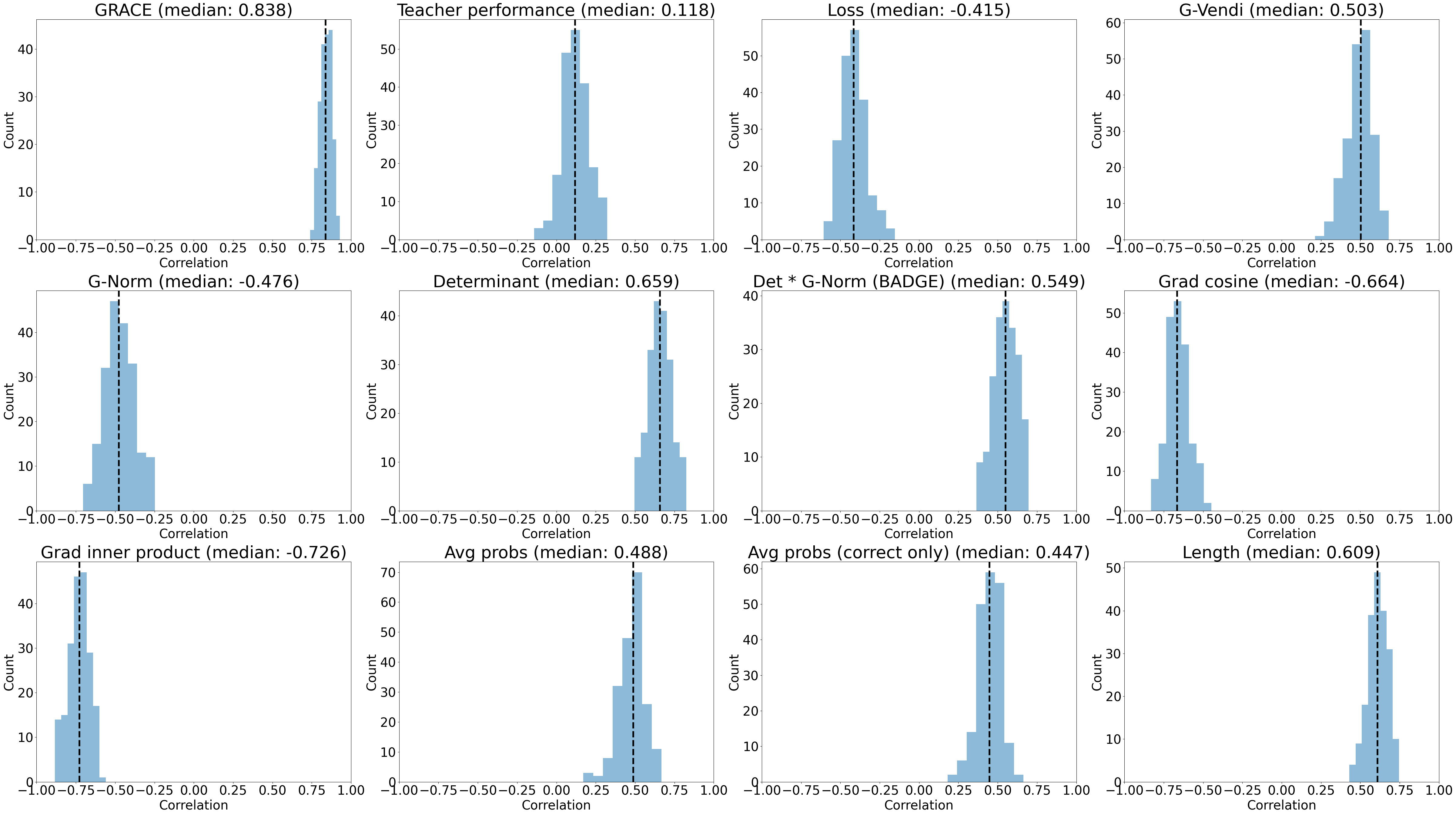}
    \caption{\textbf{Robustness of scores}: 
    We report the distribution of scores for \llamaS{} on \gsm, computed over 200 random subsets of teacher-temperature combinations, where each subset is 60\% of the full set.
    The dashed lines and numbers in parentheses mark the median.
    The proposed \gradcv{} score consistently shows stronger correlations over all other metrics.
    }
    \label{fig:ablation_robust_gsm}
\end{figure}







\section{Prompt format}

We train the student model with the following format across distillation from all teachers:

\fcolorbox{black}{gray!10}{%
  \parbox{\textwidth}{%
    $\#\#\#$ Problem: $\{$input$\}$\newline
    $\#\#\#$ Solution: $\{$output$\}$
  }%
}

One example is as follows:
\begin{tcolorbox}[
  colback=gray!10,
  colframe=black,
  boxrule=0.5pt,
  left=0.5pt,right=0.5pt,top=6pt,bottom=6pt,
  breakable,
]
{\ttfamily
\textbf{\#\#\# Problem:} Chandra has four bowls. Each one is a different color (red, blue, yellow, green). She also has exactly one glass the same color as each bowl. If she chooses a bowl and a glass from the cupboard, how many pairings are possible? One such pairing is a blue bowl and a yellow glass.

\textbf{\#\#\# Solution:}

\begin{verbatim}
To determine the total number of possible pairings between bowls and glasses, 
we need to consider the number of choices available for each item and then 
multiply these choices together.

Chandra has 4 bowls, and each bowl can be paired with any of the 4 glasses. 
Therefore, the number of possible pairings is calculated by multiplying the 
number of choices for the bowl by the number of choices for the glass. 
This can be expressed as:

\\[ 4 \\text{ (choices for bowls)} \\times 4 \\text{ (choices for glasses)} = 16 \\]

Thus, the total number of possible pairings is \\(\\boxed{16}\\).
\end{verbatim}
}
\end{tcolorbox}

We use the following formats to get responses from each teacher family, which have been taken from their model-cards on huggingface. We use vLLM 0.7.0 (\citep{kwon2023efficient}) to get responses from all models.

\begin{tcolorbox}[
  colback=green!5,
  colframe=black,
  boxrule=0.5pt,
  arc=2pt,
  left=0.5pt,right=0.5pt,top=6pt,bottom=6pt,
  width=\linewidth
]
\textbf{Prompt to LLaMA teacher models}

\begin{verbatim}
<|start_header_id|>system<|end_header_id|>You are a helpful assistant.<|eot_id|>
<|start_header_id|>user<|end_header_id|>
Please reason step by step, and put your final answer within \\boxed{{}}.
{input}
<|eot_id|>
<|start_header_id|>assistant<|end_header_id|>


\end{verbatim}
\end{tcolorbox}

\begin{tcolorbox}[
  colback=green!5,
  colframe=black,
  boxrule=0.5pt,
  arc=2pt,
  left=0.5pt,right=0.5pt,top=6pt,bottom=6pt,
  width=\linewidth
]
\textbf{Prompt to Qwen teacher models}

\begin{verbatim}
<|im_start|>system\nYou are a helpful assistant.<|im_end|>
<|im_start|>user
Please reason step by step, and put your final answer within \\boxed{{}}.
{input}
<|im_end|>
<|im_start|>assistant

\end{verbatim}
\end{tcolorbox}

\begin{tcolorbox}[
  colback=green!5,
  colframe=black,
  boxrule=0.5pt,
  arc=2pt,
  left=0.5pt,right=0.5pt,top=6pt,bottom=6pt,
  width=\linewidth
]
\textbf{Prompt to OLMo teacher models}

\begin{verbatim}
<|endoftext|><|user|>
Please reason step by step, and put your final answer within \\boxed{{}}.
{input}
<|assistant|>

\end{verbatim}
\end{tcolorbox}

\begin{tcolorbox}[
  colback=green!5,
  colframe=black,
  boxrule=0.5pt,
  arc=2pt,
  left=0.5pt,right=0.5pt,top=6pt,bottom=6pt,
  width=\linewidth
]
\textbf{Prompt to Phi teacher models}

\begin{verbatim}
<|im_start|>system<|im_sep|>
You are a medieval knight and must provide explanations to modern people.
<|im_end|>
<|im_start|>user<|im_sep|>
Please reason step by step, and put your final answer within \\boxed{{}}.
{input}
<|im_end|>
<|im_start|>assistant<|im_sep|>

\end{verbatim}
\end{tcolorbox}

\begin{tcolorbox}[
  colback=green!5,
  colframe=black,
  boxrule=0.5pt,
  arc=2pt,
  left=0.5pt,right=0.5pt,top=6pt,bottom=6pt,
  width=\linewidth
]
\textbf{Prompt to Gemma teacher models}

\begin{verbatim}
<bos><start_of_turn>user
You are a helpful assistant.
Please reason step by step, and put your final answer within \\boxed{{}}.
{input}
<end_of_turn>
<start_of_turn>model

\end{verbatim}
\end{tcolorbox}

\subsection{Qualitative analysis of responses from different teachers}
\label{app:response_analysis}

Here, we give qualitative analysis of responses from each teacher. In the table below, we show the average length of responses from different teachers. We observe that Gemma models give very concise responses, which leads to very high \gnorm{} and \gradcv{} (in \cref{fig:across_all_temps}). Response length can differentiate between the responses of different teacher families, however, it's not a perfect predictor (\cref{fig:comp_to_baselines,fig:comp_to_baselines_regret}).

We give examples of responses from Qwen-2.5 3B Instruct, LLaMA-3.3 70B Instruct, and Gemma-2 9B Instruct teachers to a question. We find that even though both Qwen-2.5 3B Instruct and LLaMA-3.3 70B Instruct qualitatively give similar amount of details in each step in their responses, Qwen teachers provide more details in their numerical computations.

\begin{table}[h]
\centering
\begin{tabular}{l r}
\toprule
\textbf{Teacher model} & \textbf{Average response length (across all temperatures)} \\
\midrule
\multicolumn{2}{l}{\textbf{LLaMA}} \\
\midrule
LLaMA-3.1-8B Instruct & 212.083 \\
LLaMA-3.2-3B Instruct & 224.492 \\
LLaMA-3.3-70B Instruct & 222.314 \\
\midrule
\multicolumn{2}{l}{\textbf{Qwen}} \\
\midrule
Qwen2.5-1.5B Instruct & 259.411 \\
Qwen2.5-3B Instruct & 289.952 \\
Qwen2.5-7B Instruct & 271.666 \\
Qwen2.5-14B Instruct & 287.611 \\
\midrule
\multicolumn{2}{l}{\textbf{Qwen-Math}} \\
\midrule
Qwen2.5-Math-1.5B Instruct & 272.988 \\
Qwen2.5-Math-7B Instruct & 332.119 \\
\midrule
\multicolumn{2}{l}{\textbf{Gemma-2}} \\
\midrule
Gemma-2-2b-it & 166.888 \\
Gemma-2-9b-it & 138.462 \\
Gemma-2-27b-it & 140.294 \\
\midrule
\multicolumn{2}{l}{\textbf{OLMo-2}} \\
\midrule
OLMo-2-7B Instruct & 286.604 \\
OLMo-2-13B Instruct & 252.683 \\
\midrule
\multicolumn{2}{l}{\textbf{Phi}} \\
\midrule
Phi-4 & 293.160 \\
\bottomrule
\end{tabular}
\caption{Average response lengths (in terms of number of tokens under the tokenizer of \llamaS{}) across all teachers on \gsm{} on  responses to a subsampled set of $512$ prompts from the training set.}
\end{table}

\begin{table}[h]
\centering
\begin{tabular}{l r}
\toprule
\textbf{Teacher model} & \textbf{Average response length (across all temperatures)} \\
\midrule
\multicolumn{2}{l}{\textbf{LLaMA}} \\
\midrule
LLaMA-3.1-8B Instruct & 519.514 \\
LLaMA-3.2-3B Instruct & 537.696 \\
LLaMA-3.3-70B Instruct & 529.951 \\
\midrule
\multicolumn{2}{l}{\textbf{Qwen}} \\
\midrule
Qwen2.5-1.5B Instruct & 572.480 \\
Qwen2.5-3B Instruct & 581.836 \\
Qwen2.5-7B Instruct & 561.407 \\
Qwen2.5-14B Instruct & 587.693 \\
\midrule
\multicolumn{2}{l}{\textbf{Qwen-Math}} \\
\midrule
Qwen2.5-Math-1.5B Instruct & 531.055 \\
Qwen2.5-Math-7B Instruct & 784.821 \\
\midrule
\multicolumn{2}{l}{\textbf{Gemma}} \\
\midrule
Gemma-2-2B Instruct & 357.610 \\
Gemma-2-9B Instruct & 341.838 \\
Gemma-2-27B Instruct & 355.279 \\
\midrule
\multicolumn{2}{l}{\textbf{OLMo}} \\
\midrule
OLMo-2-7B Instruct & 632.502 \\
OLMo-2-13B Instruct & 589.932 \\
\midrule
\multicolumn{2}{l}{\textbf{Phi}} \\
\midrule
Phi-4 & 666.602 \\
\bottomrule
\end{tabular}
\caption{Average response lengths (in terms of number of tokens under the tokenizer of \llamaM{}) across all teachers on \math{} on  responses to a subsampled set of $512$ prompts from the training set.}
\end{table}

\begin{tcolorbox}[
  colback=gray!5,
  colframe=black,
  boxrule=0.5pt,
  arc=2pt,
  left=0.5pt,right=0.5pt,top=6pt,bottom=6pt,
  width=\linewidth
]
\textbf{Question}

\begin{verbatim}
Janet’s ducks lay 16 eggs per day. She eats three for breakfast every 
morning and bakes muffins for her friends every day with four. 
She sells the remainder at the farmers' market daily for $2 per 
fresh duck egg. How much in dollars does she make every day 
at the farmers' market?
\end{verbatim}

\end{tcolorbox}

\begin{tcolorbox}[
  colback=blue!5,
  colframe=black,
  boxrule=0.5pt,
  arc=2pt,
  left=0.5pt,right=0.5pt,top=6pt,bottom=6pt,
  width=\linewidth
]

\textbf{Response from LLaMA-3.3-70B Instruct (temperature=$0.6$)}

\begin{verbatim}
To find out how much Janet makes every day at the farmers' market
, we first need to determine how many eggs she sells there. 
Here's the step-by-step process:

1. **Total eggs laid per day**: Janet’s ducks lay 16 eggs per day.

2. **Eggs eaten for breakfast**: She eats 3 eggs for breakfast every morning.

3. **Eggs used for baking muffins**: She uses 4 eggs to bake muffins 
for her friends every day.

4. **Total eggs used**: The total number of eggs she
uses (eats and bakes) is 3 (for breakfast) + 4 (for muffins) = 7 eggs.

5. **Eggs sold at the farmers' market**: The number of eggs she
sells at the market is the total number of eggs laid minus 
the eggs she uses. So, 16 (total eggs) - 7 (eggs used) = 9 eggs sold.

6. **Money made from selling eggs**: She sells each egg for $2. 

Therefore, the amount of money she makes 
from selling 9 eggs is 9 eggs * $2/egg = $18.

So, Janet makes $18 every day at the farmers' market.
The final answer is: $\\boxed{18}$
\end{verbatim}

\end{tcolorbox}

\begin{tcolorbox}[
  colback=purple!5,
  colframe=black,
  boxrule=0.5pt,
  arc=2pt,
  left=0.5pt,right=0.5pt,top=6pt,bottom=6pt,
  width=\linewidth
]

\textbf{Response from Gemma-2-9B Instruct (temperature=$0.6$)}

\begin{verbatim}
Here's how to solve the problem step-by-step:

1. **Eggs used:** Janet uses 3 eggs for breakfast + 4 eggs for muffins
= 7 eggs per day.

2. **Eggs remaining:** She has 16 eggs - 7 eggs = 9 eggs left for the market.

3. **Earnings:** She sells 9 eggs * $2 per egg = $18.

\\boxed{$18}
\end{verbatim}

\end{tcolorbox}

\begin{tcolorbox}[
  colback=orange!5,
  colframe=black,
  boxrule=0.5pt,
  arc=2pt,
  left=0.5pt,right=0.5pt,top=6pt,bottom=6pt,
  width=\linewidth
]

\textbf{Response from Qwen-2.5-3B Instruct (temperature=$0.6$)}

\begin{verbatim}
To determine how much Janet makes every day at the farmers' market,
we need to follow these steps:

1. Calculate the total number of eggs laid by the ducks per day.
2. Subtract the number of eggs Janet eats for breakfast.
3. Subtract the number of eggs used for baking muffins.
4. The remainder will be the number of eggs sold at the farmers' market.
5. Multiply the number of eggs sold by the price per egg to find 
out how much she makes each day.

Let's go through each step in detail:

1. The ducks lay 16 eggs per day.
2. Janet eats 3 eggs for breakfast.
3. Janet uses 4 eggs for baking muffins.

First, we calculate the number of eggs remaining after
Janet's consumption and baking:
\\[
16 \\text{ (total eggs)} - 3 \\text{ (eaten for breakfast)} - 
4 \\text{ (used for baking)} = 9 \\text{ (remaining eggs)}
\\]

Next, Janet sells the remaining 9 eggs at the farmers' market
for $2 per egg. So, we calculate her earnings:
\\[
9 \\text{ (remaining eggs)} \\times 2 \\text{ (dollars per egg)} 
= 18 \\text{ (dollars)}
\\]

Thus, Janet makes \\(\\boxed{18}\\) dollars every day at 
the farmers' market.
\end{verbatim}

\end{tcolorbox}

\end{document}